\documentclass[10pt,twocolumn,letterpaper]{article}

\usepackage{cvpr}

\usepackage{booktabs}
\usepackage{multirow} 
\usepackage[most]{tcolorbox}
\usepackage{amssymb}
\usepackage{amsmath}
\usepackage{graphicx}
\usepackage{algorithm}
\usepackage{algorithmic}

\definecolor{cvprblue}{rgb}{0.21,0.49,0.74}
\usepackage[pagebackref,breaklinks,colorlinks,allcolors=cvprblue]{hyperref}

\def\confName{CVPR}
\def\confYear{2027}

\title{DCR: Counterfactual Attractor Guidance for Rare Compositional Generation}

\author{
   Taewon Kang$^{1}$, Matthias Zwicker$^{1}$ \\
   $^{1}$University of Maryland at College Park, United States \\
   {\large \texttt{taewon@umd.edu, zwicker@umd.edu}} \\
}

\begin{document}
\maketitle

\begin{abstract}
    Diffusion models generate realistic visual content, yet often fail to produce \emph{rare but plausible compositions}. When prompted with combinations that are valid but underrepresented in training data (such as \emph{a snowy beach} or \emph{a rainbow at night}), the generation process frequently collapses toward more common alternatives. We identify this failure mode as \emph{default completion bias}, where denoising trajectories are implicitly attracted toward high-frequency semantic configurations. Existing guidance mechanisms do not explicitly model this competing tendency and therefore struggle to prevent such collapse. We introduce \textbf{Default Completion Repulsion (DCR)}, a training-free framework that explicitly models and suppresses default completion behavior. DCR constructs a \emph{counterfactual attractor} by relaxing the rare compositional factor while preserving surrounding semantics, inducing an alternative denoising trajectory reflecting the model's preferred completion. We define the discrepancy between target and attractor trajectories as a \emph{counterfactual drift}, and propose a projection-based repulsion mechanism that removes guidance components aligned with this drift direction. This suppresses undesired frequent completions while preserving other semantic components. DCR operates entirely within the standard diffusion sampling process without retraining or architectural modification. Experiments on rare compositional prompts show that DCR improves compositional fidelity while maintaining visual quality. Our analysis further shows that generations from the attractor prompt alone closely match the baseline outputs, confirming that it captures the model's default completion.
\end{abstract}

\section{Introduction}
\begin{figure}[t]
    \centering
    \includegraphics[width=\linewidth]{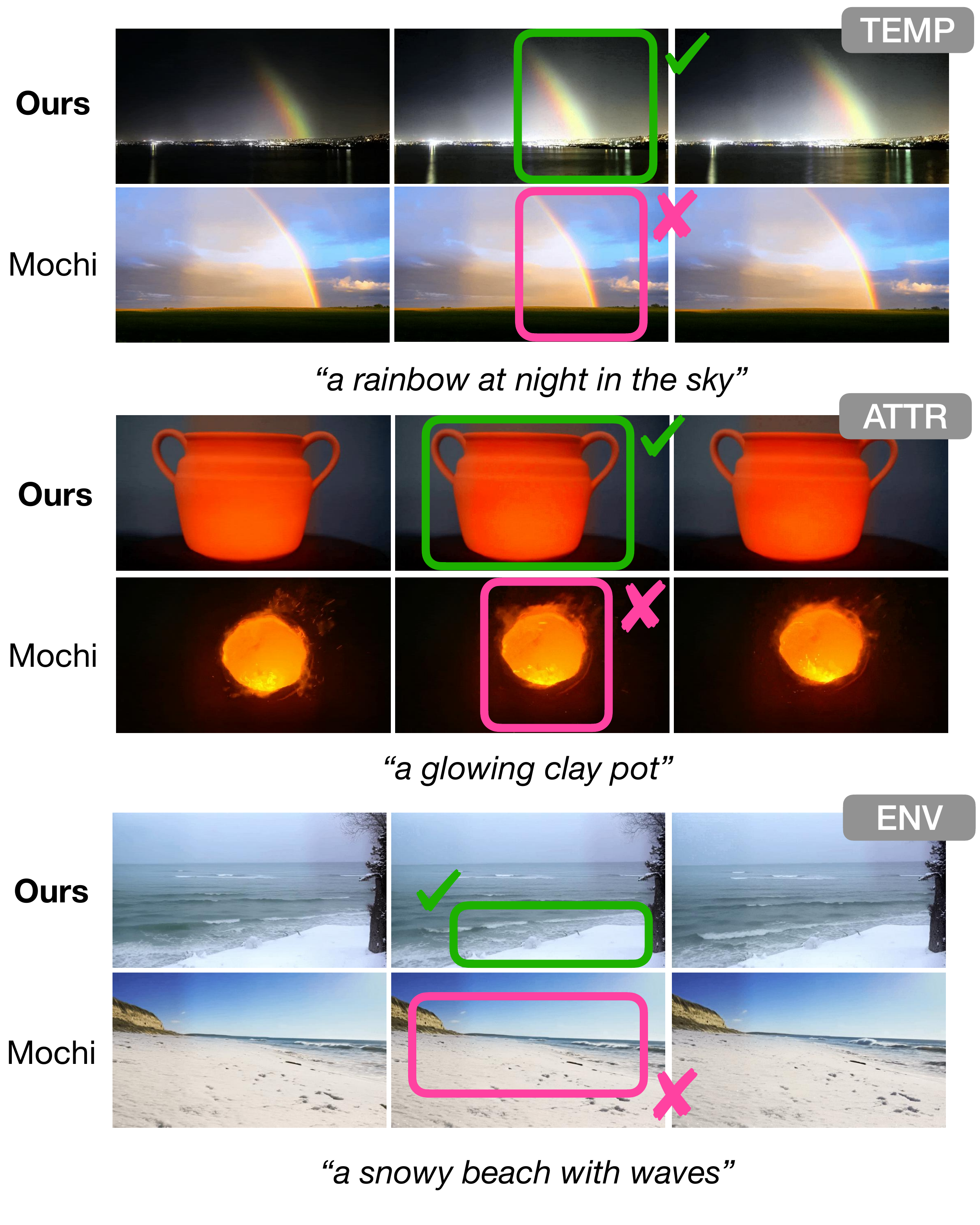}
    \caption{\textbf{Default completion bias in video diffusion models.} We show three rare compositional prompts, one per panel from top to bottom: Temporal Misalignment (TEMP, ``a rainbow at night in the sky''), Attribute Rebinding (ATTR, ``a glowing clay pot''), and Environment Recomposition (ENV, ``a snowy beach with waves''). Within each panel, the top row shows frames from our method (DCR) and the bottom row shows frames from Mochi; green and pink boxes mark a satisfied and a violated compositional factor. Mochi collapses toward statistically dominant completions, generating a daytime rainbow instead of a nocturnal one, a glowing sphere instead of a glowing clay pot, and a snow-free beach instead of a snowy one. DCR suppresses this default completion tendency via counterfactual attractor guidance and realizes both compositional factors in each case. Extended comparisons across all eight compositional categories and additional baselines are provided in Appendix~\ref{sec:detailed_qualitative}.}
    \label{fig:teaser}
\end{figure}

Diffusion models have achieved remarkable success in image and video generation~\cite{ho2020denoising, rombach2022high, ho2022video, yang2025cogvideox}. Despite this progress, they exhibit a persistent limitation: they struggle to generate \emph{rare but plausible compositions}~\cite{thrush2022winoground, yuksekgonul2022and, park2025rare}. When given prompts that combine semantically valid elements that rarely co-occur in training data (such as \emph{a penguin in a desert}, \emph{a snowy beach}, or \emph{a rainbow at night}), the generated outputs often deviate from the intended description and instead reflect more common visual patterns. This phenomenon can be understood as a form of \emph{compositional collapse}. During denoising, the model is implicitly biased toward high-frequency semantic configurations, causing the trajectory to drift away from the specified rare composition and toward a more likely alternative. This failure mode does not arise from ambiguous prompts or missing constraints but from the model's intrinsic preference for statistically dominant patterns. 
Existing approaches to controllable generation primarily focus on \emph{reinforcing the target signal}, for example by increasing guidance strength~\cite{ho2022classifier, saharia2022photorealistic}, introducing additional conditioning inputs, or modifying attention mechanisms~\cite{chefer2023attend, feng2022training}.
However, these methods do not explicitly account for the presence of competing completion tendencies within the model. As a result, even strong conditioning may fail to prevent collapse when the desired composition lies in a low-density region of the learned distribution.

In this work, we take the opposite approach: instead of strengthening the target signal, we identify and suppress the model's \emph{default completion tendency}. This tendency can be made explicit through a \emph{counterfactual attractor} constructed by relaxing the original prompt. By weakening the rare compositional factor while preserving the surrounding semantic context, the transformed prompt induces an alternative denoising trajectory that reflects the model's preferred completion. Building on this idea, we introduce \textbf{Default Completion Repulsion (DCR)}, a training-free framework that runs entirely within the standard diffusion sampling loop. At each denoising step, we compute both the target-guided update and the attractor-guided update, and define their difference as a \emph{counterfactual drift}. We then apply a projection-based repulsion mechanism that removes the component of the update aligned with this drift direction. This prevents the denoising trajectory from collapsing toward frequent alternatives while preserving other semantic components. A key property of our approach is that the attractor trajectory is not constrained to be collinear with the standard guidance direction. This allows the method to capture completion tendencies that are distinct from the guidance direction, rather than rescaling the guidance strength. We evaluate our method on a benchmark of rare compositional prompts spanning object-scene, attribute-scene, and lighting-condition combinations. Our results show that DCR significantly improves compositional fidelity compared to standard classifier-free guidance and other baselines, while maintaining high visual quality. Further analysis shows that generations conditioned on the attractor prompt alone closely match the baseline outputs, confirming that the attractor captures the model's default completion.

In summary, our contributions are as follows:
\begin{itemize}
    \item We identify \emph{compositional collapse} as a key failure mode in diffusion models for rare compositional generation.
    \item We propose a \emph{counterfactual attractor} formulation that explicitly models the model's default completion tendency via prompt relaxation.
    \item We introduce a projection-based repulsion mechanism that suppresses collapse during denoising.
    \item We demonstrate consistent improvements in compositional fidelity over standard CFG and negative-prompt baselines across all eight benchmark categories.
\end{itemize}

\section{Related Work}
\subsection{Guidance Mechanisms for Diffusion Models} 
Classifier-free guidance (CFG)~\cite{ho2022classifier} is the dominant mechanism for conditioning diffusion models on text prompts. Several variants address its limitations, including dynamic thresholding~\cite{saharia2022photorealistic}, interval-based scheduling~\cite{chung2024cfg++}, and training-free modifications such as FreeU~\cite{si2024freeu}, Universal Guidance~\cite{bansal2023universal}, energy-based constrained generation~\cite{zampini2025training}, multi-objective scheduling~\cite{xie2025dymo}, and constrained diffusion for safe control~\cite{zhang2025constrained}. Despite their effectiveness, these methods rescale or reweight the conditional signal without modeling competing semantic directions. They do not account for \emph{implicit attractors} induced by the training distribution, which bias generation toward frequent configurations. In contrast, our approach explicitly identifies default completion tendencies and applies targeted repulsive guidance. 

\subsection{Compositional and Negation-Based Generation} 
Generating outputs that faithfully combine multiple semantic concepts remains a central challenge~\cite{thrush2022winoground, yuksekgonul2022and, hsieh2023sugarcrepe}. Composable Diffusion~\cite{liu2022compositional} and GLoD~\cite{yamada2024glod} compose multiple concept-specific or layer-specific guidance signals to control global and local aspects of generation. Attend-and-Excite~\cite{chefer2023attend}, Structured Diffusion~\cite{feng2022training}, and linguistic binding~\cite{rassin2023linguistic} improve multi-concept consistency by reinforcing attention to specified tokens. However, these methods address conflicts \emph{between} explicitly specified concepts; they do not model the model's implicit tendency to favor statistically dominant alternatives when the target composition is rare. DCR is geometrically distinct from these approaches: rather than composing multiple guidance signals, it estimates the direction of the model's default completion tendency and removes its projection from the CFG update, a repulsion operation that targets implicit distributional pull rather than explicit concept composition. Prior work on rare compositional generation includes R2F~\cite{park2025rare}, which uses LLM-generated concept substitutions to steer image diffusion models toward rare concepts at the representation level. DCR differs in two respects. First, DCR operates in the video domain, where temporal co-occurrence priors introduce failure modes, such as lighting-time misalignment (TEMP) and material-state conflict (MAT), that cannot be addressed by frame-independent image methods. Second, DCR suppresses default completion at the level of denoising trajectory dynamics via projection-based repulsion, rather than modifying the target representation itself. Vision-language models are known to struggle with negation~\cite{alhamoud2025vision, zhang2025negvqa, singh2024learn, xiao2026notjustwhatsthere}, and concept erasure methods suppress specific concepts via weight modification~\cite{gandikota2023erasing, gandikota2024unified}. The closest prior work, NEGATE~\cite{anonymous2026negate}, uses contrastive guidance to enforce \emph{explicit} negation constraints stated in the prompt; we adapt its attractor-directed metric analogues in Sec.~\ref{sec:quantitative_evaluation}. DCR instead targets \emph{implicit} default completion tendencies that arise without any explicit constraint, and replaces unconditional contrastive subtraction with projection-based repulsion on the guidance update. 

\subsection{Diffusion Models for Image and Video Generation} 
Denoising diffusion~\cite{ho2020denoising} and latent diffusion models~\cite{rombach2022high} underpin scalable text-to-image generation. This framework has been extended to video synthesis across diverse architectures~\cite{ho2022video, blattmann2023align, blattmann2023stable, ho2022imagen, singer2022make, bar2024lumiere, girdhar2311emu, lian2023llm, wang2023modelscope, wang2023gen, henschel2024streamingt2v, qiu2023freenoise, he2022latent, hong2022cogvideo, villegas2022phenaki, ge2023preserve, wang2024lavie}, including closed-source systems such as Sora~\cite{openai}, Veo~\cite{veo2024}, Veo 2~\cite{veo2}, and Movie Gen~\cite{polyak2025moviegencastmedia}. Recent open models including CogVideoX~\cite{yang2025cogvideox}, HunyuanVideo~\cite{wu2025hunyuanvideo, kong2024hunyuanvideo}, and Mochi~\cite{mochi} further advance quality and temporal consistency. Despite these advances, pretrained models inherit distributional biases that manifest as \emph{default completion} behavior, drifting toward high-frequency patterns even for valid rare compositions. Our work introduces a training-free mechanism that detects and counteracts such attractors at inference time.

\section{Method}

\label{sec:method}

\begin{figure*}[t]
    \centering
    \includegraphics[width=\linewidth]{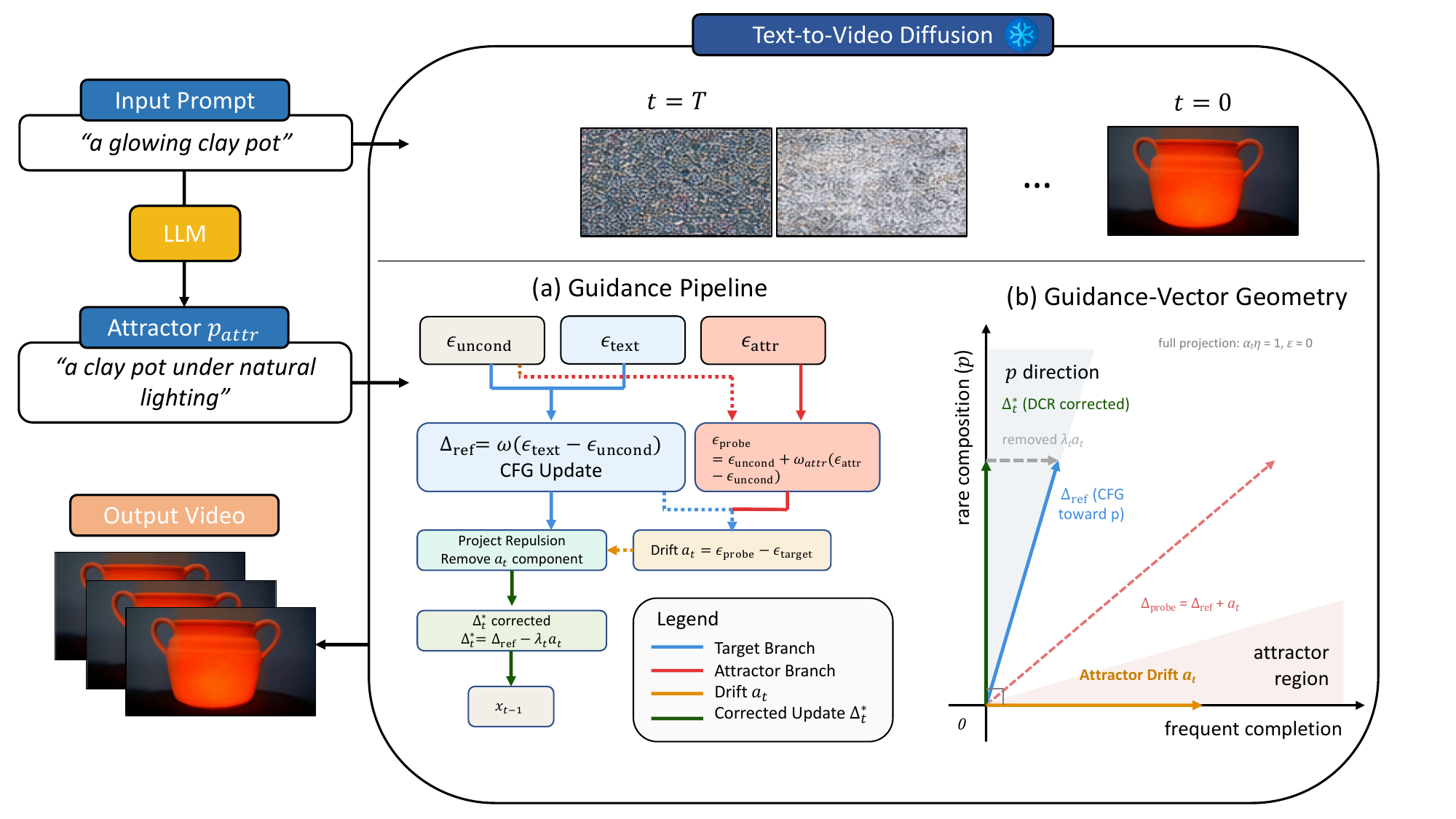}
    \caption{\textbf{Overview of Default Completion Repulsion (DCR).} Given a rare but plausible compositional prompt $p$ (e.g., ``a glowing clay pot''), an LLM generates an attractor prompt $p_{\text{attr}}$ representing the nearest frequent counterpart (e.g., ``a clay pot under natural lighting''). Both prompts enter the text-to-video diffusion process, where DCR is applied at each denoising step. \textbf{(a) Guidance Pipeline:} Three denoiser branches are computed: unconditional $\epsilon_{\text{uncond}}$, text-conditioned $\epsilon_{\text{text}}$, and attractor-conditioned $\epsilon_{\text{attr}}$. The standard CFG update is defined as $\Delta_{\text{ref}} = w(\epsilon_{\text{text}} - \epsilon_{\text{uncond}})$, and the CFG-guided prediction is $\epsilon_{\text{target}} = \epsilon_{\text{uncond}} + \Delta_{\text{ref}}$. An attractor probe $\epsilon_{\text{probe}} = \epsilon_{\text{uncond}} + w_{\text{attr}}(\epsilon_{\text{attr}} - \epsilon_{\text{uncond}})$ is constructed using a reduced guidance scale $w_{\text{attr}} < w$, and the counterfactual attractor drift is defined as $a_t = \epsilon_{\text{probe}} - \epsilon_{\text{target}}$. A projection-based repulsion mechanism removes the component of $\Delta_{\text{ref}}$ aligned with $a_t$, yielding the corrected update $\Delta_t^\star = \Delta_{\text{ref}} - \lambda_t a_t$. \textbf{(b) Guidance-Vector Geometry:} the panel is drawn for full projection ($\alpha_t \eta = 1$, $\varepsilon \approx 0$) under positive alignment. The probe increment $\Delta_{\text{probe}} = \Delta_{\text{ref}} + a_t$ leans toward the frequent completion, and the projection removes the drift-aligned component $\lambda_t a_t$ from $\Delta_{\text{ref}}$, so the corrected update $\Delta_t^\star$ is orthogonal to $a_t$ and keeps the rare-composition component. The shaded regions mark the rare-composition and frequent-completion directions and are schematic. The corrected trajectory produces compositionally faithful video generation without retraining the backbone diffusion model.}
    \label{fig:method_figure}
\end{figure*}

\subsection{Problem Setup}

\label{subsec:problem_setup}

Let $x_t \in \mathbb{R}^{C \times F \times H \times W}$ denote the latent variable at diffusion step $t$. 
Given a prompt $p$ describing a rare but plausible composition, our goal is to prevent the denoising trajectory from collapsing toward a more frequent completion favored by the pretrained model.

We denote the denoiser prediction by
\[
\epsilon_\theta(x_t, t, p),
\]
and the unconditional branch by
\[
\epsilon_{\text{uncond}} := \epsilon_\theta(x_t, t, \varnothing).
\]

Under standard classifier-free guidance (CFG), the guided update is
\begin{equation}
\Delta_{\text{ref}}
=
w \left(
\epsilon_{\text{text}} - \epsilon_{\text{uncond}}
\right),
\label{eq:delta_ref_main}
\end{equation}
where
\[
\epsilon_{\text{text}} := \epsilon_\theta(x_t,t,p),
\]
and $w>1$ is the guidance scale. The corresponding guided prediction is
\begin{equation}
\epsilon_{\text{target}}
=
\epsilon_{\text{uncond}} + \Delta_{\text{ref}}.
\label{eq:eps_target_main}
\end{equation}

Our method augments this formulation with an additional branch that exposes a competing completion tendency under relaxed semantics.

\subsection{Attractor Construction}
\label{subsec:attractor_branch}

We introduce an attractor prompt $p_{\text{attr}}$ that represents a relaxed version of the original prompt $p$. 
Specifically, $p_{\text{attr}}$ removes or weakens the rare compositional factor while preserving the surrounding semantic context.

For example, ``a snowy beach'' is relaxed into the more common completion ``a tropical beach.'' The construction proceeds in three steps, summarized together with the sampling loop in Algorithm~\ref{alg:dcr}. First, an LLM identifies the rare compositional factor of $p$, such as an unusual attribute, an atypical environment, or an unlikely object placement. Second, the LLM rewrites $p$ by replacing this factor with its most frequent counterpart while keeping all other scene elements unchanged; the fixed instruction template and decoding settings are given in Appendix~\ref{supp:attractor_prompts}. Third, the rewritten prompt is checked to confirm that it removes the rare factor and introduces no new elements. When several frequent counterparts are plausible, we keep the single completion produced at temperature 0 and verify during review that it names the dominant alternative. Projecting onto several attractor directions jointly is a natural extension and is left to future work.

We define:
\[
\epsilon_{\text{attr}} := \epsilon_\theta(x_t,t,p_{\text{attr}}).
\]
\paragraph{Interpretation.}
The attractor prompt induces a counterfactual completion trajectory that reflects how the model tends to complete the scene when the rare constraint is relaxed.

\begin{algorithm}[t]
\caption{DCR sampling with counterfactual attractor guidance}
\label{alg:dcr}
\begin{algorithmic}[1]
\REQUIRE prompt $p$, denoiser $\epsilon_\theta$, LLM rewriter $R$, scales $(w, w_{\text{attr}}, \eta)$, schedule $(r_s, r_e, \gamma)$, steps $T$
\STATE identify the rare compositional factor of $p$ with $R$
\STATE $p_{\text{attr}} \leftarrow R(p)$, replacing the rare factor with its most frequent counterpart (Appendix~\ref{supp:attractor_prompts})
\STATE verify that $p_{\text{attr}}$ removes the rare factor and adds no new elements
\FOR{$i = 0$ \TO $T-1$}
\STATE $\epsilon_{\text{uncond}} \leftarrow \epsilon_\theta(x_t, t, \varnothing)$;\ $\epsilon_{\text{text}} \leftarrow \epsilon_\theta(x_t, t, p)$;\ $\epsilon_{\text{attr}} \leftarrow \epsilon_\theta(x_t, t, p_{\text{attr}})$
\STATE $\Delta_{\text{ref}} \leftarrow w(\epsilon_{\text{text}} - \epsilon_{\text{uncond}})$;\ $\epsilon_{\text{target}} \leftarrow \epsilon_{\text{uncond}} + \Delta_{\text{ref}}$
\STATE $\epsilon_{\text{probe}} \leftarrow \epsilon_{\text{uncond}} + w_{\text{attr}}(\epsilon_{\text{attr}} - \epsilon_{\text{uncond}})$;\ $a_t \leftarrow \epsilon_{\text{probe}} - \epsilon_{\text{target}}$
\STATE $\alpha_t \leftarrow$ schedule value from $(r_s, r_e, \gamma)$
\STATE $\lambda_t \leftarrow \alpha_t \eta \, \max(\bar{a}_t^\top \bar{\Delta}_{\text{ref}}, 0) \, / \, (\lVert \bar{a}_t \rVert_2^2 + \varepsilon)$
\STATE $\Delta_t^\star \leftarrow \Delta_{\text{ref}} - \lambda_t a_t$;\ $x_{t-1} \leftarrow \mathrm{SchedulerStep}(\epsilon_{\text{uncond}} + \Delta_t^\star, t, x_t)$
\ENDFOR
\RETURN $x_0$
\end{algorithmic}
\end{algorithm}

\subsection{Counterfactual Probe Branch}
\label{subsec:probe_branch}

We construct a probe branch that applies CFG to the attractor prompt with a reduced guidance scale $w_{\text{attr}}$. Analogous to $\Delta_{\text{ref}}$ and $\epsilon_{\text{target}}$ in Eqs.~\eqref{eq:delta_ref_main} and~\eqref{eq:eps_target_main}, we define the probe increment $\Delta_{\text{probe}}$ and the probe prediction $\epsilon_{\text{probe}}$ as
\begin{equation}
\Delta_{\text{probe}}
=
w_{\text{attr}}
\left(
\epsilon_{\text{attr}} - \epsilon_{\text{uncond}}
\right),
\qquad
\epsilon_{\text{probe}}
=
\epsilon_{\text{uncond}} + \Delta_{\text{probe}},
\label{eq:eps_probe_main}
\end{equation}
where $0 \le w_{\text{attr}} < w$. The sensitivity of DCR to $w_{\text{attr}}$ is analyzed in Sec.~\ref{sec:hyperparam_sensitivity}.

\paragraph{Role of $w_{\text{attr}}$.}
The parameter $w_{\text{attr}}$ controls the influence strength of the attractor branch independently of the semantic content of $p_{\text{attr}}$. 
This enables fine-grained adjustment of the attractor trajectory magnitude without modifying the attractor prompt itself, thereby avoiding additional semantic distortion. This branch is not used for final generation, but instead serves as a probe revealing a competing direction favored by the model.

\subsection{Counterfactual Attractor Drift}
\label{subsec:drift_main}
We define the attractor drift as:
\begin{equation}
a_t
=
\epsilon_{\text{probe}} - \epsilon_{\text{target}}.
\label{eq:drift_main}
\end{equation}
Since both predictions share $\epsilon_{\text{uncond}}$, the drift is the difference of the two guidance increments,
\begin{equation}
a_t
=
\Delta_{\text{probe}} - \Delta_{\text{ref}}
=
w_{\text{attr}}
\left(
\epsilon_{\text{attr}} - \epsilon_{\text{uncond}}
\right)
-
w
\left(
\epsilon_{\text{text}} - \epsilon_{\text{uncond}}
\right),
\label{eq:drift_expanded_main}
\end{equation}
or equivalently $\Delta_{\text{probe}} = \Delta_{\text{ref}} + a_t$, the relation shown in Fig.~\ref{fig:method_figure}(b).
Because $p_{\text{attr}} \neq p$, we have $\epsilon_{\text{attr}} \neq \epsilon_{\text{text}}$ whenever the denoiser is sensitive to the relaxed compositional factor, so the two direction vectors $\epsilon_{\text{attr}} - \epsilon_{\text{uncond}}$ and $\epsilon_{\text{text}} - \epsilon_{\text{uncond}}$ are in general not collinear. Equivalently, $a_t$ cannot in general be written as $c\, (\epsilon_{\text{text}} - \epsilon_{\text{uncond}})$ for any scalar $c$. This distinguishes our update from any guidance-scale rescaling $\Delta_{\text{ref}} \mapsto c\,\Delta_{\text{ref}}$, which only modulates \emph{magnitude} along a fixed direction; the projection in Sec.~\ref{subsec:projection_main} therefore removes a distinct completion tendency rather than attenuating the guidance strength.

\subsection{Scheduled Repulsion}
\label{subsec:schedule_main}

The repulsion is weighted by a per-step coefficient $\alpha_t \in [0,1]$ that enters the projection in Sec.~\ref{subsec:projection_main}. Let $i = 0, \dots, T-1$ index the sampling steps, starting from the noisiest one, and let $\pi_t = i/(T-1) \in [0,1]$ denote the normalized progress of sampling. Repulsion is active only within an interval $[r_s, r_e] \subseteq [0,1]$, and the weight is
\begin{equation}
\alpha_t
=
\begin{cases}
\tilde{\pi}_t^{\,\gamma}, & r_s \le \pi_t \le r_e,\\
0, & \text{otherwise},
\end{cases}
\qquad
\tilde{\pi}_t = \frac{\pi_t - r_s}{r_e - r_s}.
\label{eq:schedule_main}
\end{equation}
Here $\tilde{\pi}_t$ is the relative position within the active interval, rising linearly from 0 at $r_s$ to 1 at $r_e$. The exponent $\gamma > 0$ controls how quickly the weight rises. With $\gamma = 1$ the weight grows linearly, and with $\gamma > 1$ it stays small at the start of the interval and grows toward its end, so the strongest repulsion is applied in the later part of the interval rather than in the early steps where the global scene layout forms. We use $\gamma = 2$ in all experiments. Setting $r_s > 0$ leaves the earliest steps unmodified for the same reason, and the ablation without the schedule ($\alpha_t = 1$ at every step) in Sec.~\ref{sec:ablation} shows that repulsion applied from the start disrupts the scene structure. The sensitivity of DCR to the interval $(r_s, r_e)$ is analyzed in Sec.~\ref{sec:hyperparam_sensitivity}.

\subsection{Projection-Based Repulsion}
\label{subsec:projection_main}

We flatten:
\begin{equation}
\bar{a}_t = \mathrm{vec}(a_t),
\quad
\bar{\Delta}_{\text{ref}} = \mathrm{vec}(\Delta_{\text{ref}}).
\end{equation}

The repulsion coefficient is
\begin{equation}
\lambda_t
=
\alpha_t \eta \,
\frac{\max\!\left(\bar{a}_t^\top \bar{\Delta}_{\text{ref}},\, 0\right)}{\lVert \bar{a}_t \rVert_2^2 + \varepsilon},
\label{eq:lambda_main}
\end{equation}
where $\eta > 0$ is the repulsion strength, $\varepsilon$ is a small constant for numerical stability, and the $\max(\cdot, 0)$ gate applies repulsion only when the CFG update is positively aligned with the drift. The sensitivity of DCR to $\eta$ is analyzed in Sec.~\ref{sec:hyperparam_sensitivity}.

The corrected update and final denoising step are:
\begin{equation}
\Delta_t^\star
=
\Delta_{\text{ref}} - \lambda_t a_t,
\end{equation}
\begin{equation}
\epsilon_t^\star
=
\epsilon_{\text{uncond}} + \Delta_t^\star,
\end{equation}
\begin{equation}
x_{t-1}
=
\mathrm{SchedulerStep}(\epsilon_t^\star, t, x_t).
\end{equation}

\paragraph{Interpretation.}
This correction removes only the component of the CFG update that aligns with the attractor direction, preventing collapse toward frequent completions while preserving other semantic components. When $\alpha_t \eta = 1$ and $\varepsilon$ is negligible, $\lambda_t$ reduces to $\bar{a}_t^\top \bar{\Delta}_{\text{ref}} / \lVert \bar{a}_t \rVert_2^2$ for positive alignment, so the entire component of $\Delta_{\text{ref}}$ along $a_t$ is removed and $\Delta_t^\star$ becomes orthogonal to $a_t$; smaller values of $\alpha_t \eta$ remove only a fraction of it. Figure~\ref{fig:method_figure}(b) is drawn for this full-projection case.

\section{Experiments and Results}
\subsection{Implementation Details}
Our method is implemented as a guidance-augmented inference procedure within the standard diffusion sampling loop, without modifying or retraining the backbone model. During classifier-free guidance, we compute three prediction branches: unconditional, text-conditioned, and an attractor-conditioned branch derived from the transformed prompt $p_{\text{attr}}$. Instead of directly subtracting the attractor prediction, we interpret default completion bias as a soft directional constraint on the guidance update and perform a projection-based correction that removes the component of the CFG direction aligned with the attractor drift. The repulsion strength follows a polynomial schedule $\alpha_t = \tilde{\pi}_t^\gamma$ with $\gamma = 2.0$, gradually increasing suppression of the attractor direction over the active interval. Only positive alignment between the attractor drift and the CFG update is penalized, and numerical stabilization uses $\varepsilon = 10^{-8}$. All backbone-specific parameters follow default configurations~\cite{mochi}, and no additional learnable parameters are introduced. We use Mochi as the primary backbone since it exposes a clean three-branch CFG interface for controlled ablation; the formulation is backbone-invariant by construction, and the targeted failure mode is shared across HunyuanVideo~\cite{kong2024hunyuanvideo} and CogVideoX~\cite{yang2025cogvideox} (Sec.~\ref{sec:qualitative}, Appendix~\ref{supp:backbone}).

\subsection{Benchmarking Datasets}
Standard caption-alignment benchmarks such as MS-COCO, CC, and WebVid neither control compositional rarity nor expose the drift toward frequent completions that this work targets, so we construct a controlled evaluation suite of eight categories (ENV, TEMP, OBJ, ATTR, SCALE, CTX, MAT, DENS), each isolating one compositional factor with 50 prompts, for 400 prompts in total, all physically plausible yet distributionally rare. The quantity of interest is generative rather than representational, the probability that a sample drawn from $p_\theta(\cdot \mid p)$ realizes every semantic factor of $p$ simultaneously, for example both \emph{snowy} and \emph{beach}. This probability has no closed form, so the judged metrics of Sec.~\ref{sec:quantitative_evaluation} estimate it by sampling, CVR as the frequency of attractor-directed violations and CCS as a graded relaxation of the underlying indicator. The dataset is used exclusively for evaluation; no training or fine-tuning is performed. \textbf{The formal objective, construction details, category definitions, and examples are provided in Appendix~\ref{supp:detailed_dataset}.}

\begin{figure}[t]
    \centering
    \includegraphics[width=0.875\linewidth]{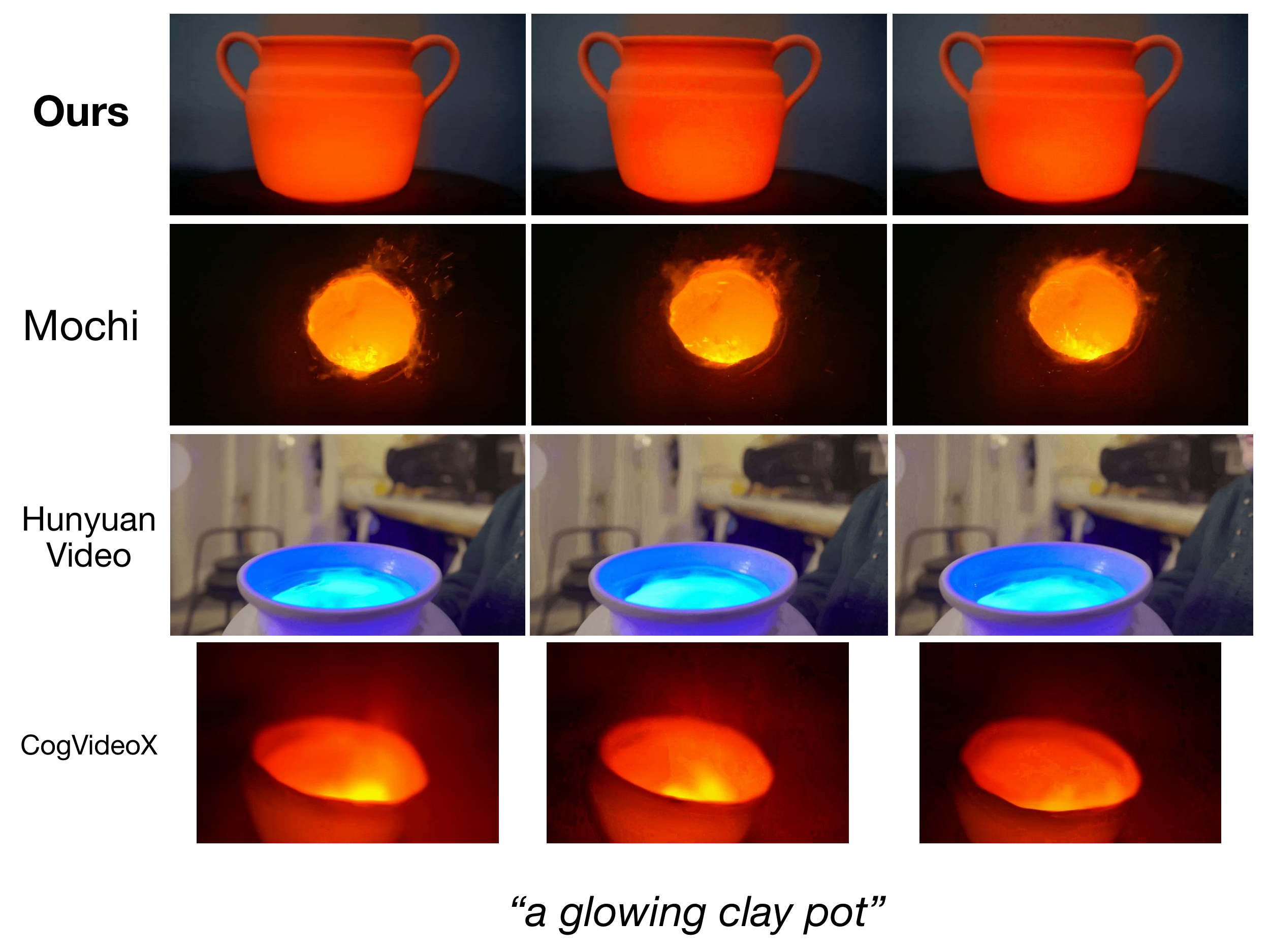}
    \caption{\textbf{Representative qualitative comparison (ATTR).}
    We compare our method with state-of-the-art diffusion baselines (Mochi, HunyuanVideo, CogVideoX) on Attribute Rebinding (ATTR): ``a glowing clay pot.'' 
    While baselines often alter object identity or material properties (e.g., generating molten objects or unrelated glowing containers), our method correctly binds the luminous attribute to the clay pot, preserving both object identity and material consistency.}
    \label{fig:qualitative_main}
\end{figure}

\subsection{Qualitative Results} 
\label{sec:qualitative} 

Figure~\ref{fig:qualitative_main} shows a representative comparison on ATTR. Baselines share one failure mode, collapse toward the statistically dominant completion. A snowy beach is rendered as a tropical one (ENV), a rainbow at night becomes a daytime rainbow or a dark frame (TEMP), objects and scenes revert to their most common contexts (OBJ, CTX), and scale or material constraints are simplified away (SCALE, MAT). DCR applies projection-based repulsion along the counterfactual attractor direction and prevents this collapse while preserving the remaining semantics; in Figure~\ref{fig:qualitative_main} it binds the luminous attribute to the clay pot without altering object identity or material consistency. \textbf{Comparisons across all eight categories are provided in Appendix~\ref{sec:detailed_qualitative}.}

\begin{table}[t]
\begin{small}
    \centering
    \resizebox{0.5\textwidth}{!}{%
    \begin{tabular}{lccccc}
        \toprule
        \textbf{Method} & \textbf{CLIPScore} $\uparrow$ & \textbf{CLIP-attr} $\downarrow$ & \textbf{BLIP} $\uparrow$ & \textbf{CCS} $\uparrow$ & \textbf{CVR} $\downarrow$ \\
        \midrule
        Mochi            & 0.2986 & 0.2718 & 0.7807 & 3.8375 & 0.4725 \\
        HunyuanVideo     & 0.3067 & 0.2725 & 0.7819 & 3.9750 & 0.3750 \\
        CogVideoX        & 0.2858 & 0.2644 & 0.7290 & 3.1925 & 0.5175 \\
        \midrule
        \textbf{Ours}    & \textbf{0.3174} & \textbf{0.2558} & \textbf{0.8075} & \textbf{4.1300} & \textbf{0.3100} \\
        \midrule
        Negative Prompt  & 0.2736 & 0.2610 & 0.7065 & 3.0375 & 0.4375 \\
        Ours w/o Attractor Prompt$^\dagger$ & 0.2986 & 0.2718 & 0.7807 & 3.8375 & 0.4725 \\
        Ours w/o Repulsion$^\dagger$        & 0.2986 & 0.2718 & 0.7807 & 3.8375 & 0.4725 \\
        Ours w/o Schedule         & 0.3115 & 0.2740 & 0.7819 & 3.9675 & 0.3550 \\
        \bottomrule
    \end{tabular}}
    \caption{\textbf{Quantitative evaluation results.} All metrics are averaged over the full evaluation set. Mochi, HunyuanVideo, and CogVideoX denote standard CFG baselines. Negative Prompt denotes the baseline where $p_{\text{attr}}$ is provided as a negative prompt within standard CFG on Mochi. Ablation variants are based on Mochi. CCS denotes the Compositional Compliance Score and CVR denotes the Compositional Violation Rate, both obtained from a GPT-4o vision-language judge. $^\dagger$By construction these variants leave the final denoising update identical to standard CFG ($\lambda_t = 0$; Sec.~\ref{subsec:projection_main}), so their rows coincide with the Mochi baseline; they are listed to make the reduction explicit.}
    \label{tab:dcr_main}
\end{small}
\end{table}

\begin{figure*}[t]
    \centering
    \includegraphics[width=0.95\linewidth]
    {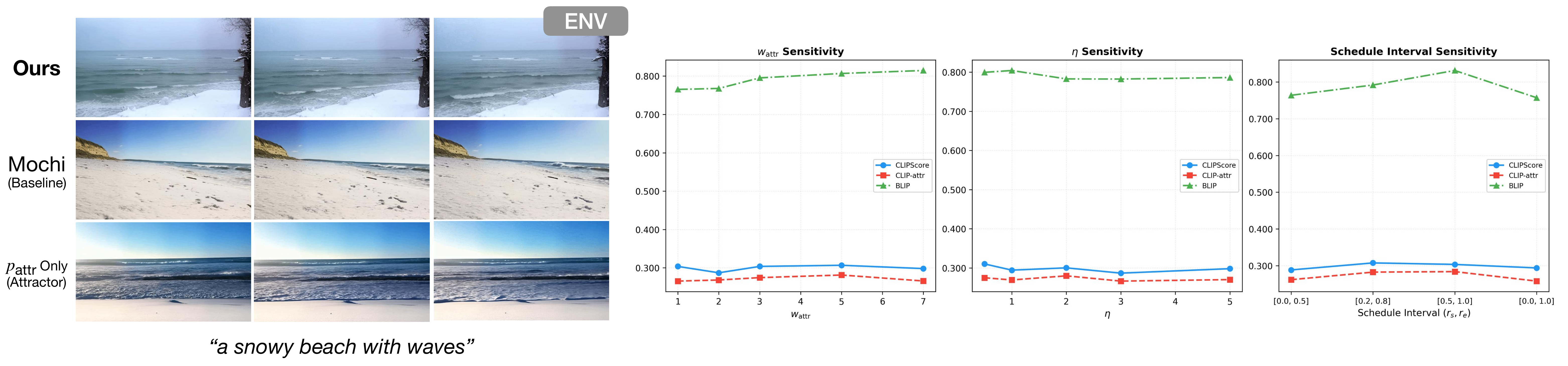}
    \caption{\textbf{Attractor trajectory visualization and hyperparameter sensitivity.} \textbf{(Left)} Comparison of Ours (DCR), Mochi (baseline), and $p_{\text{attr}}$ only (attractor) for \emph{a snowy beach with waves} (ENV). The $p_{\text{attr}}$-only output closely resembles the Mochi baseline, confirming that $p_{\text{attr}}$ captures the model's default completion tendency. \textbf{(Right)} Sensitivity of CLIPScore, CLIP-attr, and BLIP to $w_{\text{attr}}$, $\eta$, and schedule interval $(r_s, r_e)$. All three metrics remain stable across a wide range of values, demonstrating that DCR does not require careful hyperparameter tuning.}
    \label{fig:traj_sensitivity}
\end{figure*}

\subsection{Quantitative Evaluation} 
\label{sec:quantitative_evaluation} 
We report three embedding-level metrics, CLIPScore against the full prompt $p$ for global alignment, CLIP similarity to $p_{\text{attr}}$ (CLIP-attr, lower is better) for attractor suppression, and BLIP caption similarity for text-level fidelity, together with two reasoning-level metrics from a GPT-4o judge, the \emph{Compositional Compliance Score} (CCS, 1--5) and the \emph{Compositional Violation Rate} (CVR), the share of videos that collapse to the frequent counterpart. The judge follows a fixed rubric at temperature 0, and GPT-based compliance scoring has been shown to rank methods consistently with human judgment on closely related tasks~\cite{anonymous2026negate}. \textbf{Appendix~\ref{sec:detailed_quantitative_evaluation} gives full metric definitions, the judge configuration, and the aggregation protocol.} As shown in Table~\ref{tab:dcr_main}, our method is best on every metric, raising CLIPScore to 0.3174 and BLIP to 0.8075 while lowering CLIP-attr to 0.2558, and improving the reasoning-level metrics to a CCS of 4.13 and a CVR of 0.31 against 0.4725 for Mochi.

\begin{table}[t]
\begin{small}
    \centering
    \resizebox{0.5\textwidth}{!}{%
    \begin{tabular}{lcccccc}
        \toprule
        \textbf{Method} & \textbf{SC} $\uparrow$ & \textbf{BC} $\uparrow$ & \textbf{TF} $\uparrow$ & \textbf{DD} & \textbf{AQ} $\uparrow$ & \textbf{IQ} $\uparrow$ \\
        \midrule
        Mochi            & 0.9749 & 0.9581 & 0.9858 & 0.2375 & 0.5542 & 0.5777 \\
        HunyuanVideo     & 0.9890 & 0.9764 & 0.9931 & 0.1400 & 0.5863 & 0.6328 \\
        CogVideoX        & 0.9731 & 0.9685 & 0.9793 & 0.5375 & 0.4850 & 0.4902 \\
        \midrule
        \textbf{Ours}    & 0.9740 & 0.9583 & 0.9851 & 0.2600 & 0.5566 & 0.5809 \\
        \midrule
        Negative Prompt  & 0.9555 & 0.9452 & 0.9820 & 0.2400 & 0.4809 & 0.5686 \\
        Ours w/o Attractor Prompt$^\dagger$ & 0.9772 & 0.9612 & 0.9868 & 0.1950 & 0.5595 & 0.5893 \\
        Ours w/o Repulsion$^\dagger$        & 0.9763 & 0.9619 & 0.9865 & 0.1775 & 0.5578 & 0.5925 \\
        Ours w/o Schedule         & 0.9739 & 0.9588 & 0.9858 & 0.2050 & 0.5534 & 0.5882 \\
        \bottomrule
    \end{tabular}}
    \caption{\textbf{Reference-free perceptual quality (VBench).} Six VBench quality dimensions evaluated on the same generated videos in custom-input mode. SC, subject consistency; BC, background consistency; TF, temporal flickering; DD, dynamic degree; AQ, aesthetic quality; IQ, imaging quality. DD measures the share of videos with large motion and carries no preferred direction. $^\dagger$These variants reduce analytically to standard CFG, so their deviation from the Mochi row reflects run-to-run sampling variance.}
    \label{tab:dcr_vbench}
\end{small}
\end{table}

Compositional gains would mean little if they came at the cost of perceptual quality, so Table~\ref{tab:dcr_vbench} additionally reports reference-free quality scores from VBench~\cite{huang2024vbench} on the same generated videos, covering six quality dimensions evaluated in custom-input mode. Our method matches the Mochi backbone on every dimension, with gaps of at most $0.001$ on subject consistency, background consistency, and temporal flickering and slightly higher scores on dynamic degree, aesthetic quality, and imaging quality. The two ablation variants that reduce analytically to standard CFG provide a direct estimate of run-to-run sampling variance, and on every dimension the gap between Ours and Mochi is smaller than the variation between those rows and the Mochi row, so DCR leaves perceptual quality unchanged within sampling noise. The contrast with the Negative Prompt baseline is sharp. Placing $p_{\text{attr}}$ in the negative slot of CFG lowers subject consistency to $0.9555$, background consistency to $0.9452$, and aesthetic quality to $0.4809$, confirming that unconditional negation pays a quality price that the gated projection avoids. HunyuanVideo attains the highest raw quality scores, consistent with its stronger backbone, yet Table~\ref{tab:dcr_main} shows it remains well behind on compositional fidelity, which is the failure mode this work targets.

\begin{figure}[htb!]
\centering
\includegraphics[width=0.875\linewidth]{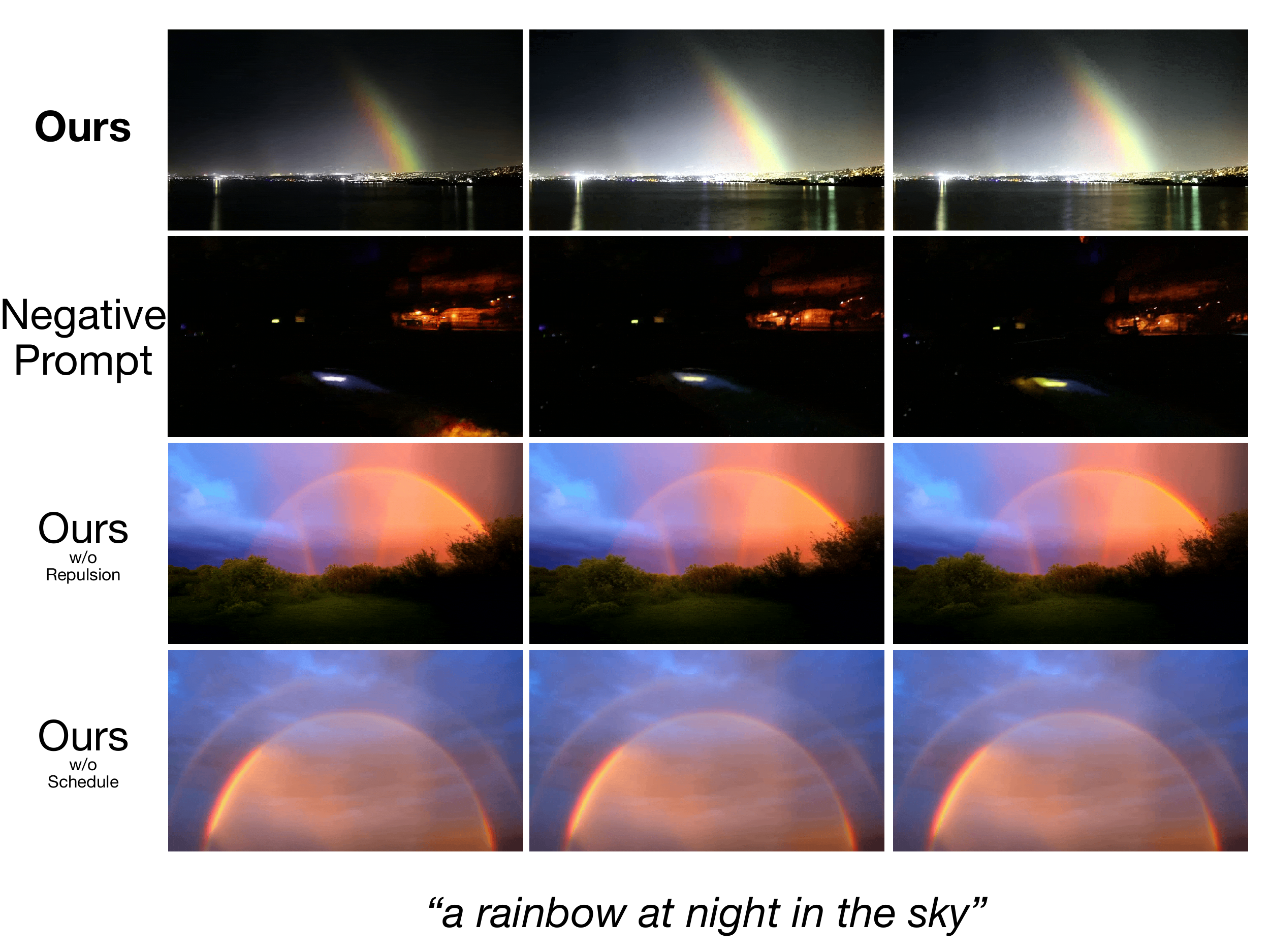}
\caption{\textbf{Ablation study on rare compositional generation.} We compare our full model against three ablation variants on the prompt \textit{``a rainbow at night in the sky.''} \textbf{Negative Prompt} replaces the attractor prompt $p_\text{attr}$ into the negative slot of standard CFG, producing dark, near-black frames that entirely fail to render the rainbow, confirming that naive negation via CFG cannot encode compositional rarity. \textbf{w/o Repulsion} retains the attractor probe but sets $\lambda_t = 0$, so the final update is identical to standard CFG by construction; the output accordingly reverts to the model's default daylight completion, showing the behavior the projection must overcome. \textbf{w/o Schedule} applies repulsion uniformly ($\alpha_t = 1$) across all denoising steps; early-step interference disrupts global structure formation, yielding an oversaturated close-up arc that loses the intended nocturnal scene context. Our full model correctly synthesizes a rainbow against a dark night sky while maintaining temporal coherence across frames.}
\label{results:ablation}
\end{figure}

\subsection{Ablation Study}
\label{sec:ablation}
We ablate the attractor prompt, the repulsion update, the schedule, and a negative prompt baseline (Table~\ref{tab:dcr_main}, Figure~\ref{results:ablation}). Negative prompting degrades most severely, dropping CLIPScore to $0.2736$ and BLIP to $0.7065$ with near-black frames, since unconditional CFG repulsion suppresses the shared content along with the unwanted attribute. The w/o Repulsion and w/o Attractor Prompt variants reduce analytically to standard CFG, the former by setting $\lambda_t = 0$ and the latter because the drift becomes anti-parallel to $\Delta_{\text{ref}}$ and the gate closes (Sec.~\ref{subsec:projection_main}), so their rows coincide with the Mochi baseline and the gap to the full model (CCS $3.8375 \to 4.1300$, CVR $0.4725 \to 0.3100$) measures the net contribution of the attractor-guided projection. Removing only the schedule raises CVR to $0.3550$, with premature early-step repulsion corrupting global structure (Figure~\ref{results:ablation}). \textbf{More detailed analysis is provided in Appendix~\ref{sec:detailed_ablation}.}

\subsection{Attractor Trajectory Visualization}
Figure~\ref{fig:traj_sensitivity} (left) compares videos generated by Ours (DCR), Mochi (standard CFG baseline), and $p_{\text{attr}}$ only for \emph{a snowy beach with waves}. The $p_{\text{attr}}$-only output closely resembles the Mochi baseline, both producing a snow-free sunny beach, confirming that $p_{\text{attr}}$ accurately captures the model's default completion tendency. This validates the core motivation of DCR: by explicitly modeling the attractor direction, the method can apply targeted repulsion to prevent compositional collapse. Extended comparisons across additional categories are provided in Appendix~\ref{supp:attractor_visualization}. 

\subsection{Hyperparameter Sensitivity}
\label{sec:hyperparam_sensitivity} 
Figure~\ref{fig:traj_sensitivity} (right) reports the sensitivity of our method to three key hyperparameters: the attractor guidance scale $w_{\text{attr}}$, the repulsion strength $\eta$, and the schedule interval $(r_s, r_e)$. Across all three sweeps, CLIPScore and CLIP-attr remain stable within a variation of approximately 0.01, indicating that DCR does not require careful hyperparameter tuning to achieve effective attractor suppression. BLIP shows a mild monotonic trend with respect to $w_{\text{attr}}$ and $\eta$, but remains competitive across all evaluated settings. For the schedule interval, the [0.2, 0.8] and [0.5, 1.0] configurations both perform well on embedding-based metrics, while our default [0.2, 0.8] provides a favorable balance between early structural formation and late-stage attractor suppression. These results confirm that DCR is robust to hyperparameter variation across a wide range of values, and that our default configuration ($w_{\text{attr}} = 3.0$, $\eta = 1.0$, $(r_s, r_e) = [0.2, 0.8]$) reliably achieves strong compositional fidelity without exhaustive tuning.

\subsection{Guidance Geometry Diagnostics}
\label{sec:guidance_diag}
The projection is derived from a geometric claim, that the attractor-guided and text-guided updates disagree in a measurable direction which the gate isolates. To verify this claim directly we log the guidance geometry at every denoising step while regenerating two groups of prompts from the main evaluation, 40 prompts the judge scored as successes (five per category, sampled at random) and the 124 prompts it scored as violations. Let $\varphi_t$ denote the angle between the raw guidance directions $\epsilon_{\text{attr}} - \epsilon_{\text{uncond}}$ and $\epsilon_{\text{text}} - \epsilon_{\text{uncond}}$, and $\rho_t$ their norm ratio; the gate condition $s_t > 0$ is equivalent to $\rho_t \cos\varphi_t > w / w_{\text{attr}}$, so these two quantities fully determine when the projection acts. 

\begin{figure}[t]
\centering
\includegraphics[width=\linewidth]{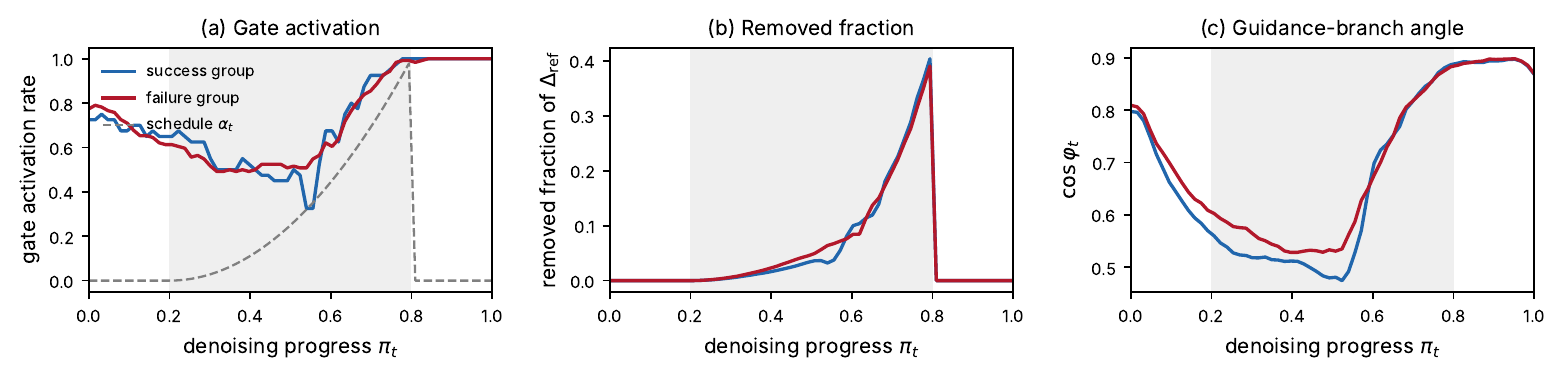}
\caption{\textbf{Guidance geometry across denoising steps.} Step-wise means over the regenerated success and failure groups; the shaded band marks the scheduling window $[r_s, r_e] = [0.2, 0.8]$. \textbf{(a)} Gate activation rate, with the schedule $\alpha_t$ dashed. \textbf{(b)} Fraction of $\Delta_{\text{ref}}$ removed by the projection. \textbf{(c)} Alignment $\cos\varphi_t$ between the attractor- and text-guidance directions. The branches disagree most in mid-generation, where composition is decided, and the removal concentrates there through the schedule.}
\label{fig:diag_steps}
\end{figure}

Figure~\ref{fig:diag_steps} shows the step-wise behavior averaged over all clips. The two guidance branches are far from collinear throughout, with $\cos\varphi_t$ dipping to about $0.5$ in the middle of the scheduling window, precisely the phase in which scene composition is decided, and re-aligning above $0.9$ only in the final steps. Steps with $\lvert\cos\varphi_t\rvert > 0.99$ account for at most $0.5\%$ of the window. Within the window the gate is open on $63.9\%$ of steps for both groups, and the removed fraction of $\Delta_{\text{ref}}$ follows the schedule, negligible early, rising to about $40\%$ just before the window closes, and averaging $9\%$ over the window. The projection is therefore selective twice over, in time through the schedule and in direction through the gate, rather than a uniform attenuation of the guidance signal. Success and failure groups show nearly identical statistics ($\cos\varphi$ of $0.617$ against $0.642$, identical gate rates, effective removal of $8.8\%$ against $9.1\%$), which rules out one candidate explanation of the residual failures, that the mechanism simply fails to engage on hard prompts. The failure analysis in Appendix~\ref{supp:failure_analysis} locates the residual failures instead in where the correction points and how much it removes; per-category diagnostics are given in Appendix~\ref{supp:diagnostics}.

\section{Conclusions} 
We formulated \emph{default completion bias}, the tendency of pretrained diffusion models to drift from a rare requested composition toward its high-frequency counterpart, as an explicit, correctable component of the guidance update. A counterfactual attractor prompt, obtained by relaxing the rare compositional factor, exposes the model's preferred completion; the discrepancy between the target- and attractor-guided updates defines a drift direction; and a scheduled projection removes the component of the CFG update aligned with that drift, leaving the remaining semantics untouched. The method requires no retraining or architectural change, and on our eight-category benchmark it improves compositional fidelity over standard CFG and negative-prompt baselines while preserving visual quality, with ablations attributing the gains separately to the attractor construction, the projection, and the schedule. Because the correction is defined on the classifier-free guidance update alone, the formulation carries over to other backbones, although our controlled quantitative study is confined to Mochi. Extending the attractor construction beyond rare co-occurrence to negation, quantification, and relational constraints is a natural next step. \textbf{Limitations and Future Work. }DCR reduces default completion but does not guarantee that it is avoided. The judge still reports a compositional violation for 31\% of the videos generated with DCR (CVR of 0.31, Table~\ref{tab:dcr_main}), compared with 47\% for Mochi, because the projection removes only the component of the CFG update aligned with the estimated drift. A manual analysis of regenerated failure cases (Appendix~\ref{supp:failure_analysis}) finds no single dominant cause. In 29\% of the analyzed failures the projection removes the shared scene content together with the drift, in 25\% the correction fires but removes too little to displace the default completion, in 24\% the intervention is effectively inactive, and in 12\% the clip collapses to a completion that the attractor prompt does not capture. Our quantitative study is also confined to Mochi, since transferring DCR to other backbones requires recalibrating $(w_{\text{attr}}, \eta, r_s, r_e)$. Finally, CCS and CVR come from the same model family that generates the attractor prompts, and independent human validation remains future work. \textbf{Appendix~\ref{supp:detailed_limitation} discusses these limitations in detail.}

{
    \small
    \bibliographystyle{ieeenat_fullname}
    \bibliography{main}
}

\newpage
\onecolumn
\appendix
\section{Detailed Limitations and Future Works}
\label{supp:detailed_limitation}
DCR reduces default completion but does not guarantee that it is avoided. With DCR, the judge still reports a compositional violation for 31\% of the generated videos (CVR of 0.31, Table~\ref{tab:dcr_main}), compared with 47\% for Mochi, and the remaining rate varies by category, from 18\% in MAT to 40\% in TEMP and ATTR (Table~\ref{tab:dcr_category}). The projection only removes the component of the CFG update that is aligned with the estimated drift, so it cannot help when the attractor prompt misses the actual default completion, when the gate is inactive, or when the rare factor is weakly represented in the conditional prediction itself.

Our current formulation approximates default completion bias through a counterfactual attractor prompt $p_{\text{attr}}$, which requires identifying a semantically meaningful frequent counterpart for each target composition. While we automate this via LLM-based prompt relaxation, cases where the attractor is ambiguous or difficult to define precisely may limit suppression effectiveness. Additionally, the projection-based repulsion operates at the level of global latent directions and may not fully capture spatially localized or temporally varying compositional constraints. DCR is validated on Mochi as a single-backbone setting that allows clean attribution of the attractor probe, the projection-based repulsion, and the polynomial schedule; the formulation is invariant to backbone architecture by construction, and Appendix~\ref{supp:backbone} discusses this scope together with the empirical observation, drawn from our comparisons against HunyuanVideo and CogVideoX, that compositional collapse is shared across the current generation of text-to-video diffusion models. Our judged metrics (CCS and CVR) are produced by the same model family that generates the attractor prompts; the protocol fixes the rubric and decoding temperature, but independent human validation on DCR-Bench remains future work. In addition, preliminary exploration on other backbones indicates that per-backbone calibration of $(w_{\text{attr}}, \eta, r_s, r_e)$ is more involved than matching guidance scales alone, so we confine quantitative claims to Mochi. Future work may extend this formulation to broader classes of distributional constraints beyond rare co-occurrence, including negation, quantification, and relational compositional structure. Integrating learned attractor estimation, rather than prompt-based relaxation, with neural trajectory control could further reduce dependence on explicit attractor specification.

\section{Ethics Statement}
\label{supplementary:ethics}

\begin{tcolorbox}[breakable, title=Ethics Statement] 
This work introduces Default Completion Repulsion (DCR), a training-free inference-time guidance method for text-to-video diffusion models. We address the following ethical considerations. \textbf{Intended Use.} DCR is designed to improve compositional fidelity in generative video models by suppressing distributional biases that cause models to ignore rare but valid semantic combinations. The method is intended for research purposes and for applications where faithful prompt adherence is required, such as creative content generation, simulation, and visual prototyping. \textbf{Potential Misuse.} As with all generative video methods, DCR could in principle be applied to produce misleading, harmful, or deceptive content. We note that DCR does not introduce new generative capabilities beyond those of the underlying backbone model; it modifies guidance at inference time to better reflect the intended prompt. The risks associated with DCR are therefore no greater than those of the backbone diffusion models on which it operates. \textbf{Benchmark and Data.} The DCR-Bench evaluation suite is constructed solely for research evaluation purposes. All prompts are manually designed to be physically plausible and free of harmful, offensive, or privacy-violating content. No personally identifiable information is collected or used. The benchmark is released for reproducibility and to facilitate future research on compositional generation. \textbf{Computational Resources.} DCR operates entirely at inference time and requires no additional training. The primary computational overhead relative to standard CFG is a single additional denoiser forward pass per step for the attractor branch. All experiments were conducted on standard research GPU hardware. \textbf{Broader Impact.} By improving the ability of generative models to faithfully realize rare but valid compositions, DCR may contribute to more reliable and controllable AI-generated content. We hope this work encourages further research into distributional bias in generative models and inference-time methods for controllable generation. \textbf{Artifact Licensing and Compatibility.} The pretrained video diffusion backbones evaluated throughout this study are obtained from their respective official releases and used in accordance with the licenses declared by their authors. Mochi~1~\cite{mochi} and CogVideoX~\cite{yang2025cogvideox} are each released under Apache~2.0, granting unrestricted research, modification, and redistribution rights. HunyuanVideo~\cite{kong2024hunyuanvideo} is governed by the Tencent Hunyuan Community License, under which academic research is permitted within the licensed territory; all reported experiments fall entirely within this scope. The two text-image evaluation models invoked by our metrics, CLIP and BLIP, are made available under MIT and BSD-3-Clause respectively, both of which authorize the inference-only research usage applied here. Throughout this work, no model weights are retrained, fine-tuned, modified, or hosted for redistribution; all backbones are accessed strictly as frozen inference endpoints, which keeps our usage compatible with the original research-oriented terms of every artifact employed. For the resource produced by this paper, the DCR-Bench prompt suite, we adopt the CC~BY-NC~4.0 license. This choice mirrors the non-commercial spirit of the most restrictive upstream license among the evaluated backbones and is intended to keep the benchmark aligned with academic dissemination practices in the compositional generation literature.
\end{tcolorbox}

\section{Detailed Benchmarking Dataset}
\label{supp:detailed_dataset}

Unlike prior works that evaluate compositional generalization through retrieval or captioning accuracy, our objective is to assess \emph{compositional fidelity in generative diffusion trajectories}. This changes what the evaluation dataset must contain.

\paragraph{Why Standard Datasets Are Insufficient.}
Conventional vision-language benchmarks such as MS-COCO, CC, or WebVid are designed for descriptive captioning and semantic alignment. Rare compositional prompts are underrepresented, unstructured, and not systematically controlled. More importantly, these datasets evaluate representation-level similarity rather than the distributional behavior of generated samples under compositional constraints.

In representation learning settings, evaluation typically takes the form:
\[
S(I, T_{\text{target}}) > S(I, T_{\text{distractor}}),
\]
which measures embedding alignment. In contrast, our problem is generative and distributional:
\begin{equation}
\mathbb{P}_{x_0 \sim p_\theta(\cdot \mid p)} 
\left[
\mathcal{C}(x_0; p) = 1
\right],
\label{eq:dataset_objective}
\end{equation}
where $\mathcal{C}(x_0; p)$ denotes whether the generated sample satisfies the target composition specified by prompt $p$.

Standard datasets do not provide structured prompts to estimate Eq.~\eqref{eq:dataset_objective}, nor do they expose systematic deviations toward frequent alternatives. Therefore, we construct a controlled evaluation suite tailored to compositional fidelity.

\paragraph{Dataset Scale.}
We construct eight categories, each containing 50 prompts, resulting in a total of 400 evaluation samples. Each prompt is designed to isolate a specific compositional failure mode. The dataset is used solely for evaluation; no training or fine-tuning is performed.

\paragraph{Compositional Constraint Formulation.}
Each prompt $p$ implicitly defines a compositional constraint over generated samples:
\begin{equation}
\mathcal{C}(x_0; p) = 
\prod_{k=1}^{K} 
\mathbf{1}\left[g_k(x_0) \in \mathcal{S}_k(p)\right],
\end{equation}
where $g_k(x_0)$ denotes a semantic factor (e.g., environment, object, scale), and $\mathcal{S}_k(p)$ defines the allowable set specified by the prompt.

Compositional collapse occurs when
\begin{equation}
g_k(x_0) \notin \mathcal{S}_k(p),
\end{equation}
typically drifting toward a high-probability alternative under the model prior:
\begin{equation}
g_k(x_0) \in \mathcal{S}_k^{\text{prior}},
\quad
\mathcal{S}_k^{\text{prior}} \neq \mathcal{S}_k(p).
\end{equation}

Our dataset is explicitly designed to induce such conflicts between target composition and prior preference.

\paragraph{Design Principles.}
The dataset is constructed under four key principles:
\begin{itemize}
    \item \textbf{Physical Plausibility:} All prompts are visually realizable.
    \item \textbf{Controlled Rarity:} Prompts lie in low-density regions of $p_\theta(x_0 \mid p)$.
    \item \textbf{Failure Isolation:} Each category isolates a distinct semantic factor $g_k$.
    \item \textbf{Non-Triviality:} Prompts avoid degenerate solutions such as object removal.
\end{itemize}

\subsubsection*{(1) ENV: Environment Recomposition}

This category evaluates rare combinations of environmental factors:
\begin{quote}
``A snowy beach with waves.'' \\
``A desert oasis surrounded by snow.'' \\
``A lake shore with snow and blooming flowers.''
\end{quote}

Here, the constraint applies to the environment factor:
\[
g_{\text{env}}(x_0) \in \mathcal{S}_{\text{env}}(p),
\]
where $\mathcal{S}_{\text{env}}(p)$ includes atypical co-occurrences. Models often collapse toward a dominant single-environment mode.

\subsubsection*{(2) TEMP: Temporal Misalignment}

This category evaluates rare temporal and astronomical co-occurrences:
\begin{quote}
``A rainbow at night in the sky.'' \\
``A sunrise with stars still visible in the sky.'' \\
``A sunrise sky with a visible rainbow and moon together.''
\end{quote}

The constraint is defined over temporal factors:
\[
g_{\text{time}}(x_0) \in \mathcal{S}_{\text{time}}(p),
\]
which conflicts with learned time-of-day priors.

\subsubsection*{(3) OBJ: Object Relocation}

This category evaluates object-environment mismatches:
\begin{quote}
``A lighthouse located in a grassy meadow.'' \\
``A canoe placed in a dry canyon.'' \\
``A surfboard standing in a grassy field.''
\end{quote}

We require:
\[
(g_{\text{obj}}(x_0), g_{\text{env}}(x_0)) \in \mathcal{S}_{\text{joint}}(p),
\]
where the joint configuration is rare but valid.

\subsubsection*{(4) ATTR: Attribute Rebinding}

This category evaluates unusual attribute-material bindings:
\begin{quote}
``A glowing clay pot.'' \\
``A reflective wooden table like a mirror.'' \\
``A solid cloud shaped like a rock.''
\end{quote}

The constraint operates on attribute consistency:
\[
g_{\text{attr}}(x_0) \in \mathcal{S}_{\text{attr}}(p),
\]
which may conflict with learned material priors.

\subsubsection*{(5) SCALE: Scale Shift}

This category evaluates deviations in object scale:
\begin{quote}
``A giant cat larger than a building.'' \\
``A tiny chair placed on a finger.'' \\
``A tiny bench on a coin.''
\end{quote}

The constraint is:
\[
g_{\text{scale}}(x_0) \in \mathcal{S}_{\text{scale}}(p),
\]
which often conflicts with typical size distributions.

\subsubsection*{(6) CTX: Contextual Relocation}

This category evaluates unusual scene placement:
\begin{quote}
``A kitchen in the middle of a highway.'' \\
``A library on a beach.'' \\
``A restaurant dining area in the middle of a highway.''
\end{quote}

The constraint is defined over context:
\[
g_{\text{ctx}}(x_0) \in \mathcal{S}_{\text{ctx}}(p),
\]
which competes with strong context priors.

\subsubsection*{(7) MAT: Material and State Conflict}

This category evaluates inconsistencies in material or physical state:
\begin{quote}
``A melting ice sculpture in a snowy field.'' \\
``A steaming ice cube on a snowy ground.'' \\
``A wet fire burning with visible flames.''
\end{quote}

Here:
\[
g_{\text{state}}(x_0) \in \mathcal{S}_{\text{state}}(p),
\]
which may violate typical thermodynamic or material consistency.

\subsubsection*{(8) DENS: Composition Density Variation}

This category evaluates deviations in population or object density:
\begin{quote}
``A lightly crowded subway platform during peak hours.'' \\
``A lightly crowded shopping street during sales season.'' \\
``A lightly crowded bus stop during commute time.''
\end{quote}

The constraint is:
\[
g_{\text{dens}}(x_0) \le \tau(p),
\]
where $\tau(p)$ specifies a low-density threshold rather than strict absence.

\paragraph{Summary.}
Together, ENV, TEMP, OBJ, ATTR, SCALE, CTX, MAT, and DENS form a structured evaluation suite that probes multiple dimensions of compositional collapse, including environment recomposition, temporal misalignment, object-context mismatch, attribute binding, scale consistency, contextual placement, material-state coherence, and density variation.

Each category isolates a specific semantic factor $g_k$, enabling fine-grained analysis of failure modes in diffusion models. By modeling compositional constraints as in Eq.~\eqref{eq:dataset_objective}, the benchmark supports systematic evaluation of compositional fidelity in diffusion models.

\begin{figure*}[t] 
    \centering 
    \includegraphics[width=0.7\linewidth] {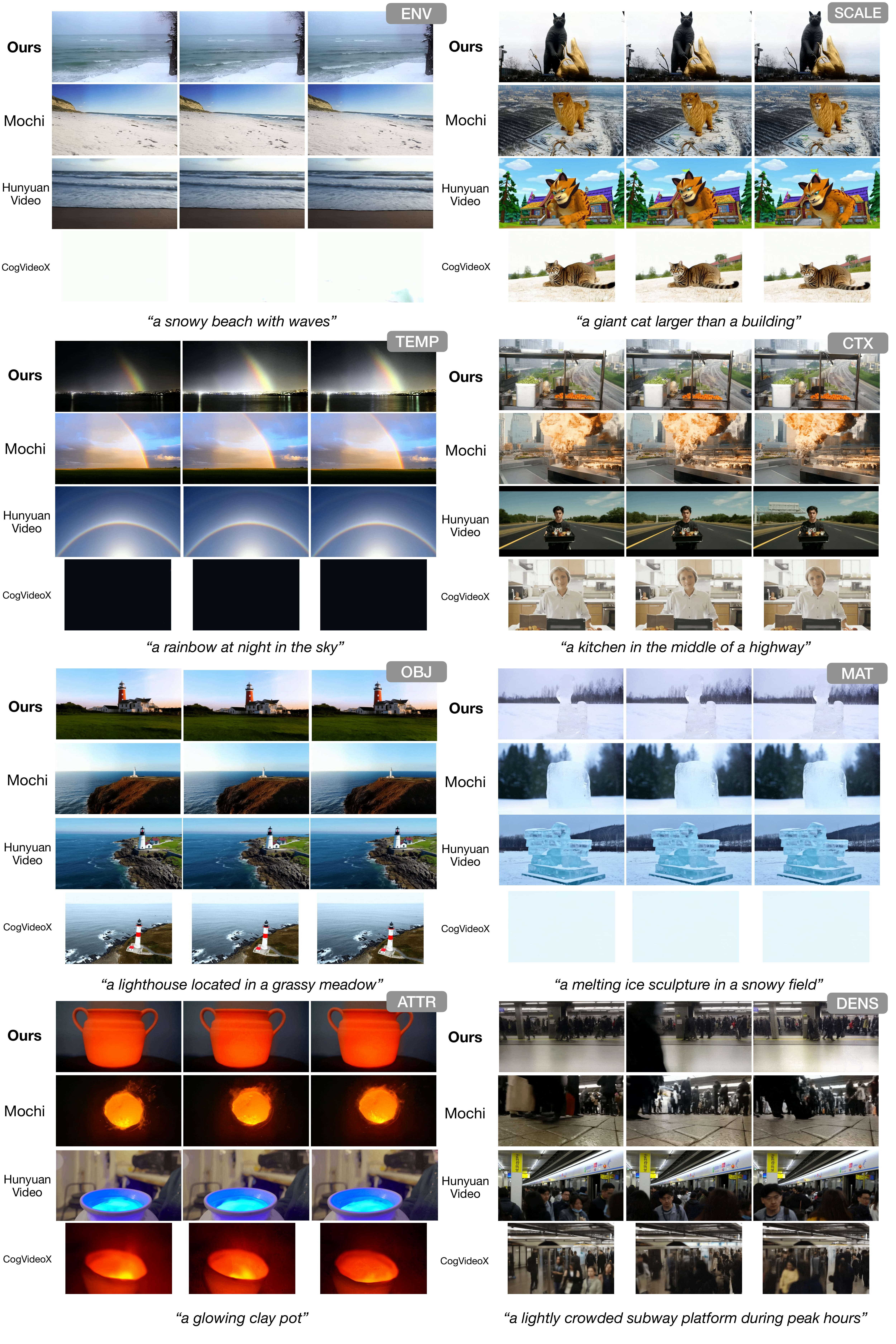} 
    \caption{\textbf{Qualitative comparison across diverse compositional scenarios (Set 1).} We evaluate our method against state-of-the-art video diffusion models (Mochi, HunyuanVideo, CogVideoX) on eight rare compositional categories: ENV (Environment Recomposition), TEMP (Temporal Misalignment), OBJ (Object Relocation), ATTR (Attribute Rebinding), SCALE (Scale Shift), CTX (Contextual Relocation), MAT (Material and State Conflict), and DENS (Composition Density Variation). Each row shows frames sampled from generated videos for a representative prompt per category. Baseline models consistently collapse toward frequent semantic alternatives (for instance, generating a daytime rainbow instead of a nocturnal one, or a tropical beach instead of a snowy one), whereas our method faithfully realizes both compositional factors simultaneously, resisting default completion bias while preserving overall scene coherence.} 
    \label{fig:detailed_qualitative_supp1} 
\end{figure*}

\begin{figure*}[t]
    \centering
    \includegraphics[width=0.7\linewidth]
    {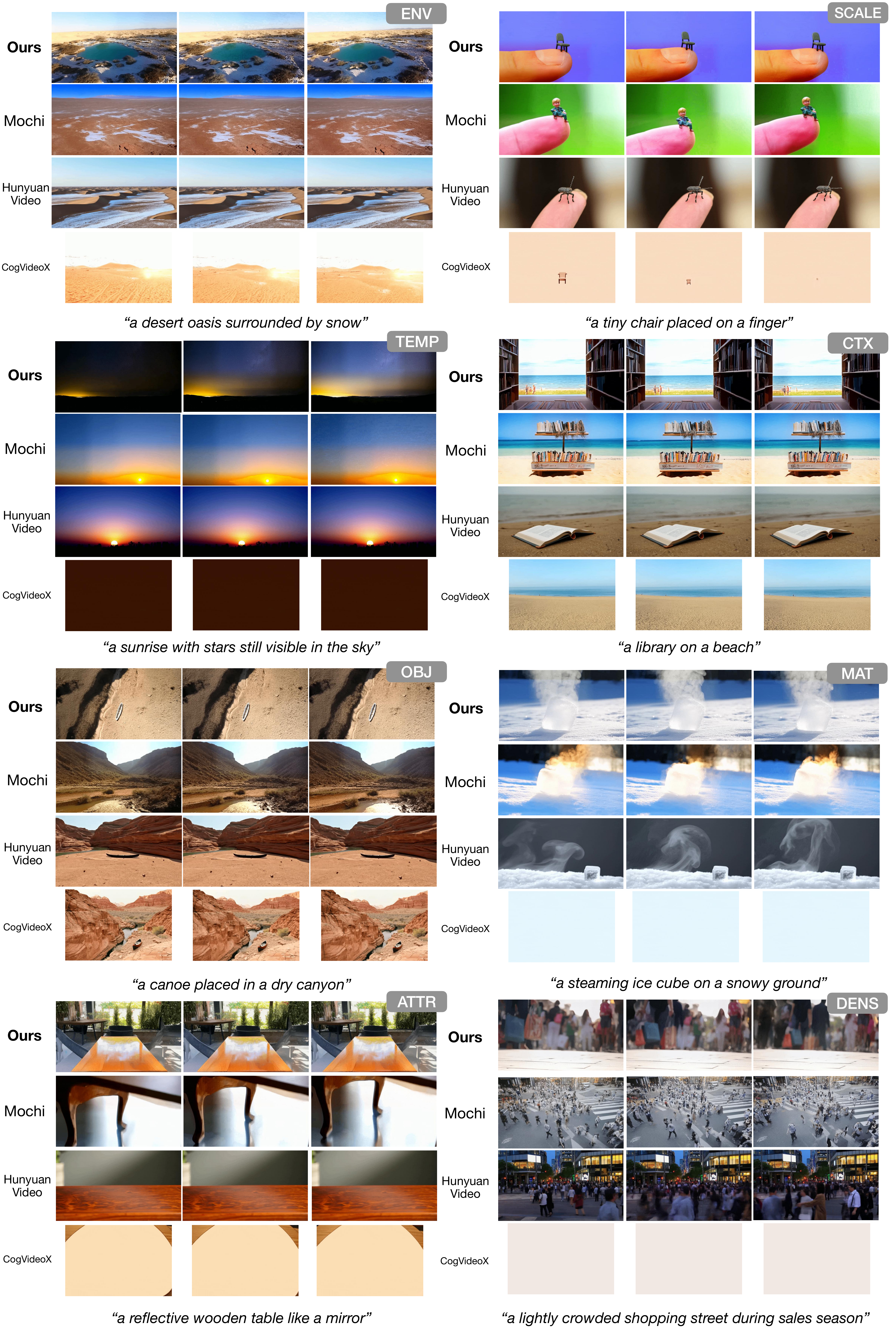}
    \caption{\textbf{Qualitative comparison across diverse
    compositional scenarios (Set 2).} This figure presents a second set of representative prompts covering the same eight compositional categories as in Figure~\ref{fig:detailed_qualitative_supp1} (ENV, TEMP, OBJ, ATTR, SCALE, CTX, MAT, DENS). These examples include more challenging compositions such as \emph{a desert oasis surrounded by snow}, \emph{a tiny chair placed on a finger}, and \emph{a steaming ice cube on snowy ground}. While baseline models frequently drift toward high-frequency alternatives or fail to realize the intended rare factor, our method consistently suppresses attractor-driven collapse and produces semantically coherent outputs across diverse scene contexts, from natural environments to structured indoor and object-level compositions.} \label{fig:detailed_qualitative_supp2}
\end{figure*}

\begin{figure*}[t]
    \centering
    \includegraphics[width=0.7\linewidth]
    {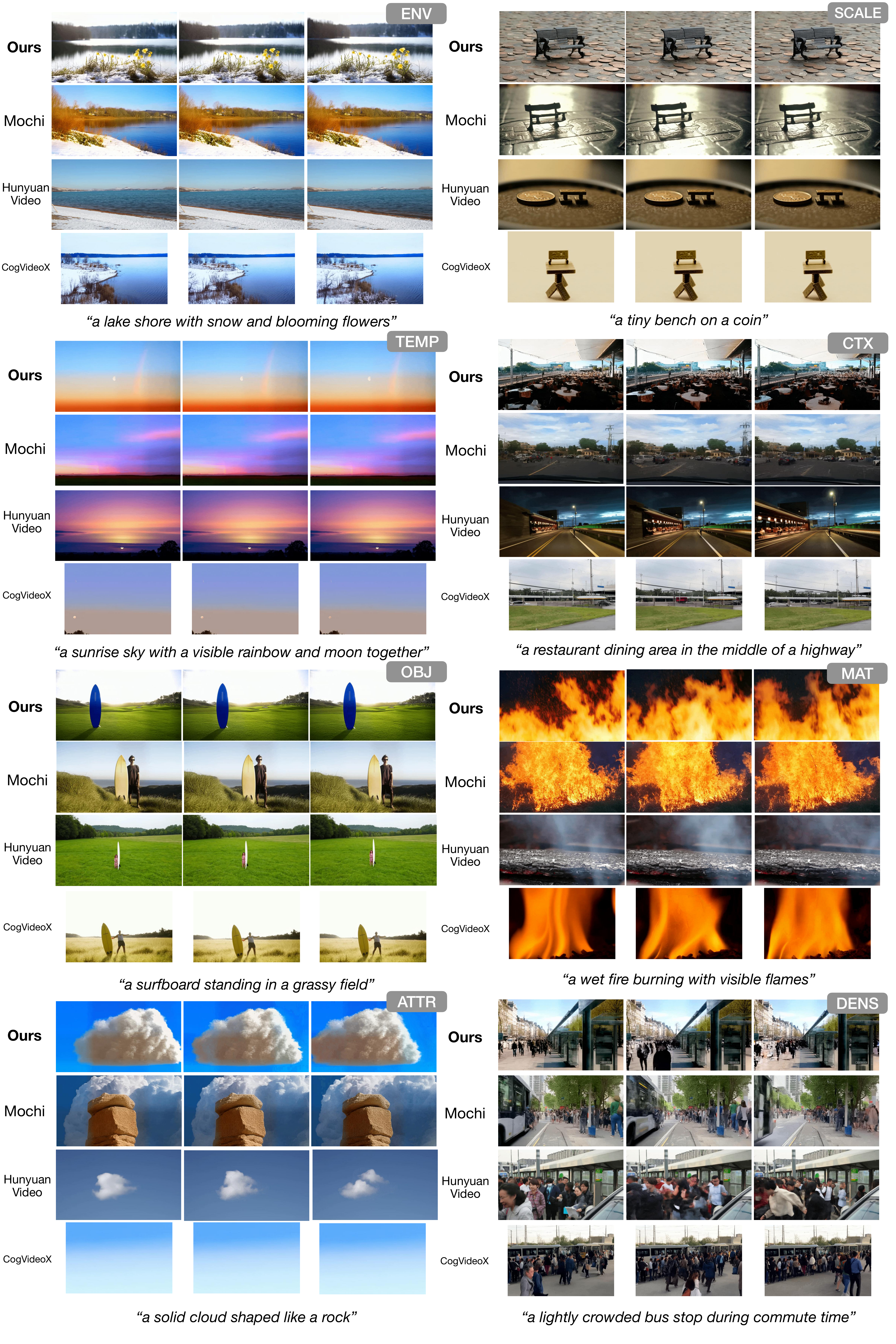}
    \caption{\textbf{Qualitative comparison across diverse
    compositional scenarios (Set 3).} This figure presents a third set of prompts spanning
    all eight compositional categories, including \emph{a lake shore with snow and blooming flowers}, \emph{a surfboard standing in a grassy field}, \emph{a solid cloud shaped like a rock}, and \emph{a wet fire burning with visible flames}. These cases probe more subtle compositional conflicts, such as simultaneous material-state inconsistency and attribute rebinding. Across all settings, baseline models exhibit attractor collapse toward more common scene configurations, whereas our method applies targeted repulsion to maintain the intended rare composition without degrading visual quality or scene realism.} \label{fig:detailed_qualitative_supp3}
\end{figure*}

\section{Detailed Qualitative Results} 
\label{sec:detailed_qualitative} 
Figures~\ref{fig:detailed_qualitative_supp1}, \ref{fig:detailed_qualitative_supp2}, and~\ref{fig:detailed_qualitative_supp3} present qualitative comparisons across all eight compositional categories of DCR-Bench, spanning three sets of representative prompts. For each prompt, we show uniformly sampled frames from videos generated by Mochi, HunyuanVideo, CogVideoX, and our method (DCR applied to Mochi). 

\paragraph{ENV and TEMP.} Environment recomposition and temporal misalignment represent cases where the rare compositional factor directly conflicts with a dominant distributional prior. In ENV, baseline models consistently collapse toward a single-environment mode: for example, \emph{a snowy beach with waves} is rendered as a standard tropical or sandy beach by all baselines, while our method preserves both the beach setting and the snow simultaneously. In TEMP, cases such as \emph{a rainbow at night in the sky} reveal a strong daytime prior in all backbone models: Mochi and HunyuanVideo generate brightly lit scenes with daytime rainbows, and CogVideoX produces a fully dark frame, failing to realize either factor. Our method generates a coherent nocturnal scene with a visible rainbow, successfully realizing both temporal and atmospheric components. 

\paragraph{OBJ and CTX.} Object relocation and contextual relocation probe whether models can place objects or scene elements in environments that deviate from typical co-occurrence patterns. In OBJ, prompts such as \emph{a lighthouse located in a grassy meadow} and \emph{a canoe placed in a dry canyon} require placing an object in an atypical environment. Baselines frequently revert the setting to the most common context for the object: placing the lighthouse on a rocky coast, or the canoe on a river. Our method maintains the intended environment while correctly placing the object within it. In CTX, prompts such as \emph{a kitchen in the middle of a highway} and \emph{a restaurant dining area in the middle of a highway} are particularly challenging, as both contextual elements must be jointly realized. Baselines tend to generate either the kitchen or the highway in isolation, while our method produces coherent compositions that integrate both elements. 

\paragraph{ATTR and SCALE.} Attribute rebinding and scale shift test the model's ability to assign unusual properties to objects. In ATTR, \emph{a glowing clay pot} elicits strong responses: Mochi and CogVideoX produce glowing outputs, but with incorrect material renderings, while HunyuanVideo generates a completely different glowing object. Our method correctly combines the clay pot form with a luminous attribute. For SCALE, prompts such as \emph{a giant cat larger than a building} and \emph{a tiny bench on a coin} require scale relationships that are drastically different from learned priors. Baselines consistently generate objects at typical scales, ignoring the compositional constraint. Our method realizes the intended scale relationship, placing a miniature bench on a coin surface or rendering a cat at building scale with spatial coherence. 

\paragraph{MAT and DENS.} Material-state conflict and density variation represent the most subtle compositional challenges. In MAT, prompts such as \emph{a melting ice sculpture in a snowy field} and \emph{a wet fire burning with visible flames} require physically inconsistent but visually describable states. Mochi and HunyuanVideo tend to generate one state or the other (either a static ice sculpture or a standard fire), while our method produces outputs that incorporate both states. In DENS, prompts such as \emph{a lightly crowded subway platform during peak hours} require the model to generate a scene that contradicts its density prior. Baselines produce crowded platforms consistent with typical peak-hour scenes, whereas our method generates a visibly sparse platform that nonetheless preserves the environmental context. 

\paragraph{Overall Observations.} Across all three sets, a consistent failure pattern emerges in baseline models: the denoising trajectory collapses toward the most statistically dominant completion, ignoring one or both compositional factors specified in the prompt. This manifests as environment substitution in ENV, temporal collapse in TEMP, context reversion in OBJ and CTX, attribute neglect in ATTR and SCALE, and state simplification in MAT. Our method applies targeted repulsion along the attractor direction at each denoising step, preventing this collapse while preserving the remaining semantic components of the guidance update. The results show that Default Completion Repulsion substantially improves the realization of rare but semantically valid compositions that standard diffusion models consistently fail to produce.

\begin{table*}[t]
\begin{small}
    \centering
    \resizebox{\textwidth}{!}{%
    \begin{tabular}{lccccc}
        \toprule
        \textbf{Method} & \textbf{CLIPScore} $\uparrow$ & \textbf{CLIP-attr} $\downarrow$ & \textbf{BLIP} $\uparrow$ & \textbf{CCS} $\uparrow$ & \textbf{CVR} $\downarrow$ \\
        \midrule
        Mochi            & 0.2986 $\pm$ 0.0300 & 0.2718 $\pm$ 0.0395 & 0.7807 $\pm$ 0.0834 & 3.8375 $\pm$ 1.2865 & 0.4725 $\pm$ 0.4998 \\
        HunyuanVideo     & 0.3067 $\pm$ 0.0312 & 0.2725 $\pm$ 0.0368 & 0.7819 $\pm$ 0.0811 & 3.9750 $\pm$ 1.2662 & 0.3750 $\pm$ 0.4847 \\
        CogVideoX        & 0.2858 $\pm$ 0.0491 & 0.2644 $\pm$ 0.0454 & 0.7290 $\pm$ 0.1180 & 3.1925 $\pm$ 1.6830 & 0.5175 $\pm$ 0.5003 \\
        \midrule
        \textbf{Ours}    & \textbf{0.3174 $\pm$ 0.0318} & \textbf{0.2558 $\pm$ 0.0349} & \textbf{0.8075 $\pm$ 0.0592} & \textbf{4.1300 $\pm$ 1.2931} & \textbf{0.3100 $\pm$ 0.4630} \\
        \midrule
        Negative Prompt  & 0.2736 $\pm$ 0.0363 & 0.2610 $\pm$ 0.0362 & 0.7065 $\pm$ 0.0813 & 3.0375 $\pm$ 1.5466 & 0.4375 $\pm$ 0.4966 \\
        Ours w/o Attractor Prompt$^\dagger$ & 0.2986 $\pm$ 0.0300 & 0.2718 $\pm$ 0.0395 & 0.7807 $\pm$ 0.0834 & 3.8375 $\pm$ 1.2865 & 0.4725 $\pm$ 0.4998 \\
        Ours w/o Repulsion$^\dagger$        & 0.2986 $\pm$ 0.0300 & 0.2718 $\pm$ 0.0395 & 0.7807 $\pm$ 0.0834 & 3.8375 $\pm$ 1.2865 & 0.4725 $\pm$ 0.4998 \\
        Ours w/o Schedule         & 0.3115 $\pm$ 0.0309 & 0.2740 $\pm$ 0.0402 & 0.7819 $\pm$ 0.0851 & 3.9675 $\pm$ 1.3155 & 0.3550 $\pm$ 0.4791 \\
        \bottomrule
    \end{tabular}}
    \caption{\textbf{Quantitative evaluation results.} All metrics are reported as mean $\pm$ standard deviation over the full evaluation set. Mochi, HunyuanVideo, and CogVideoX denote standard CFG baselines. Negative Prompt denotes the baseline where $p_{\text{attr}}$ is provided as a negative prompt within standard CFG on Mochi. Ablation variants are based on Mochi. CCS denotes the Compositional Compliance Score and CVR denotes the Compositional Violation Rate, both obtained from a GPT-4o vision-language judge. $^\dagger$By construction these variants leave the final denoising update identical to standard CFG ($\lambda_t = 0$; Sec.~\ref{subsec:projection_main}), so their rows coincide with the Mochi baseline; they are listed to make the reduction explicit.}
    \label{tab:dcr_supp}
\end{small}
\end{table*}

\section{Detailed Quantitative Evaluation}
\label{sec:detailed_quantitative_evaluation}
We evaluate whether generated videos satisfy the full compositional constraint using five complementary metrics. Together, CLIPScore, CLIP-attr, and BLIP capture alignment with the intended composition, suppression of the frequent alternative, and semantic consistency across modalities, while CCS and CVR evaluate compositional fidelity at the semantic reasoning level rather than within embedding space. Our evaluation focuses on compositional fidelity: whether the generated video simultaneously realizes both constituent factors of the intended rare composition, and whether it resists collapse toward the more frequent attractor completion.

\paragraph{CLIPScore.}
We compute CLIPScore by measuring cosine similarity between each generated frame and the full textual prompt $p$. For a video consisting of $T$ frames $\{I_t\}_{t=1}^T$ and prompt $p$, we compute
\[
\text{CLIPScore}
= \frac{1}{T} \sum_{t=1}^{T}
\cos\!\big(E_{\text{img}}(I_t),\, E_{\text{text}}(p)\big).
\]
Higher CLIPScore indicates stronger global alignment between the generated video and the full intended composition.

\paragraph{CLIP Similarity to Attractor Prompt (CLIP-attr).}
To measure the degree of collapse toward the frequent counterpart, we compute the cosine similarity between generated frames and the attractor prompt $p_{\text{attr}}$:
\[
\text{CLIP-attr}
= \frac{1}{T} \sum_{t=1}^{T}
\cos\!\big(E_{\text{img}}(I_t),\, E_{\text{text}}(p_{\text{attr}})\big).
\]
Lower CLIP-attr values indicate stronger suppression of the default completion tendency. This metric is the DCR analogue of CLIP-neg used in negation-aware generation evaluation~\cite{anonymous2026negate}, adapted to measure attractor-directional collapse rather than forbidden concept presence. Because $p_{\text{attr}}$ is explicitly defined as part of our framework, this metric directly reflects the core failure mode our method targets.

\paragraph{BLIP Caption Alignment.}
We evaluate semantic fidelity using BLIP-based captioning. For each frame, we generate a caption $\hat{C}_t$ using a pretrained BLIP model and compute its similarity to the original prompt:
\[
\text{BLIP}
= \frac{1}{T} \sum_{t=1}^{T}
\cos\!\big(E_{\text{text}}(\hat{C}_t),\,
E_{\text{text}}(p)\big).
\]
This provides a complementary text-level consistency measure based on generated descriptions rather than direct image-text matching. Unlike CLIPScore, which compares visual features against the prompt embedding directly, BLIP evaluates whether the semantic content of generated frames, as interpreted by an independent captioning model, is consistent with the intended composition.

Together, CLIPScore, CLIP-attr, and BLIP quantify (1) global prompt alignment, (2) attractor suppression, and (3) caption-level compositional fidelity, providing embedding-space measurements that are complementary in their sensitivity to different aspects of generation quality.

\paragraph{Vision-Language Compositional Compliance Evaluation (CCS and CVR).}
While the preceding metrics operate within embedding spaces, they do not directly verify whether both compositional factors are simultaneously present and coherently realized in the generated video. In particular, embedding-based metrics may fail to capture cases where one compositional factor dominates the visual representation while the other is absent, or where the generation reflects the attractor completion at a semantic level not fully captured by cosine similarity. To address this limitation, we introduce two complementary metrics based on a direct vision-language judge: the \emph{Compositional Compliance Score (CCS)} and the \emph{Compositional Violation Rate (CVR)}.

\textbf{Evaluation Principle.}
Instead of comparing embeddings, we directly query a multimodal large language model capable of joint visual and textual reasoning. The judge receives (i) the full original textual prompt $p$ describing the intended rare composition, (ii) the two constituent compositional factors extracted from $p$ (e.g., \emph{snowy} and \emph{beach} for the prompt \emph{a snowy beach}), (iii) the attractor prompt $p_{\text{attr}}$ representing the frequent counterpart (e.g., \emph{a tropical beach}), and (iv) multiple frames uniformly sampled from the generated video across its temporal span. The frames are resized with preserved aspect ratio to ensure consistent visual input resolution. No cropping, object masking, or post-processing is applied.

\textbf{Judge Configuration.}
We use GPT-4o as a multimodal judge accessed through the OpenAI API. The model is instructed to evaluate \emph{compositional fidelity only}, explicitly assessing whether both specified factors are simultaneously present and coherently composed in the video, and whether the output reflects the intended rare composition rather than the frequent attractor counterpart. The evaluation rubric is fixed and provided verbatim to the judge to ensure consistency across all evaluated samples. The model operates with deterministic decoding (temperature set to 0) to reduce stochastic variability.

\textbf{Scoring Rubric.}
For each video, the judge assigns an integer score $s \in \{1,2,3,4,5\}$ defined as:
\begin{itemize}
    \item 1: Neither compositional factor is present; the output reflects neither the intended composition nor a semantically coherent alternative. 
    \item 2: Only one compositional factor is present, or the output has collapsed entirely toward the attractor completion $p_{\text{attr}}$. 
    \item 3: Both factors are partially present but incoherently composed, or the output is ambiguous between the intended composition and the attractor. 
    \item 4: Both factors are present and mostly coherently composed, with minor ambiguity or imperfection. 
    \item 5: Both factors are fully and coherently present; the output clearly reflects the intended rare composition rather than the frequent alternative.
\end{itemize}

In addition, the judge outputs a binary indicator \texttt{collapsed}, set to \texttt{True} when the generated video is judged to reflect the attractor prompt $p_{\text{attr}}$ rather than the intended rare composition $p$, based on the judge's multimodal reasoning over the provided frames.

\paragraph{Compositional Compliance Score (CCS).}
Let $s_i$ denote the integer compliance score assigned to video $i$ among $N$ evaluated samples. We define:
\[
\text{CCS} = \frac{1}{N} \sum_{i=1}^{N} s_i.
\]
CCS ranges from 1 to 5. Higher values indicate stronger simultaneous realization of both compositional factors in the generated video. Because a high score requires both factors to be simultaneously present and coherently realized, degenerate outputs that suppress the attractor by degrading overall content, such as the near-black Negative Prompt generations, receive low CCS rather than being rewarded; CCS therefore also guards against quality-destroying solutions.

\paragraph{Compositional Violation Rate (CVR).}
Let $v_i$ be defined as:
\[
v_i =
\begin{cases}
1 & \text{if } \texttt{collapsed}_i = \texttt{True}, \\
0 & \text{otherwise}.
\end{cases}
\]
We define:
\[
\text{CVR} = \frac{1}{N} \sum_{i=1}^{N} v_i.
\]
CVR measures the empirical frequency of attractor collapse across the evaluation set. Lower values indicate stronger resistance to default completion bias. CVR serves as the DCR analogue of the Negation Violation Rate (NVR) introduced in~\cite{anonymous2026negate}, adapted to measure collapse toward frequent compositional alternatives rather than presence of forbidden concepts.

\textbf{Aggregation Protocol.}
All CCS and CVR values are computed over the full evaluation set without thresholding, sample exclusion, or post-hoc correction. Ambiguous cases (score $= 3$) are retained in the computation. The binary collapse indicator directly reflects the judge's decision regarding attractor-directed generation, without manual reinterpretation.

\textbf{Rationale.}
Unlike embedding-based metrics, CCS and CVR evaluate compositional fidelity through direct multimodal reasoning over visual evidence. This removes reliance on CLIP similarity space and avoids potential circularity arising from the fact that CLIP embeddings may not fully represent rare compositional semantics. By combining embedding-based alignment metrics with direct vision-language compliance scoring, we obtain complementary measurements of compositional fidelity and attractor suppression that operate at both the representational and reasoning levels.

\paragraph{Overall Performance.} As summarized in Table~\ref{tab:dcr_supp}, our method achieves the best performance across all evaluation metrics. In terms of embedding-based alignment, it improves CLIPScore (0.3174) over strong baselines such as HunyuanVideo (0.3067), while simultaneously reducing CLIP-attr (0.2558), indicating effective suppression of attractor bias without sacrificing global prompt alignment. Negative Prompt reduces CLIP-attr to 0.261, but at the cost of substantially lower CLIPScore (0.2736) and BLIP (0.7065), suggesting that naive attractor suppression via negative conditioning degrades overall semantic fidelity. In contrast, our method achieves lower CLIP-attr (0.2558) while simultaneously improving CLIPScore and BLIP (0.8075), demonstrating that projection-based repulsion avoids this trade-off. Our method also attains the highest BLIP score (0.8075), confirming improved semantic consistency at the caption level. Gains are larger still on the reasoning-based metrics. Our approach achieves the highest CCS (4.1300) and the lowest CVR (0.3100), substantially reducing attractor-driven collapse compared to Mochi (CVR: 0.4725) and Negative Prompt (CVR: 0.4375). This demonstrates that improvements are not limited to embedding similarity, but translate to actual compositional correctness under multimodal reasoning. Ablation results further localize the contribution of the mechanism. The w/o Repulsion and w/o Attractor Prompt variants reduce analytically to standard CFG (Sec.~\ref{subsec:projection_main}), so their rows coincide with the Mochi baseline, and the full gap to the complete method (CCS $3.8375 \to 4.1300$, CVR $0.4725 \to 0.3100$) is attributable to the attractor-guided projection. Removing only the schedule retains part of the projection's benefit (CCS 3.9675) but raises CVR to 0.3550, showing that restricting repulsion to a targeted interval prevents over-correction of the denoising trajectory.

\begin{table*}[t]
\centering
\small
\begin{tabular}{llccc}
    \toprule
    \textbf{Category} & \textbf{Method} & \textbf{CLIPScore} $\uparrow$ & \textbf{CCS} $\uparrow$ & \textbf{CVR} $\downarrow$ \\
    \midrule
    \multirow{3}{*}{ENV}
    & Mochi           & 0.2977 & 3.5400 & 0.4800 \\
    & Negative Prompt & 0.2777 & 3.3200 & 0.4000 \\
    & \textbf{Ours}   & \textbf{0.3142} & \textbf{4.3600} & \textbf{0.3600} \\
    \midrule
    \multirow{3}{*}{TEMP}
    & Mochi           & 0.3017 & 4.1400 & 0.4600 \\
    & Negative Prompt & 0.2747 & 3.0400 & 0.4800 \\
    & \textbf{Ours}   & \textbf{0.3095} & \textbf{4.2200} & \textbf{0.4000} \\
    \midrule
    \multirow{3}{*}{OBJ}
    & Mochi           & 0.3092 & 3.8800 & 0.4200 \\
    & Negative Prompt & 0.3076 & 3.9400 & 0.4200 \\
    & \textbf{Ours}   & \textbf{0.3367} & \textbf{4.2400} & \textbf{0.2800} \\
    \midrule
    \multirow{3}{*}{ATTR}
    & Mochi           & 0.2958 & 3.8200 & 0.5800 \\
    & Negative Prompt & 0.2600 & 2.7600 & 0.4400 \\
    & \textbf{Ours}   & \textbf{0.3082} & \textbf{3.8400} & \textbf{0.4000} \\
    \midrule
    \multirow{3}{*}{SCALE}
    & Mochi           & 0.2912 & 3.8000 & 0.3600 \\
    & Negative Prompt & 0.2782 & 3.5600 & 0.3000 \\
    & \textbf{Ours}   & \textbf{0.3266} & \textbf{3.9600} & \textbf{0.2800} \\
    \midrule
    \multirow{3}{*}{CTX}
    & Mochi           & 0.3017 & 4.1000 & 0.3200 \\
    & Negative Prompt & 0.2743 & 2.1200 & 0.5400 \\
    & \textbf{Ours}   & \textbf{0.3362} & \textbf{4.1800} & \textbf{0.2600} \\
    \midrule
    \multirow{3}{*}{MAT}
    & Mochi           & 0.2975 & 3.3800 & 0.7200 \\
    & Negative Prompt & 0.2736 & 2.7400 & 0.5400 \\
    & \textbf{Ours}   & \textbf{0.3129} & \textbf{4.1400} & \textbf{0.1800} \\
    \midrule
    \multirow{3}{*}{DENS}
    & Mochi           & 0.2939 & 4.0400 & 0.4400 \\
    & Negative Prompt & 0.2426 & 2.8200 & 0.3800 \\
    & \textbf{Ours}   & \textbf{0.2952} & \textbf{4.1000} & \textbf{0.3200} \\
    \bottomrule
\end{tabular}
\caption{\textbf{Category-wise quantitative evaluation on DCR-Bench.}
We report CLIPScore ($\uparrow$), CCS ($\uparrow$), and CVR ($\downarrow$)
across all eight compositional categories.
All methods use Mochi as the backbone unless otherwise noted.
Categories are defined in Appendix~\ref{supp:detailed_dataset}.}
\label{tab:dcr_category}
\end{table*}

\paragraph{Category-wise Compositional Analysis.} 
Table~\ref{tab:dcr_category} reports CLIPScore, CCS, and CVR across all eight compositional categories. Several consistent trends emerge from this breakdown. First, the benefits of DCR are most pronounced in categories that require suppression of strong distributional priors. In MAT, our method reduces CVR from $0.72$ (Mochi) to $0.18$, the largest absolute reduction across all categories, while improving CCS from $3.38$ to $4.14$. This indicates that material-state conflicts, such as \emph{a melting ice sculpture in a snowy field}, are particularly susceptible to attractor collapse, and that targeted repulsion is highly effective in these settings. Similarly, in ENV, CVR drops from $0.48$ to $0.36$ with a CCS gain of $+0.82$, confirming that environment recomposition benefits strongly from explicit attractor suppression. Second, SCALE and CTX show the largest CLIPScore improvements ($+0.035$ and $+0.034$ respectively), alongside substantial CVR reductions. In CTX, Negative Prompt produces the lowest CCS ($2.12$) among all methods and categories, confirming that naive negative conditioning is particularly harmful when both contextual elements must be jointly realized. Third, TEMP and CTX represent cases where Mochi itself achieves relatively high CCS ($4.14$ and $4.10$), yet DCR still improves both CCS and CVR. This indicates that even when baseline models partially succeed, the attractor direction remains active and DCR provides meaningful additional suppression. Finally, DENS shows the smallest absolute CVR improvement ($0.44 \to 0.32$), consistent with the observation that density constraints are more diffuse and harder to localize within the guidance update direction. Nevertheless, our method maintains Ours as the best-performing variant across all eight categories on all three reported metrics, confirming the broad applicability of the DCR framework.

\paragraph{Rationale for Automated Evaluation.} 
Unlike embedding-based metrics that operate within CLIP feature space, CCS and CVR are obtained from GPT-4o, a multimodal model that directly reasons over video frames and assesses whether both compositional factors are simultaneously present and coherently realized, the same judgment a human annotator would perform. The judge is configured with a fixed rubric, deterministic decoding (temperature~$=0$), and explicit instructions to evaluate compositional fidelity only, minimizing subjectivity and ensuring consistency across all 400 evaluated samples. The validity of this protocol is supported by prior work on compositional video generation~\cite{anonymous2026negate}, which demonstrated that GPT-based compliance scoring and human perceptual evaluation yield consistent rankings across methods on closely related tasks, confirming that vision-language judge scores serve as a reliable proxy for human judgment of compositional fidelity. This configuration also addresses the potential circularity of using the same model family both to construct attractor prompts and to evaluate compliance, a concern we revisit in the Limitations; the fixed rubric and deterministic decoding leave the judge no channel through which the attractor construction could influence its scores.

\section{Detailed Ablation Study}
\label{sec:detailed_ablation}

We provide extended quantitative and qualitative analysis of the four ablation variants reported in the main paper. Results are summarized in Table~\ref{tab:dcr_supp} and illustrated in Figure~\ref{results:ablation}.

\paragraph{Negative Prompt Baseline.}
Providing $p_{\text{attr}}$ as a CFG negative prompt constitutes the most direct alternative to our method, yet yields the worst overall performance. CLIPScore drops to $0.2736 \pm 0.0363$ and BLIP to $0.7065 \pm 0.0813$, the lowest among all variants, while CCS falls to $3.0375 \pm 1.5466$ and CVR remains elevated at $0.4375 \pm 0.4966$. Qualitatively, the generated frames are near-black and featureless, failing to render any compositional content. This failure arises because standard CFG negative guidance exerts an unconditional, undirected push away from $p_{\text{attr}}$ at every denoising step, suppressing not only the unwanted attribute but the broader semantic context necessary to synthesize the rare composition. The results confirm that naive negation via CFG is unsuitable for compositional rarity control.

\paragraph{Effect of Removing the Attractor Prompt.}
Without a semantically distinct $p_{\text{attr}}$, the probe branch is conditioned on the main prompt itself, so the drift reduces to $a_t = (w_{\text{attr}} - w)(\epsilon_{\text{text}} - \epsilon_{\text{uncond}})$ with $w_{\text{attr}} < w$, which is anti-parallel to $\Delta_{\text{ref}}$. The positive-alignment gate then yields $\lambda_t = 0$ at every step, and the final update is identical to standard CFG; the corresponding row in Table~\ref{tab:dcr_supp} coincides with the Mochi baseline by construction. This variant nonetheless delimits the method: the repulsion mechanism has no well-defined direction to suppress unless the attractor prompt is semantically distinct from the target, so the LLM relaxation step is a prerequisite for the projection rather than an optional refinement.

\paragraph{Effect of Removing the Repulsion Update.}
Setting $\lambda_t = 0$ while retaining the probe disables the correction term, so the final update equals $\Delta_{\text{ref}}$ and the generation is identical to standard CFG by definition; the corresponding row in Table~\ref{tab:dcr_supp} coincides with the Mochi baseline. The attractor branch is still evaluated at every step (its cost is visible in Appendix~\ref{supp:computation_text}) but its output is unused, which makes this variant a useful implementation check: any deviation from the baseline under a shared seed would indicate a discrepancy between implementation and formulation. The gap between this row and the full model is therefore the net effect of the projection itself.

\paragraph{Effect of Removing Constraint Scheduling.}
Fixing $\alpha_t = 1$ uniformly applies repulsion throughout the entire denoising trajectory, including early high-noise steps where global scene layout is determined. This produces the highest CVR among the three model-based ablations ($0.3550 \pm 0.4791$) and the largest CLIP-attr ($0.2740 \pm 0.0402$), despite a CLIPScore comparable to the full model ($0.3115$). Qualitatively, early-step interference disrupts coarse structure formation, yielding an oversaturated close-up arc that loses the intended nocturnal scene scale and context. The elevated CVR variance further indicates that suppression becomes unstable across prompts when the scheduling window is removed. These results confirm that temporally restricting repulsion to a targeted interval is essential: premature enforcement corrupts the structural scaffold before semantic detail can be reliably steered.

\begin{figure*}[t]
    \centering
    \includegraphics[width=0.65\linewidth]{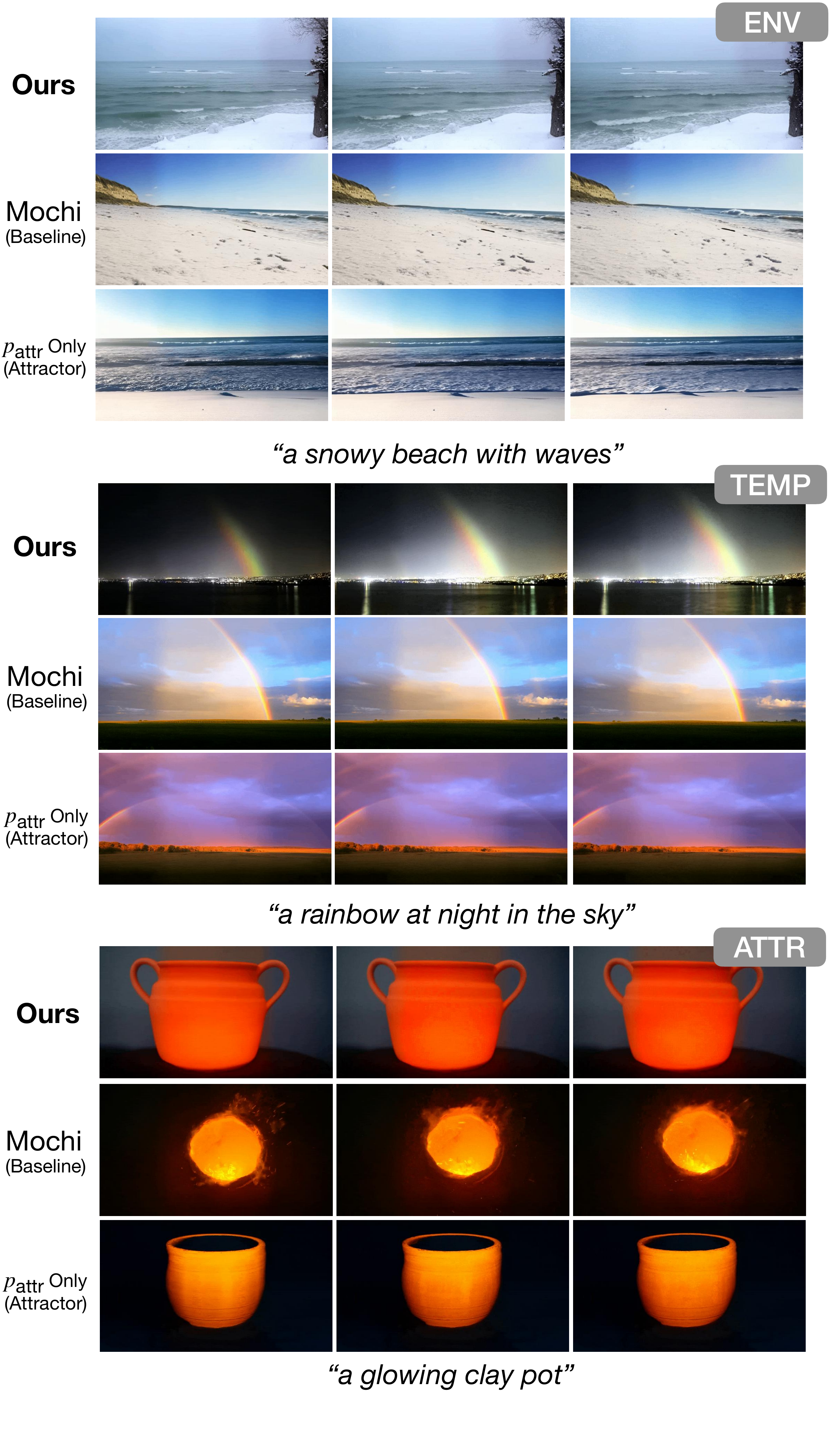}
    \caption{\textbf{Attractor trajectory visualization across three compositional categories.} For each prompt, we compare videos generated by Ours (DCR), Mochi (standard CFG baseline), and $p_{\text{attr}}$ only (standard CFG applied to the attractor prompt). In all three cases, the $p_{\text{attr}}$-only output reflects the model's default completion tendency (a sunny beach instead of a snowy one in ENV, a daytime rainbow instead of a nocturnal one in TEMP, and a dimly lit clay pot instead of a glowing one in ATTR) and closely resembles the Mochi baseline output. This confirms that $p_{\text{attr}}$ serves as an accurate proxy for the model's implicit attractor, and that DCR effectively suppresses this tendency to generate rare but semantically valid compositions.}
    \label{fig:traj_supp}
\end{figure*}

\section{Attractor Trajectory Visualization} 
\label{supp:attractor_visualization} 
To validate that the attractor prompt $p_{\text{attr}}$ accurately captures the model's default completion tendency, we compare three generation conditions for each prompt: our full method (DCR), the Mochi baseline (standard CFG on prompt $p$), and $p_{\text{attr}}$ only (standard CFG on the attractor prompt alone, without repulsion). If $p_{\text{attr}}$ correctly represents the model's implicit attractor, then the $p_{\text{attr}}$-only output should closely resemble the Mochi baseline output, both reflecting the frequent completion that DCR is designed to suppress. Figure~\ref{fig:traj_supp} presents this comparison across three compositional categories. In ENV (\emph{a snowy beach with waves}), both Mochi and the $p_{\text{attr}}$-only condition generate a snow-free sunny beach, while DCR produces a beach with visible snow and waves. In TEMP (\emph{a rainbow at night in the sky}), Mochi generates a daytime or twilight rainbow, and the $p_{\text{attr}}$-only condition produces a similar warm-toned twilight scene, whereas DCR generates a coherent nocturnal scene with a visible rainbow. In ATTR (\emph{a glowing clay pot}), Mochi collapses to a glowing spherical object, while the $p_{\text{attr}}$-only condition produces a clay pot under dimmer lighting, closer to the intended relaxed completion. DCR correctly realizes the glowing attribute while preserving the clay pot form. Across all three categories, the $p_{\text{attr}}$-only outputs consistently reflect the model's high-frequency completion preference and closely resemble the Mochi baseline, confirming that $p_{\text{attr}}$ serves as an accurate proxy for the implicit attractor. This validates the core motivation of DCR: by explicitly modeling the attractor direction through $p_{\text{attr}}$, the method can apply targeted repulsion to prevent compositional collapse without degrading overall generation quality.

\section{Attractor Prompt Generation Details} 
\label{supp:attractor_prompts} 

For each target prompt $p$, we generate the attractor prompt $p_{\text{attr}}$ using a large language model (GPT-4o) via a structured rewriting instruction. The transformation is designed to remove or weaken the rare compositional factor while preserving the surrounding semantic context, producing a prompt that reflects the model's most probable default completion. \paragraph{Prompting Template.} We use the following fixed instruction template: 

\begin{quote} 
\texttt{Given the following text prompt describing a rare but plausible visual composition, generate a single alternative prompt that represents the most common or frequently occurring version of the same scene. Remove or replace only the rare compositional factor (e.g., unusual attribute, atypical environment, or unlikely object placement) while preserving all other scene elements. Output only the rewritten prompt, with no explanation.}

\texttt{Input prompt: \{p\}}
\end{quote}

All attractor prompts are generated with temperature set to 0 for deterministic output. Manual review is conducted to verify that each $p_{\text{attr}}$ (i) removes the rare factor, (ii) preserves surrounding scene semantics, and (iii) does not introduce new compositional elements absent from the original prompt.

\paragraph{Examples.} Table~\ref{tab:pattr_examples} lists representative $(p, p_{\text{attr}})$ pairs across all eight benchmark categories.

\begin{table*}[t]
\centering
\small
\begin{tabular}{l p{6.2cm} p{6.2cm}}
\toprule
\textbf{Category} & \textbf{Prompt $p$} &
\textbf{Attractor $p_{\text{attr}}$} \\
\midrule
ENV  & a snowy beach with waves
     & a tropical beach with waves \\
ENV  & a desert oasis surrounded by snow
     & a desert oasis surrounded by sand dunes \\
TEMP & a rainbow at night in the sky
     & a rainbow during the day in the sky \\
TEMP & a sunrise with stars still visible in the sky
     & a clear sunrise sky \\
OBJ  & a lighthouse located in a grassy meadow
     & a lighthouse on a rocky coast \\
OBJ  & a canoe placed in a dry canyon
     & a canoe on a calm river \\
ATTR & a glowing clay pot
     & a clay pot under natural lighting \\
ATTR & a reflective wooden table like a mirror
     & a wooden table under natural lighting \\
SCALE & a giant cat larger than a building
      & a cat near a building \\
SCALE & a tiny bench on a coin
      & a bench on the ground \\
CTX  & a kitchen in the middle of a highway
     & a kitchen inside a house \\
CTX  & a library on a beach
     & a library inside a building \\
MAT  & a melting ice sculpture in a snowy field
     & an ice sculpture in a snowy field \\
MAT  & a wet fire burning with visible flames
     & a fire burning with visible flames \\
DENS & a lightly crowded subway platform during peak hours
     & a crowded subway platform during peak hours \\
DENS & a lightly crowded shopping street during sales season
     & a busy shopping street during sales season \\
\bottomrule
\end{tabular}
\caption{\textbf{Representative $(p, p_{\text{attr}})$ pairs across all eight benchmark categories.} Each attractor prompt removes or replaces the rare compositional factor while preserving the surrounding scene context.}
\label{tab:pattr_examples}
\end{table*}

\section{Guidance Geometry Diagnostics Details}
\label{supp:diagnostics}
This appendix details the logging protocol behind Sec.~\ref{sec:guidance_diag} and reports the per-category statistics. During generation we record, at every denoising step and without altering the update, the alignment $\cos\theta_t$ between the drift $a_t$ and the reference update $\Delta_{\text{ref}}$, the gate indicator $\mathbf{1}[s_t > 0]$, the removed fraction $\lambda_t \lVert a_t \rVert / \lVert \Delta_{\text{ref}} \rVert$, the alignment $\cos\varphi_t$ between the raw guidance directions $d_{\text{attr}} = \epsilon_{\text{attr}} - \epsilon_{\text{uncond}}$ and $d_{\text{text}} = \epsilon_{\text{text}} - \epsilon_{\text{uncond}}$, and their norm ratio $\rho_t = \lVert d_{\text{attr}} \rVert / \lVert d_{\text{text}} \rVert$. Expanding $s_t = \langle a_t, \Delta_{\text{ref}} \rangle$ with $a_t = w_{\text{attr}} d_{\text{attr}} - w\, d_{\text{text}}$ gives the equivalence
\begin{equation}
s_t > 0 \;\Longleftrightarrow\; \rho_t \cos\varphi_t > \frac{w}{w_{\text{attr}}},
\end{equation}
so gating is decided by the joint per-step behavior of $\rho_t$ and $\cos\varphi_t$. Because the two quantities co-vary strongly across steps, window means of $\rho_t$ and $\cos\varphi_t$ do not determine the gate rate on their own, which is why Table~\ref{tab:diag_category} reports all quantities separately. All statistics are computed over steps inside the scheduling window $[r_s, r_e] = [0.2, 0.8]$ only; outside the window $\alpha_t = 0$ and the projection has no effect regardless of the gate. The removed fraction is reported in two forms, averaged over gate-active steps only (Rem.) and averaged over all window steps with inactive steps counted as zero (Rem-eff.), the latter giving the unconditional effective removal.

Table~\ref{tab:diag_category} breaks the statistics down by category for both regenerated groups. Two observations stand out. First, the guidance-branch alignment varies widely across categories, from $\cos\varphi \approx 0.32$ in CTX to $\approx 0.86$ in MAT. Categories with low alignment and a high norm ratio, such as CTX, are those whose attractor prompt describes a scene that differs structurally from the target, and these are also the categories where over-suppression dominates the failure modes in Appendix~\ref{supp:failure_analysis}, consistent with a drift direction that overlaps the shared scene content. Second, within every category the success and failure groups show similar gate rates and removal, confirming at the category level what Sec.~\ref{sec:guidance_diag} reports in aggregate, that residual failures are not explained by the mechanism failing to engage.

\begin{table}[t]
\begin{small}
    \centering
    \resizebox{0.5\textwidth}{!}{%
    \begin{tabular}{llcccccc}
        \toprule
        \textbf{Group} & \textbf{Cat.} & $\cos\theta$ & \textbf{Gate\%} & \textbf{Rem.\%} & \textbf{Rem-eff.\%} & $\cos\varphi$ & $\rho$ \\
        \midrule
        \multirow{9}{*}{\shortstack[l]{Success\\(40 clips)}}
        & ENV   & 0.139  & 69.5 & 12.2 & 7.6  & 0.684 & 1.20 \\
        & TEMP  & 0.106  & 71.6 & 10.4 & 7.3  & 0.670 & 1.12 \\
        & OBJ   & -0.105 & 47.9 & 11.5 & 5.6  & 0.409 & 1.25 \\
        & ATTR  & 0.358  & 83.2 & 15.1 & 12.7 & 0.778 & 1.18 \\
        & SCALE & -0.049 & 51.1 & 16.2 & 7.0  & 0.460 & 1.27 \\
        & CTX   & -0.104 & 40.5 & 18.9 & 7.2  & 0.359 & 1.41 \\
        & MAT   & 0.255  & 74.2 & 18.2 & 13.3 & 0.785 & 1.06 \\
        & DENS  & 0.136  & 73.7 & 13.9 & 9.7  & 0.788 & 0.98 \\
        & All   & 0.092  & 63.9 & 14.5 & 8.8  & 0.617 & 1.18 \\
        \midrule
        \multirow{9}{*}{\shortstack[l]{Failure\\(124 clips)}}
        & ENV   & 0.157  & 67.3 & 15.8 & 10.0 & 0.628 & 1.29 \\
        & TEMP  & 0.113  & 64.1 & 12.3 & 7.4  & 0.706 & 1.03 \\
        & OBJ   & -0.036 & 47.2 & 18.8 & 8.2  & 0.512 & 1.20 \\
        & ATTR  & 0.223  & 70.8 & 13.1 & 9.2  & 0.720 & 1.19 \\
        & SCALE & -0.062 & 48.7 & 14.1 & 5.9  & 0.527 & 1.10 \\
        & CTX   & -0.143 & 37.9 & 17.6 & 6.2  & 0.319 & 1.61 \\
        & MAT   & 0.436  & 88.9 & 16.8 & 15.1 & 0.863 & 1.08 \\
        & DENS  & 0.344  & 86.5 & 15.1 & 12.4 & 0.834 & 1.07 \\
        & All   & 0.127  & 63.9 & 15.1 & 9.1  & 0.642 & 1.19 \\
        \bottomrule
    \end{tabular}}
    \caption{\textbf{Per-category guidance geometry.} Window-only means per regenerated group. $\cos\theta$, alignment between the drift $a_t$ and $\Delta_{\text{ref}}$; Gate\%, share of window steps with $s_t > 0$; Rem.\%, removed fraction of $\Delta_{\text{ref}}$ over gate-active steps; Rem-eff.\%, the same fraction over all window steps; $\cos\varphi$, alignment between the two guidance directions; $\rho$, their norm ratio.}
    \label{tab:diag_category}
\end{small}
\end{table}

\section{Residual Failure Analysis}
\label{supp:failure_analysis}
DCR lowers the violation rate but does not eliminate it, and this appendix asks why the remaining failures occur. We regenerated the 124 prompts that the judge scored as violations in the main evaluation, one clip each under the default configuration, with the diagnostics of Appendix~\ref{supp:diagnostics} logged during generation. The authors then watched every regenerated clip and recorded binary annotations, whether the clip still violates the composition, whether the rare factor is visible anywhere and in each temporal third, whether the remaining non-rare content is visible, whether the clip shows the typical completion of the attractor prompt, and, where it does not, whether the completion it shows instead matches the output of the CFG baseline for the same prompt. Of the 124 regenerated clips, 72 still violate; this share is not comparable to the CVR in Table~\ref{tab:dcr_main} because the regenerated set deliberately contains only previously failing prompts.

Each violating clip is assigned to the first matching rule in a fixed order. (1)~\emph{Weak intervention}, the in-window gate rate is below $\tau = 0.25$ or the effective removed fraction is below $\tau' = 0.05$, so the projection barely acted. (2)~\emph{Over-suppression}, the shared scene content is gone, whether or not the rare factor remains. (3)~\emph{Attractor mismatch}, the clip collapsed to a typical scene that differs from the attractor completion, so $p_{\text{attr}}$ missed the model's actual default. (4)~\emph{Temporal collapse}, the rare factor is present in some thirds of the clip but not all. (5)~\emph{Insufficient repulsion}, the projection was active yet the clip still shows the attractor completion without the rare factor. (6)~\emph{Other}, both factors are visible but the composition is judged incoherent. Table~\ref{tab:failure_taxonomy} reports the result.

\begin{table}[t]
\begin{small}
    \centering
    \resizebox{0.5\textwidth}{!}{%
    \begin{tabular}{lccccccc}
        \toprule
        \textbf{Cat.} & \textbf{Total} & \textbf{Over-sup.} & \textbf{Insuf.} & \textbf{Weak} & \textbf{Attr.} & \textbf{Temp.} & \textbf{Other} \\
        \midrule
        ENV   & 9  & 4  & 3  & 1  & 0 & 0 & 1 \\
        TEMP  & 14 & 1  & 2  & 8  & 1 & 1 & 1 \\
        OBJ   & 2  & 0  & 0  & 0  & 1 & 0 & 1 \\
        ATTR  & 17 & 8  & 3  & 3  & 3 & 0 & 0 \\
        SCALE & 7  & 0  & 2  & 3  & 1 & 0 & 1 \\
        CTX   & 10 & 7  & 0  & 2  & 0 & 0 & 1 \\
        MAT   & 5  & 1  & 2  & 0  & 2 & 0 & 0 \\
        DENS  & 8  & 0  & 6  & 0  & 1 & 0 & 1 \\
        \midrule
        All   & 72 & 21 & 18 & 17 & 9 & 1 & 6 \\
        \bottomrule
    \end{tabular}}
    \caption{\textbf{Residual failure taxonomy.} Primary assignment of the 72 still-violating regenerated clips, first matching rule in the order weak intervention, over-suppression, attractor mismatch, temporal collapse, insufficient repulsion, other, with $\tau = 0.25$ on the gate rate and $\tau' = 0.05$ on the effective removed fraction. Column order follows overall frequency.}
    \label{tab:failure_taxonomy}
\end{small}
\end{table}

No single cause dominates, and the four main modes carry comparable weight. \emph{Over-suppression} (21 of 72) concentrates in ATTR and CTX, the categories whose attractor prompts differ structurally from the target scene and whose guidance branches are least aligned (Table~\ref{tab:diag_category}), so the drift direction there overlaps the shared content and projecting it away removes the scene itself. \emph{Insufficient repulsion} (18) is the mirror case, the gate was open and the projection acted, yet the clip still shows the attractor completion; these clips removed only $13.2\%$ of $\Delta_{\text{ref}}$ on average, which suggests headroom in the repulsion strength $\eta$ for hard prompts. \emph{Weak intervention} (17) is driven almost entirely by the removal criterion rather than the gate criterion, 16 of the 17 clips fall below $\tau'$ while only 4 fall below $\tau$, so the typical pattern is a gate that opens on steps where there is little to remove. \emph{Attractor mismatch} is the primary label for 9 clips, but the annotation records a completion different from the attractor prompt in 42 of the 72 failures overall, many of which the earlier rules absorb, so the quality of the LLM-constructed attractor remains a first-order limitation even where it is not the primary label. A direct comparison with the CFG baseline output was recorded for 29 of these 42 clips, and in none of them does the failure completion match the baseline's default completion either. The diverted trajectory therefore tends to settle in a third mode, one captured neither by the attractor prompt nor by the default the baseline realizes, rather than remaining in the default basin. \emph{Temporal collapse} is nearly absent (1 clip), indicating that when the rare factor is realized at all it tends to persist across the clip. Because the rule order can shift counts between adjacent modes, we also checked each condition independently, and we report the sensitivity of the dominant threshold, the weak-intervention count takes the values 16, 17, and 23 as $\tau$ varies over $\{0.15, 0.25, 0.35\}$ at fixed $\tau'$, leaving the ordering of the modes unchanged.

\section{Backbone Choice and Method Generality} 
\label{supp:backbone} 
\paragraph{Scope.} 
The scientific contribution of this paper is the identification of \emph{default completion bias} as a generative failure mode and the introduction of a projection-based repulsion mechanism that suppresses it at inference time. Establishing this contribution requires (i) demonstrating that the failure mode exists in modern text-to-video diffusion models and (ii) demonstrating that projection-based repulsion suppresses it in a controlled setting where each component of the method can be cleanly isolated. We address (i) across three backbones and (ii) on Mochi as a controlled testbed. Cross-backbone \emph{calibration} of repulsion hyperparameters is a complementary engineering question and is discussed as future work. 

\paragraph{The target failure mode is empirically backbone-agnostic.} 
A central premise of DCR is that compositional collapse is not a quirk of any single model but a structural tendency of large-scale text-to-video diffusion systems. This is supported empirically by Sec.~\ref{sec:qualitative} and Appendix~\ref{sec:detailed_qualitative}: across all eight categories of DCR-Bench, Mochi, HunyuanVideo, and CogVideoX exhibit consistent collapse toward statistically dominant completions: daytime rainbows instead of nocturnal ones, snow-free beaches instead of snowy ones, glowing spheres instead of glowing clay pots. These three backbones differ in architecture (DiT-based vs. transformer-based), text encoder (T5 vs. LLM-based), scheduler family (flow-matching vs. v-prediction), and training data. The fact that all three exhibit the same failure pattern establishes that default completion bias is a property of the model \emph{class} rather than of any specific backbone. The need for an explicit suppression mechanism is therefore established at the level of the class.

\paragraph{DCR's formulation is backbone-invariant by construction.} 
Given any diffusion model that supports classifier-free guidance with free-form text conditioning, DCR is fully specified by three operations: (i) three denoiser forward passes on the existing model, conditioned respectively on $\varnothing$, $p$, and $p_{\text{attr}}$, requiring no modification to the transformer, scheduler, or text encoder; (ii) two linear combinations in noise-prediction space, $\Delta_{\text{ref}} = w(\epsilon_{\text{text}} - \epsilon_{\text{uncond}})$ and $\epsilon_{\text{probe}} = \epsilon_{\text{uncond}} + w_{\text{attr}}(\epsilon_{\text{attr}} - \epsilon_{\text{uncond}})$; and (iii) one projection $\Delta_t^\star = \Delta_{\text{ref}} - \lambda_t a_t$, computed via an inner product and a norm over flattened latent tensors. None of these operations references the architecture, the scheduler family, the parameterization (noise-prediction vs.\ flow), or the text encoder of the underlying model. The formulation is therefore backbone-invariant in the same sense that classifier-free guidance~\cite{ho2022classifier} itself is.

\paragraph{What varies across backbones is calibration, not mechanism.} 
The only quantity DCR inherits from each backbone is the \emph{magnitude scale} of $\epsilon_{\text{text}} - \epsilon_{\text{uncond}}$, which determines appropriate values of $(w, w_{\text{attr}}, \eta)$. Different backbones already use different default guidance scales for this reason, independently of DCR. Calibrating these hyperparameters per backbone is therefore a per-backbone hyperparameter exercise of the same kind that any guidance-based method inherits, not a test of whether the repulsion mechanism itself generalizes. The sensitivity analysis in Sec.~\ref{sec:hyperparam_sensitivity} shows that DCR is robust within a broad range of these values on Mochi; transferring to a new backbone nonetheless requires recalibrating them, and we do not claim that the Mochi settings carry over directly. 

\paragraph{Why Mochi specifically.} 
Within this scope, we select Mochi as the primary testbed for three reasons. First, it exposes a clean classifier-free guidance interface through the standard \texttt{diffusers} library, allowing the three-branch forward pass to be implemented without modifying any internal component. Second, its T5-based conditioning and flow-matching scheduler represent a widely adopted configuration in recent open-source text-to-video systems, making it a representative rather than idiosyncratic choice. Third, its publicly available, self-contained inference pipeline ensures that all experiments are fully reproducible without proprietary components. These properties make Mochi a controlled environment for isolating the effect of the proposed mechanism via the ablations in Table~\ref{tab:dcr_main}. 

\paragraph{Future work.} 
A systematic multi-backbone quantitative study, including HunyuanVideo, CogVideoX, and image-domain backbones such as SDXL and SD3, would require per-backbone calibration of $(w, w_{\text{attr}}, \eta, r_s, r_e)$ alongside backbone-specific evaluation protocols. We view this as a natural and important extension that builds on the mechanism established here, and leave it to future work.

\section{Computation Time and Memory Consumption}
\label{supp:computation_text}

\begin{table}[htb!]
    \small
    \centering
    \resizebox{0.5\textwidth}{!}{
    \begin{tabular}{l|l|c|c}
    \toprule
    \textbf{Model} & \textbf{Step} &
    \textbf{Memory Consumption} &
    \textbf{Computation Time} \\
    \midrule
    \textbf{Ours} & Inference & 26.3096 GB / 80.0000 GB (26,941 MiB) & 192 sec \\
    \midrule
    Ours w/o Attractor Prompt & Inference & 25.0264 GB / 80.0000 GB (25,627 MiB) & 190 sec \\
    Ours w/o Repulsion & Inference & 25.4287 GB / 80.0000 GB (26,039 MiB) & 176 sec \\
    Ours w/o Schedule & Inference & 25.0264 GB / 80.0000 GB (25,627 MiB) & 187 sec \\
    \midrule
    Negative Prompt & Inference & 26.3096 GB / 80.0000 GB (26,941 MiB) & 174 sec \\
    Mochi & Inference & 24.4385 GB / 80.0000 GB (25,025 MiB) & 114 sec \\
    HunyuanVideo & Inference & 38.5186 GB / 80.0000 GB (39,443 MiB) & 396 sec \\
    CogVideoX & Inference & 15.6318 GB / 80.0000 GB (16,007 MiB) & 61 sec \\
    \bottomrule
    \end{tabular}
    }
    \caption{\textbf{Inference-time memory and runtime on a single NVIDIA H100 (80GB).} Peak GPU memory and end-to-end inference time are reported for our full method, ablation variants, and all baselines. All methods share the same backbone (Mochi) except HunyuanVideo and CogVideoX. All measurements are conducted under identical hardware and evaluation conditions. GB values are binary gigabytes (MiB / 1024) as reported by \texttt{nvidia-smi}.}
    \label{tab:computation}
\end{table}

Table~\ref{tab:computation} reports peak GPU memory and end-to-end inference time on a single NVIDIA H100 (80GB) for all evaluated methods.

\paragraph{Computational overhead of DCR.}
Our full method requires three denoiser forward passes per step: unconditional, text-conditioned, and attractor-conditioned. Compared to Mochi (standard two-branch CFG, 114 sec), DCR adds approximately 78 sec (+68\%), reflecting the cost of the additional attractor branch forward pass. Peak memory increases from 24.44 GB (Mochi) to 26.31 GB (Ours), a moderate +1.87 GB corresponding to the activation footprint of the additional attractor branch, as the three branches share the same model weights. The projection-based repulsion contributes negligible overhead beyond the forward pass, as it requires only a scalar inner-product and norm computation over flattened latent tensors.

\paragraph{Ablation variants.}
The w/o Attractor Prompt (190 sec, 25.03 GB) and w/o Repulsion (176 sec, 25.43 GB) variants still evaluate all three branches even though their final updates are identical to standard CFG (Sec.~\ref{subsec:projection_main}); their timings therefore isolate the cost of the additional branch and of the per-step correction arithmetic, independently of any effect on the output. Removing the schedule (187 sec, 25.03 GB) applies repulsion uniformly across all steps; runtime is comparable to the full model since the active interval $[r_s, r_e] = [0.2, 0.8]$ already covers the majority of denoising steps. We also note that the attractor branch influences the update only inside the active interval, so skipping its evaluation outside $[r_s, r_e]$ would reduce the overhead further; the timings reported here are unoptimized.

\paragraph{Comparison across backbones.}
Negative Prompt uses the same three-branch structure as DCR but applies the attractor branch as a standard CFG negative input, resulting in 174 sec and the same peak memory as DCR (26.31 GB), consistent with the shared three-branch structure. Among the backbone baselines, HunyuanVideo requires the most resources at 38.52 GB and 396 sec, while CogVideoX is the most efficient at 15.63 GB and 61 sec. Our method's overhead relative to Mochi is modest and represents a practical trade-off for improved compositional fidelity.

\newpage

\begin{figure*}[t]
    \centering
    \includegraphics[width=0.95\linewidth]{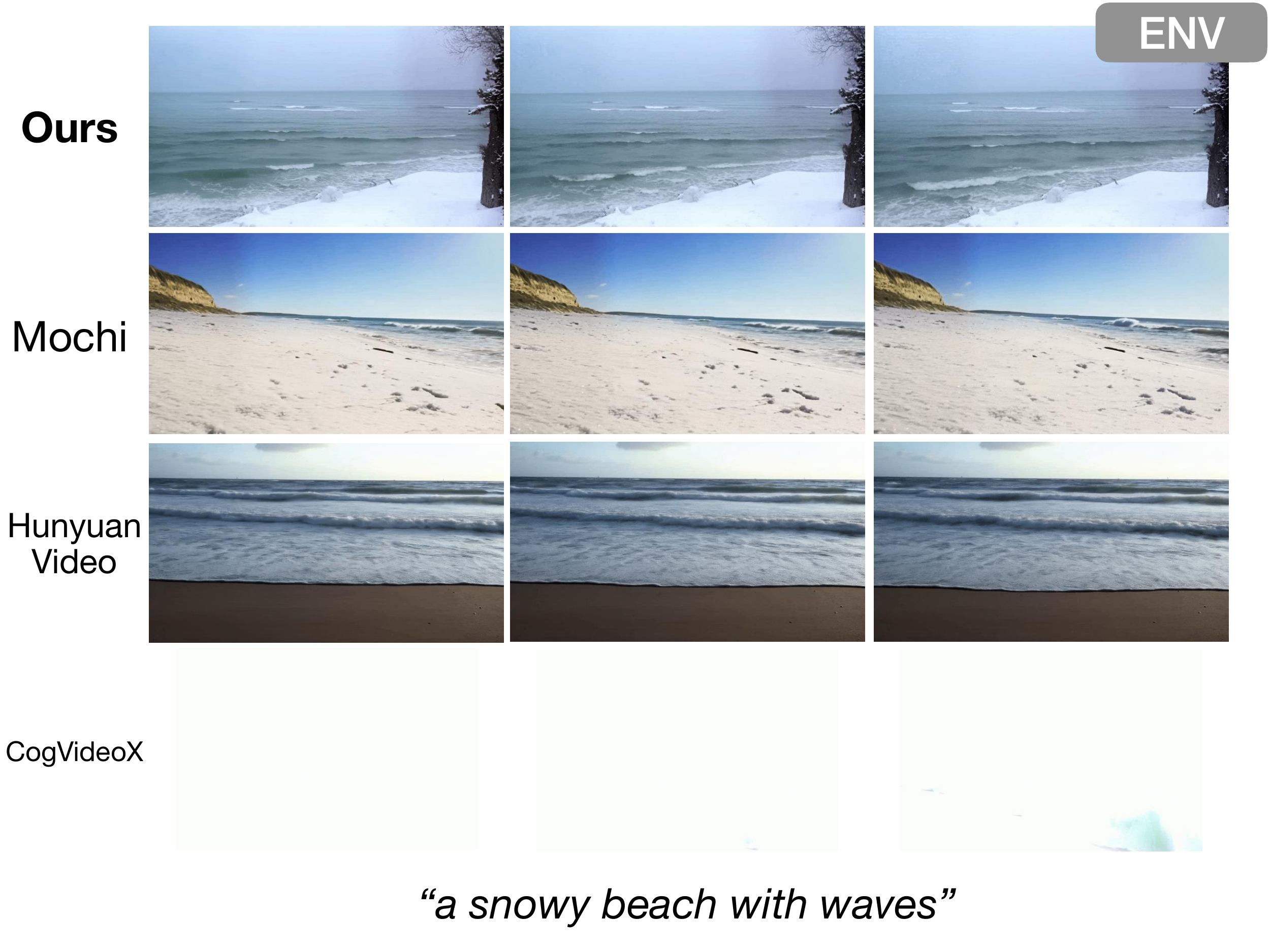}
    \caption{\textbf{ENV (Environment Recomposition).} 
    Prompt: ``a snowy beach with waves.'' 
    Our method preserves both snow and beach simultaneously, while baselines collapse to a standard beach.}
    \label{fig:qual_env}
\end{figure*}

\begin{figure*}[t]
    \centering
    \includegraphics[width=0.95\linewidth]{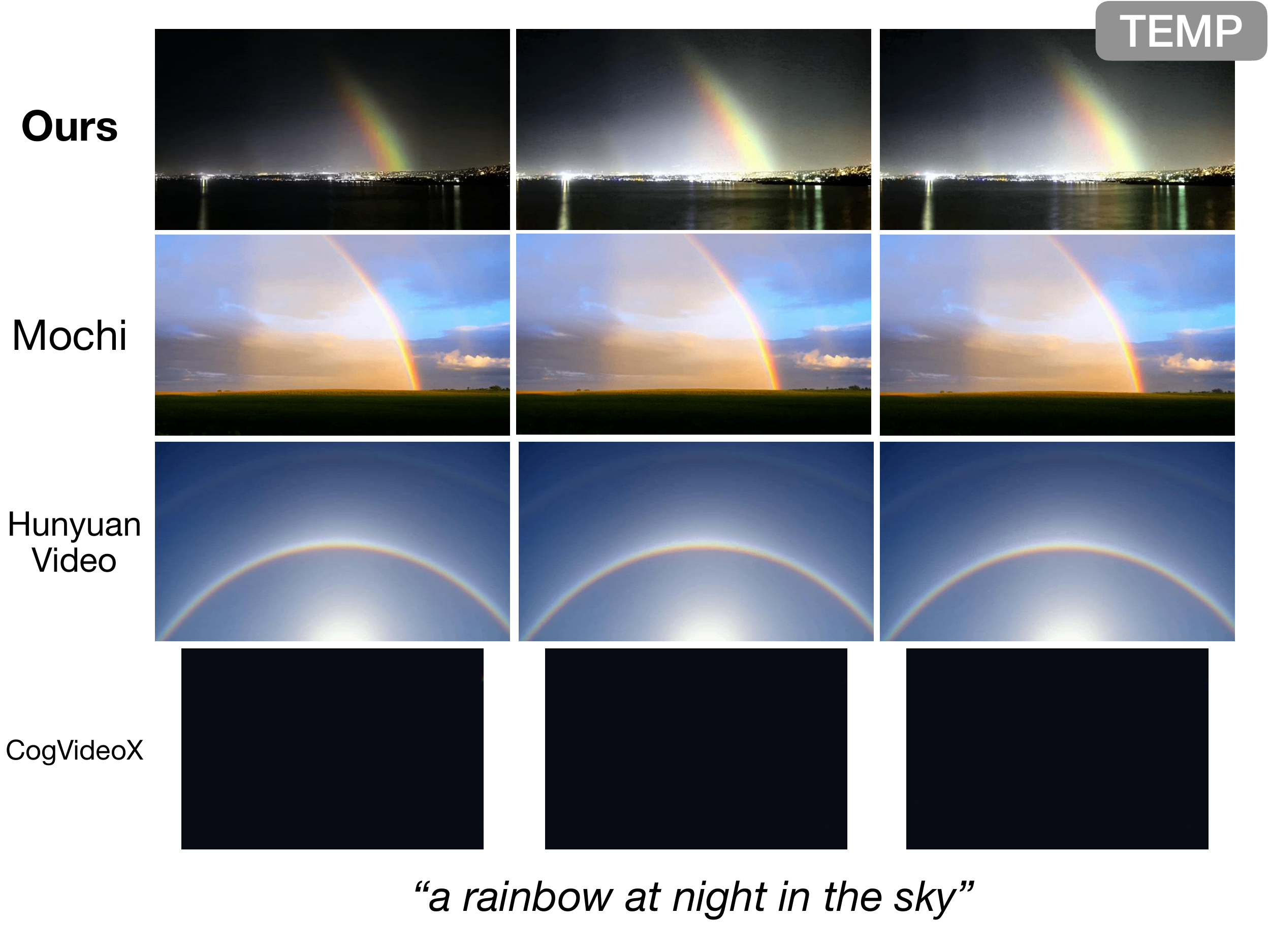}
    \caption{\textbf{TEMP (Temporal Misalignment).} 
    Prompt: ``a rainbow at night in the sky.'' 
    Our method generates a nocturnal rainbow, whereas baselines revert to daytime scenes.}
    \label{fig:qual_temp}
\end{figure*}

\begin{figure*}[t]
    \centering
    \includegraphics[width=0.95\linewidth]{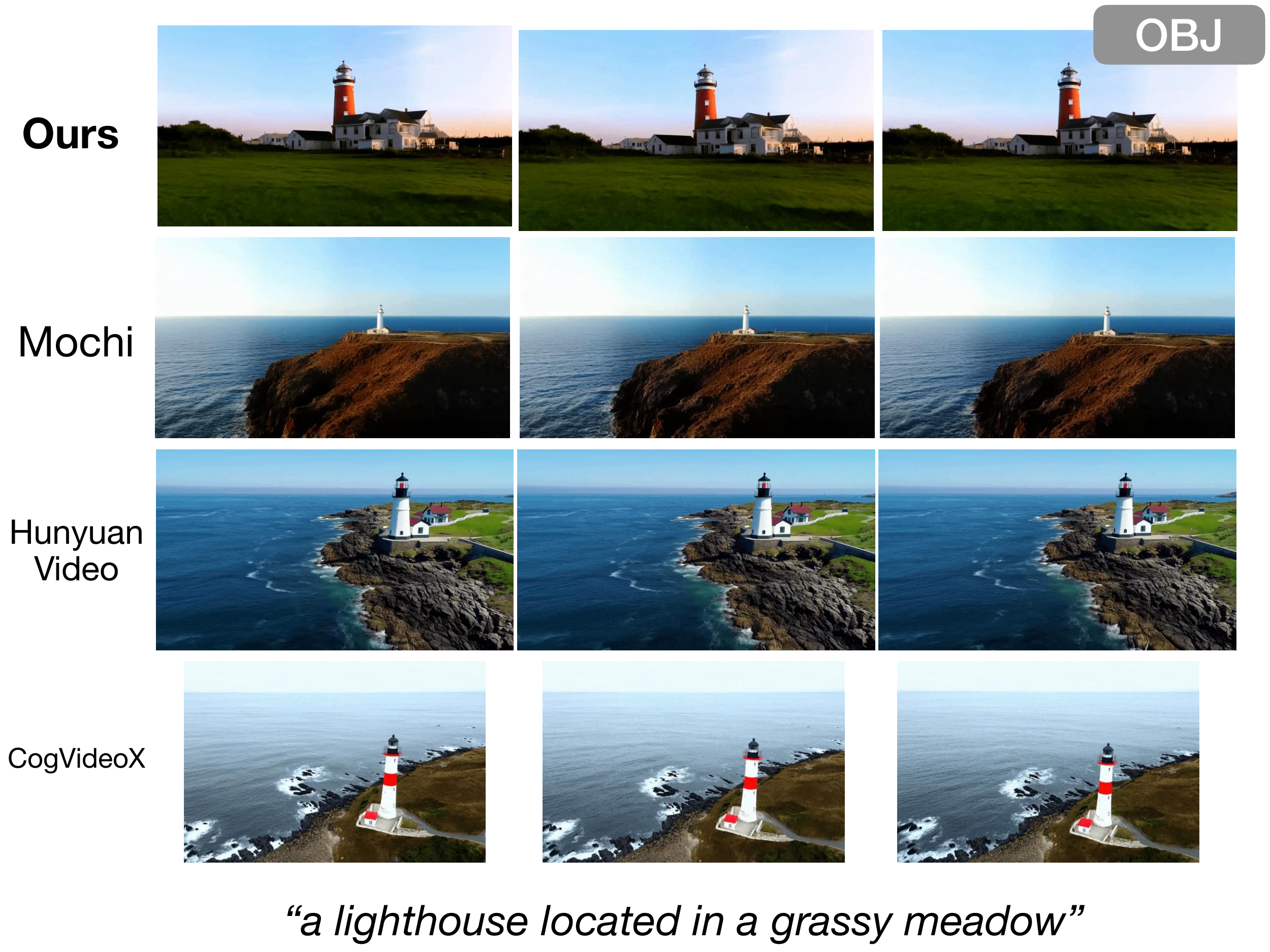}
    \caption{\textbf{OBJ (Object Relocation).} 
    Prompt: ``a lighthouse located in a grassy meadow.'' 
    Our method maintains the rare object-environment pairing, while baselines relocate the object to coastal regions.}
    \label{fig:qual_obj}
\end{figure*}

\begin{figure*}[t]
    \centering
    \includegraphics[width=0.95\linewidth]{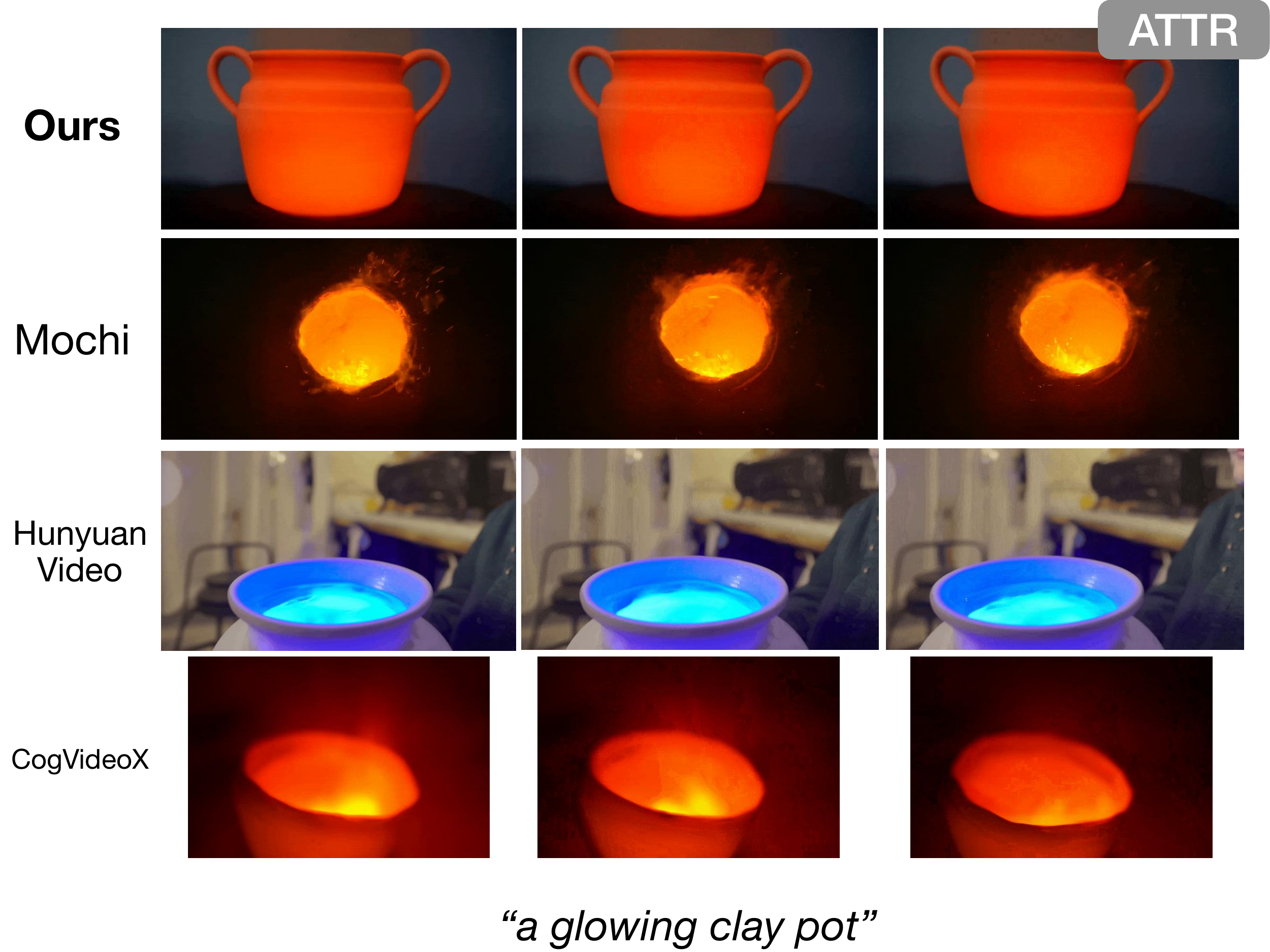}
    \caption{\textbf{ATTR (Attribute Rebinding).} 
    Prompt: ``a glowing clay pot.'' 
    Our method correctly binds the glowing attribute to the clay material, while baselines alter object identity or material.}
    \label{fig:qual_attr}
\end{figure*}

\begin{figure*}[t]
    \centering
    \includegraphics[width=0.95\linewidth]{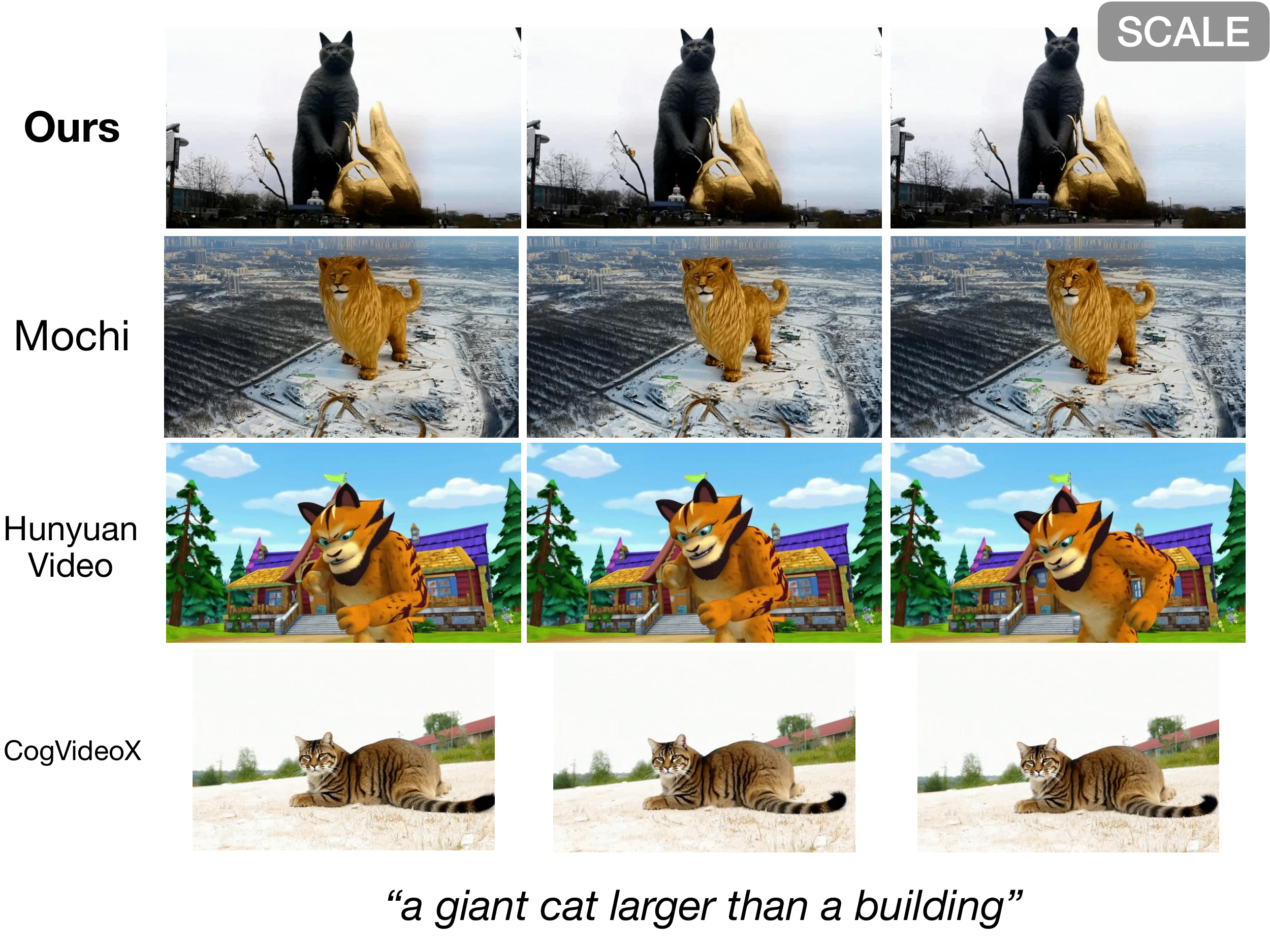}
    \caption{\textbf{SCALE (Scale Shift).} 
    Prompt: ``a giant cat larger than a building.'' 
    Our method realizes extreme scale relationships, while baselines revert to typical object sizes.}
    \label{fig:qual_scale}
\end{figure*}

\begin{figure*}[t]
    \centering
    \includegraphics[width=0.95\linewidth]{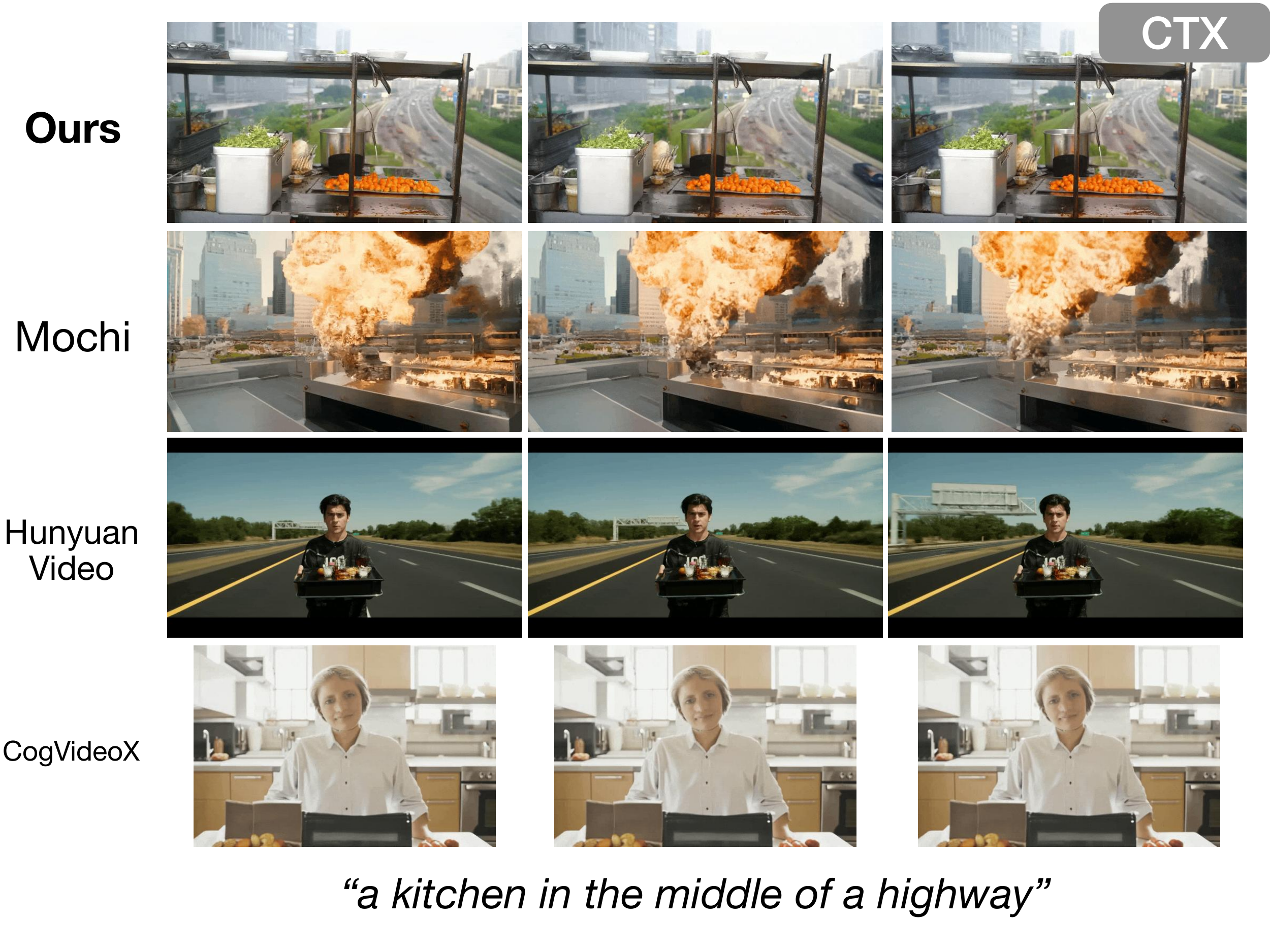}
    \caption{\textbf{CTX (Contextual Relocation).} 
    Prompt: ``a kitchen in the middle of a highway.'' 
    Our method integrates both contexts coherently, while baselines generate only one dominant scene.}
    \label{fig:qual_ctx}
\end{figure*}

\begin{figure*}[t]
    \centering
    \includegraphics[width=0.95\linewidth]{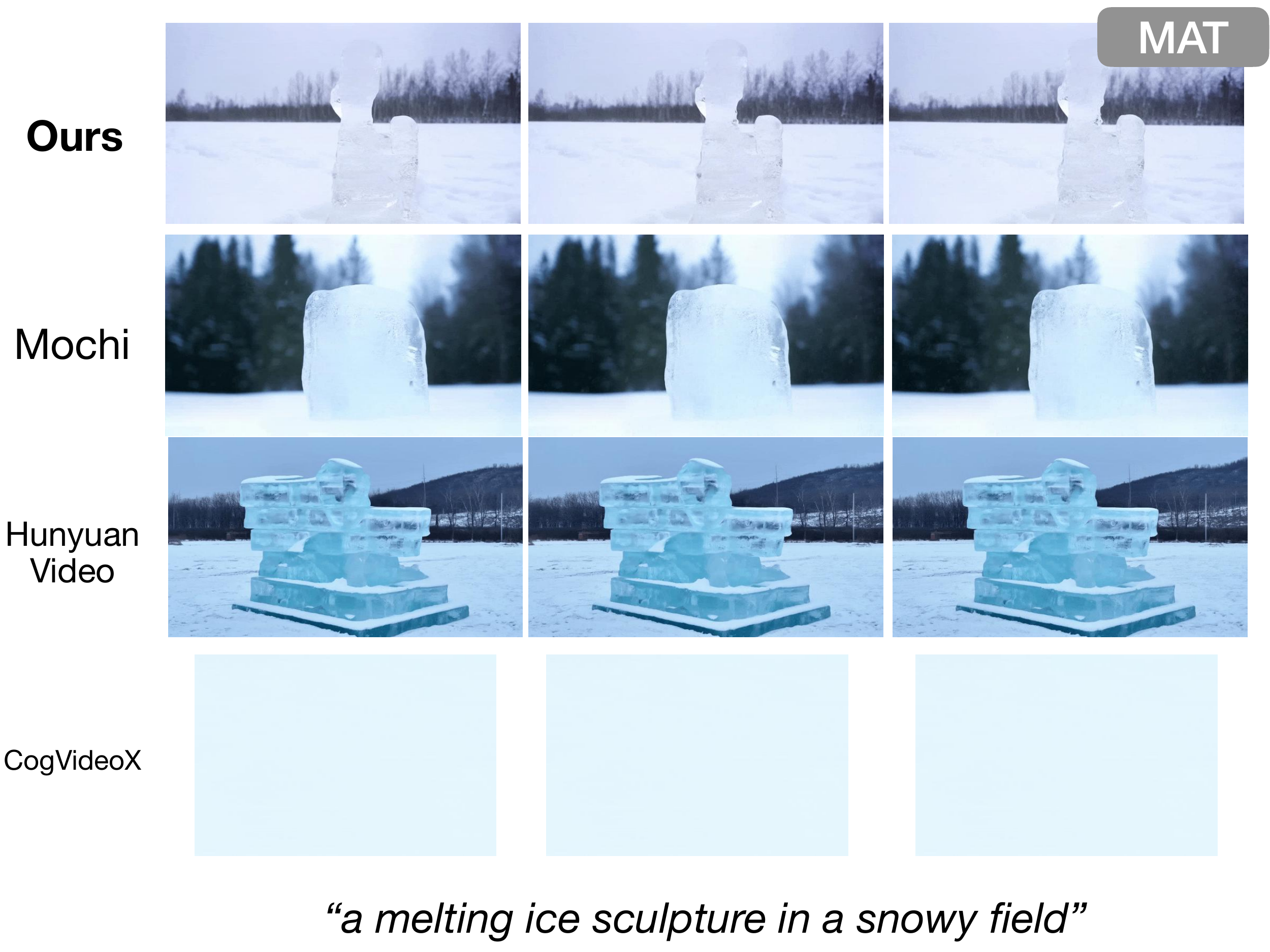}
    \caption{\textbf{MAT (Material-State Conflict).} 
    Prompt: ``a melting ice sculpture in a snowy field.'' 
    Our method preserves both melting and snowy conditions, while baselines collapse to a single state.}
    \label{fig:qual_mat}
\end{figure*}

\begin{figure*}[t]
    \centering
    \includegraphics[width=0.95\linewidth]{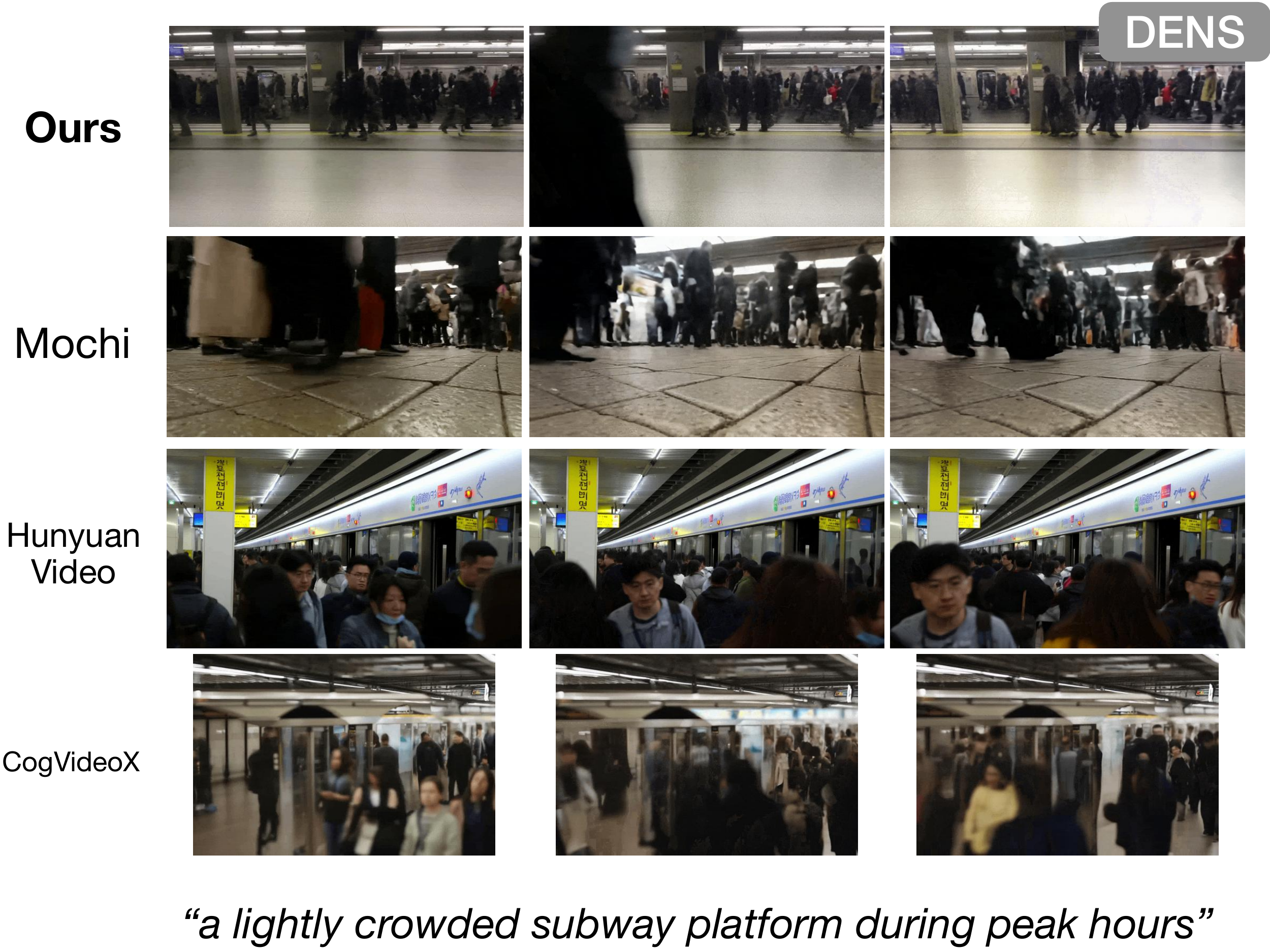}
    \caption{\textbf{DENS (Density Variation).} 
    Prompt: ``a lightly crowded subway platform during peak hours.'' 
    Our method maintains low density despite strong priors for crowded scenes.}
    \label{fig:qual_dens}
\end{figure*}

\begin{figure*}[t]
\centering
\includegraphics[width=0.95\linewidth]{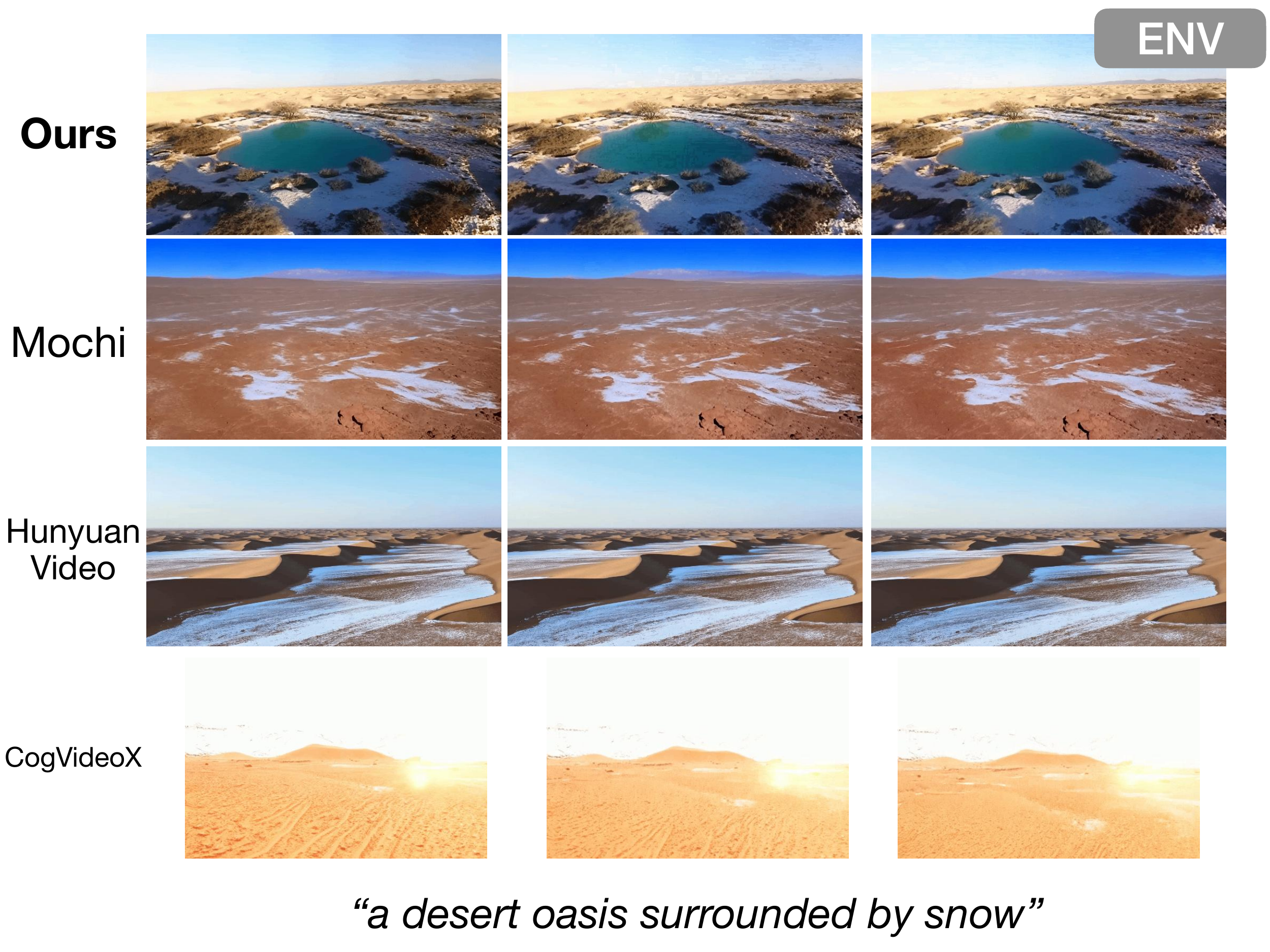}
\caption{\textbf{ENV.} Prompt: ``a desert oasis surrounded by snow.''}
\label{fig:supp_env_1}
\end{figure*}

\begin{figure*}[t]
\centering
\includegraphics[width=0.95\linewidth]{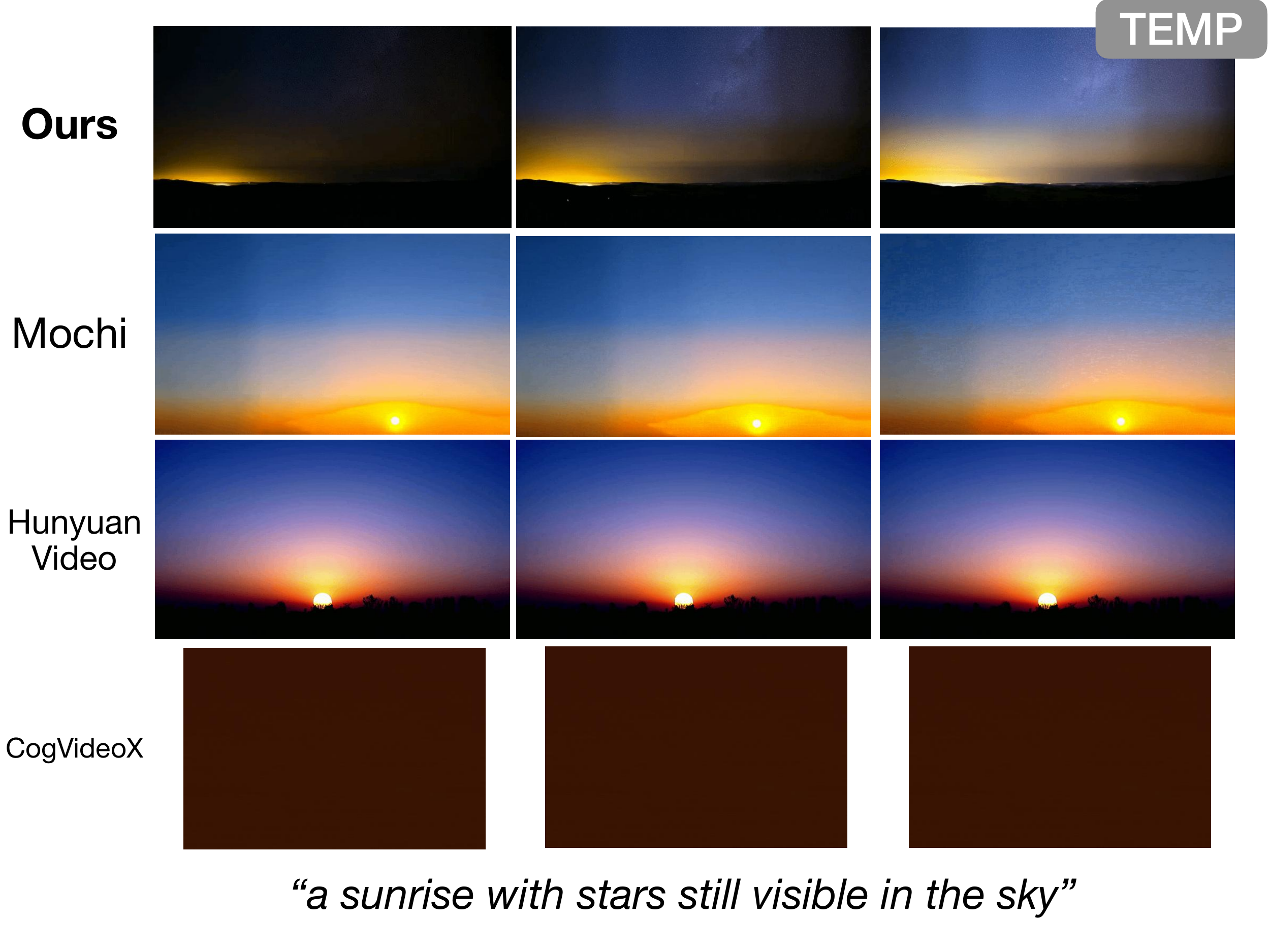}
\caption{\textbf{TEMP.} Prompt: ``a sunrise with stars still visible in the sky.''}
\label{fig:supp_temp_1}
\end{figure*}

\begin{figure*}[t]
\centering
\includegraphics[width=0.95\linewidth]{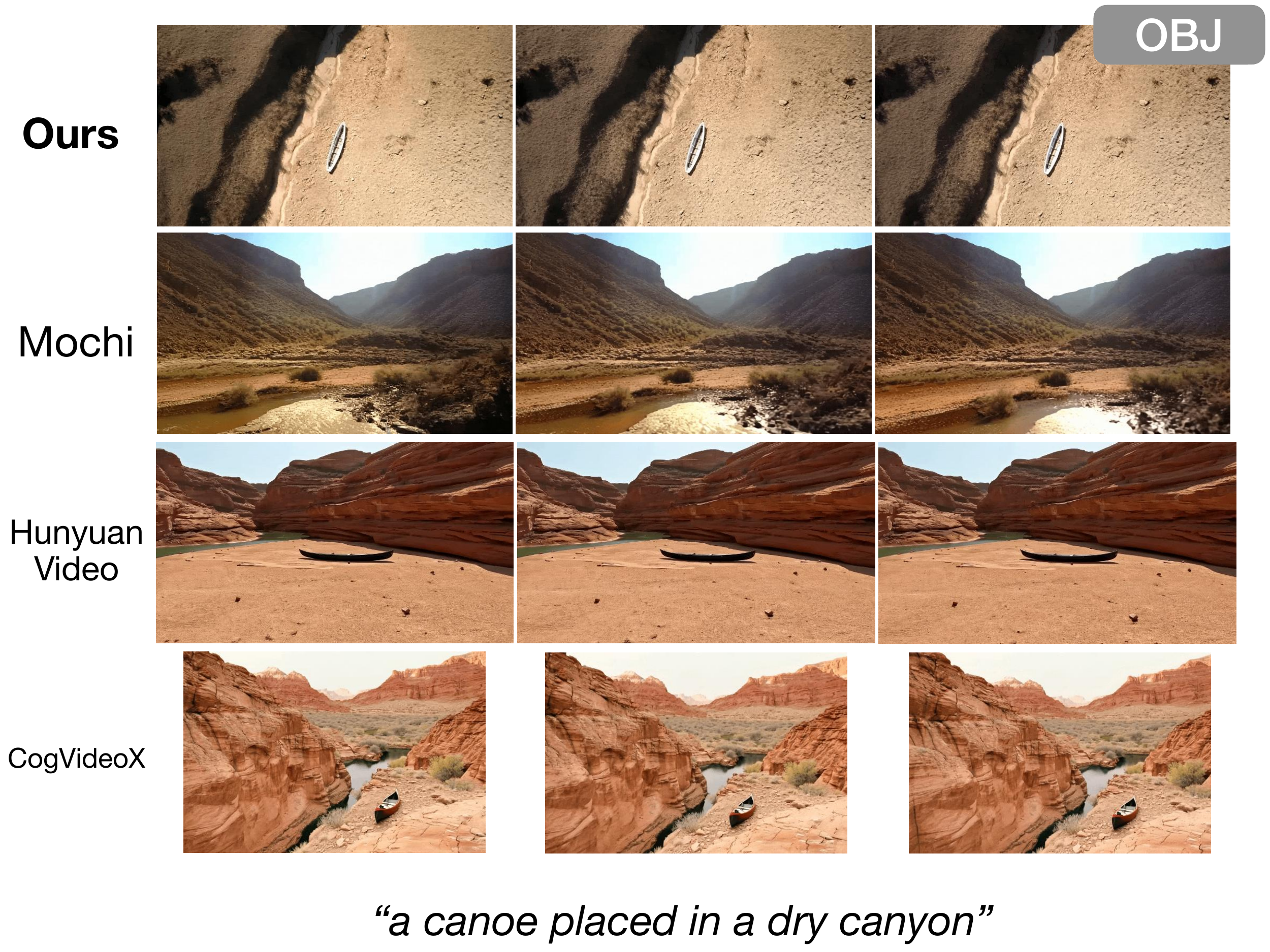}
\caption{\textbf{OBJ.} Prompt: ``a canoe placed in a dry canyon.''}
\label{fig:supp_obj_1}
\end{figure*}

\begin{figure*}[t]
\centering
\includegraphics[width=0.95\linewidth]{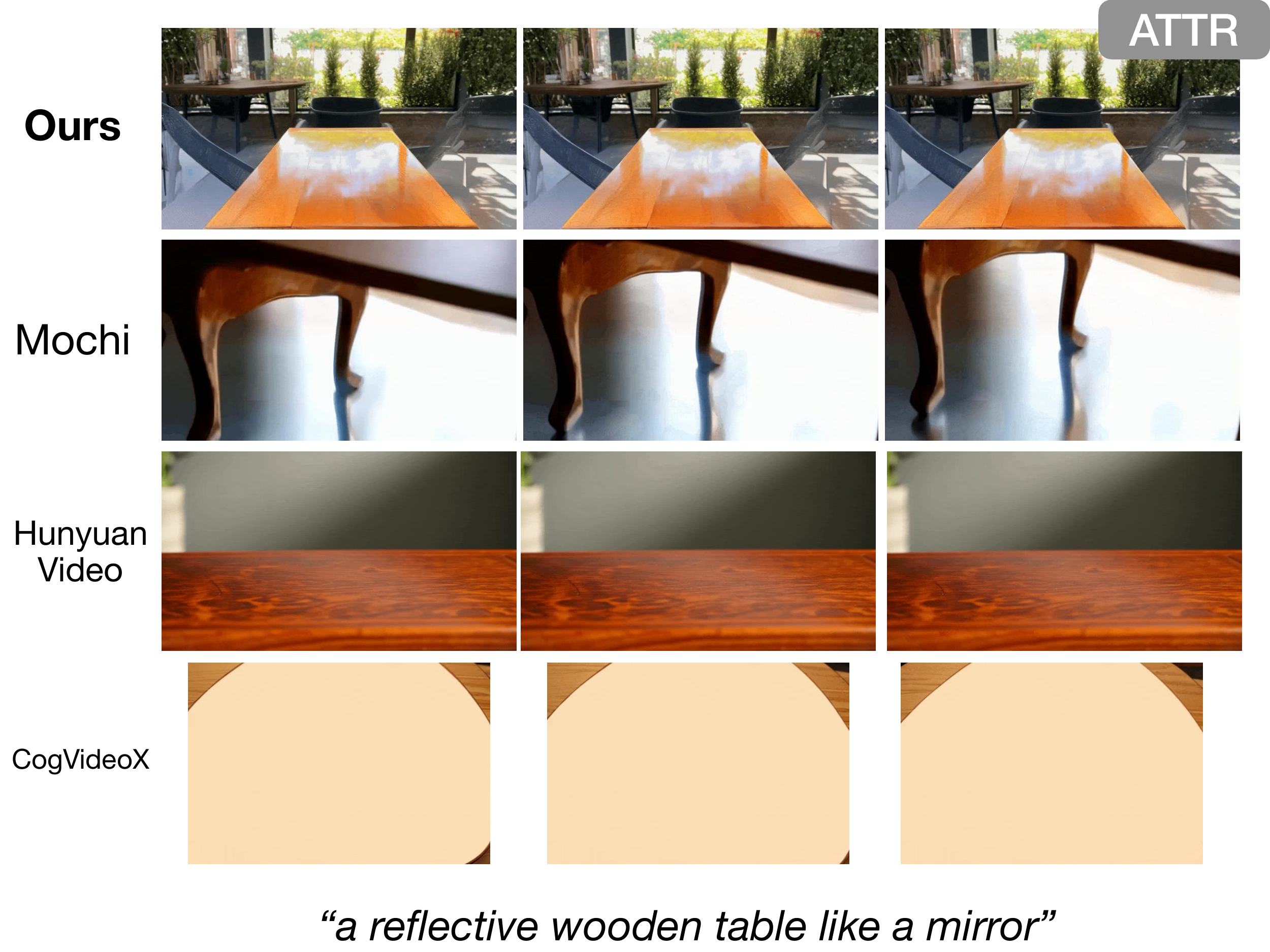}
\caption{\textbf{ATTR.} Prompt: ``a reflective wooden table like a mirror.''}
\label{fig:supp_attr_1}
\end{figure*}

\begin{figure*}[t]
\centering
\includegraphics[width=0.95\linewidth]{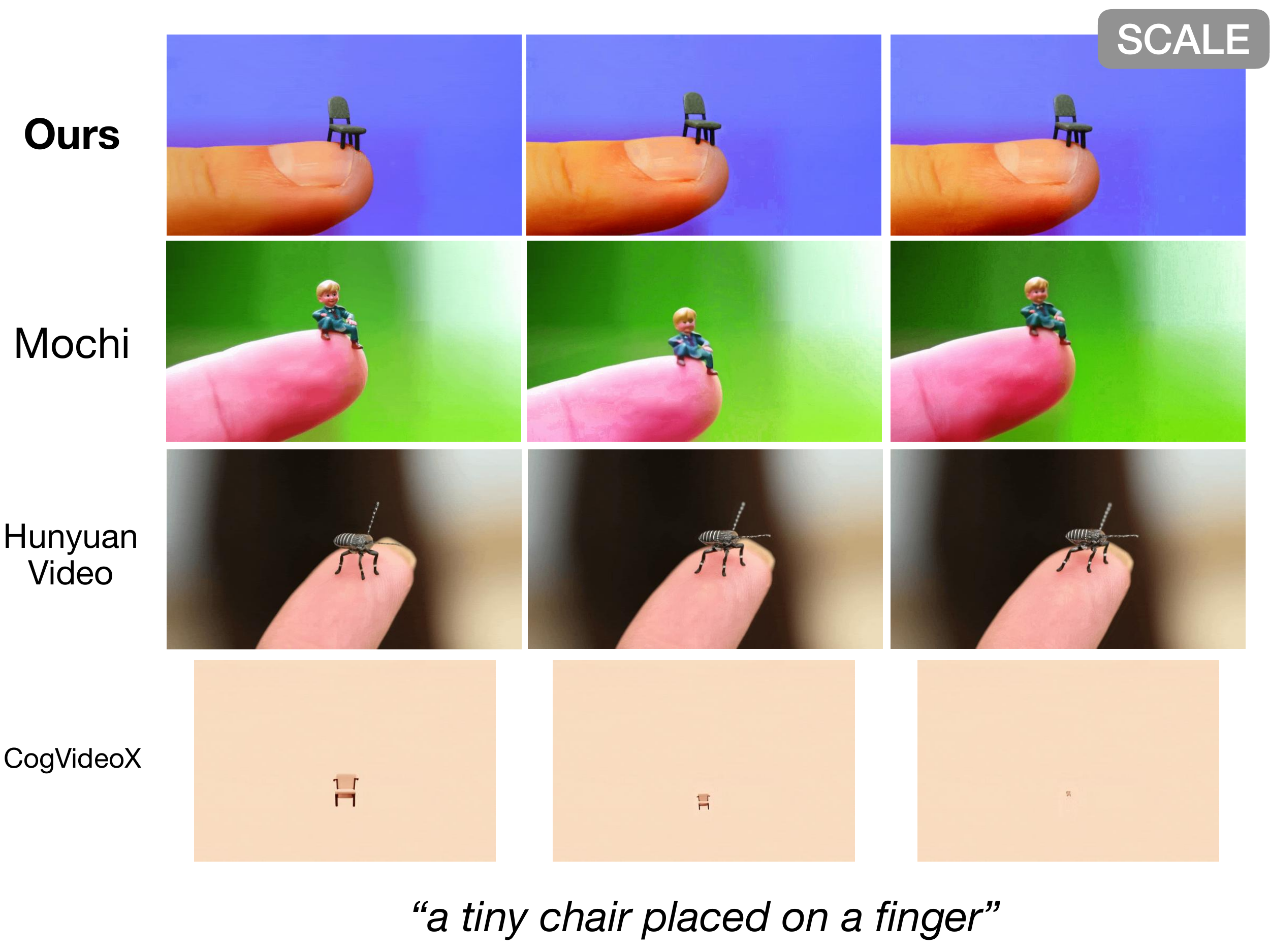}
\caption{\textbf{SCALE.} Prompt: ``a tiny chair placed on a finger.''}
\label{fig:supp_scale_1}
\end{figure*}

\begin{figure*}[t]
\centering
\includegraphics[width=0.95\linewidth]{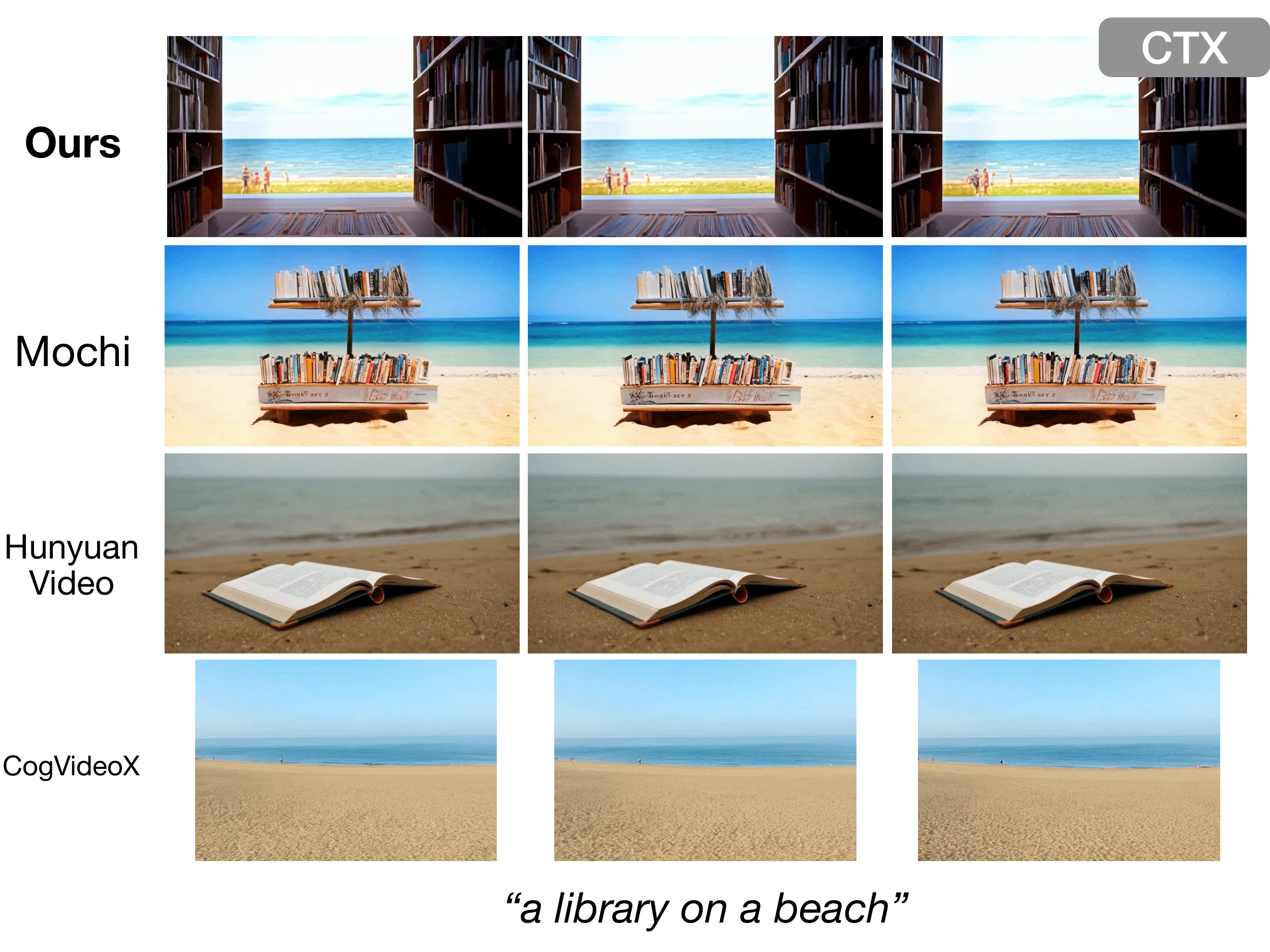}
\caption{\textbf{CTX.} Prompt: ``a library on a beach.''}
\label{fig:supp_ctx_1}
\end{figure*}

\begin{figure*}[t]
\centering
\includegraphics[width=0.95\linewidth]{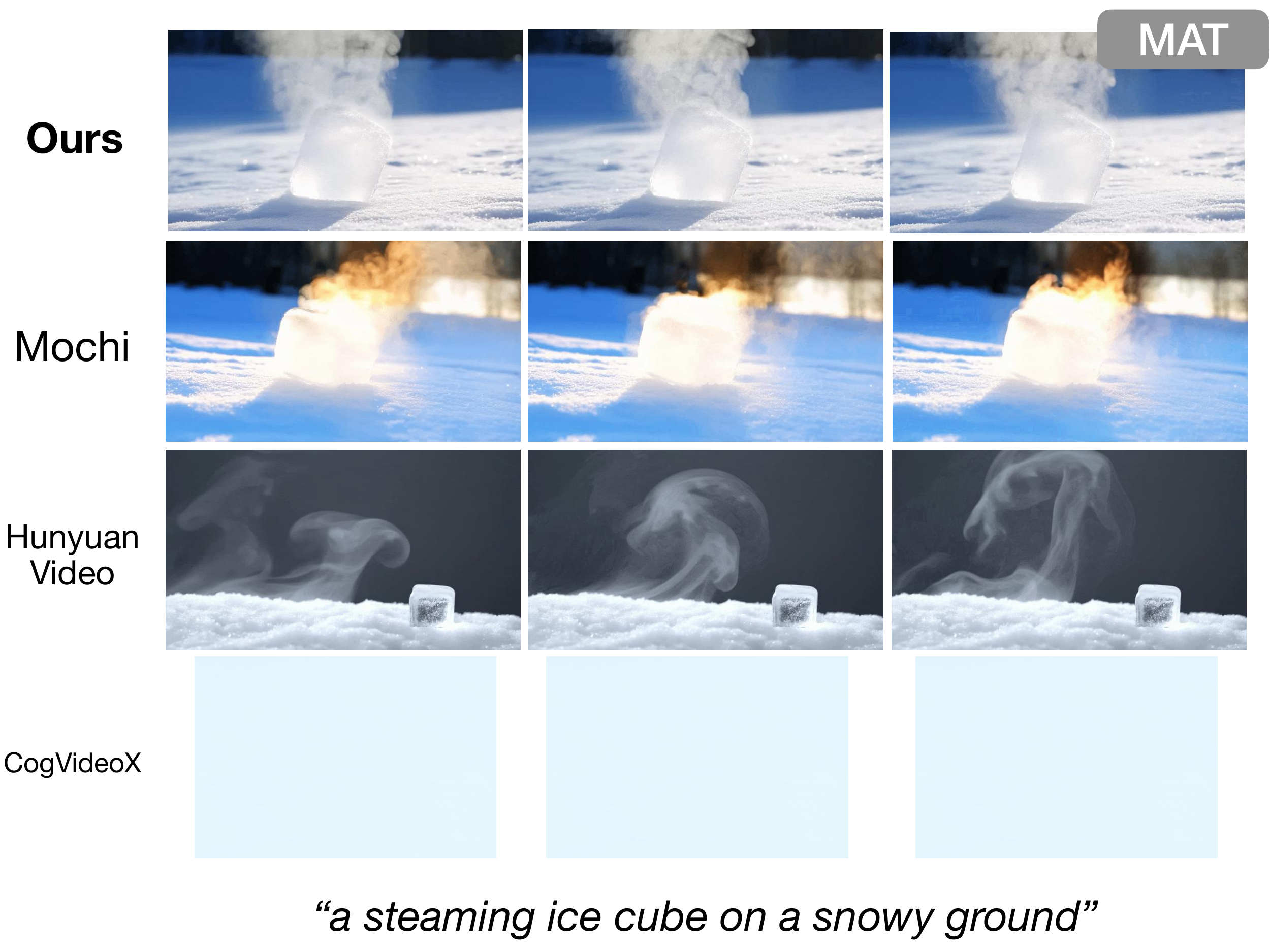}
\caption{\textbf{MAT.} Prompt: ``a steaming ice cube on a snowy ground.''}
\label{fig:supp_mat_1}
\end{figure*}

\begin{figure*}[t]
\centering
\includegraphics[width=0.95\linewidth]{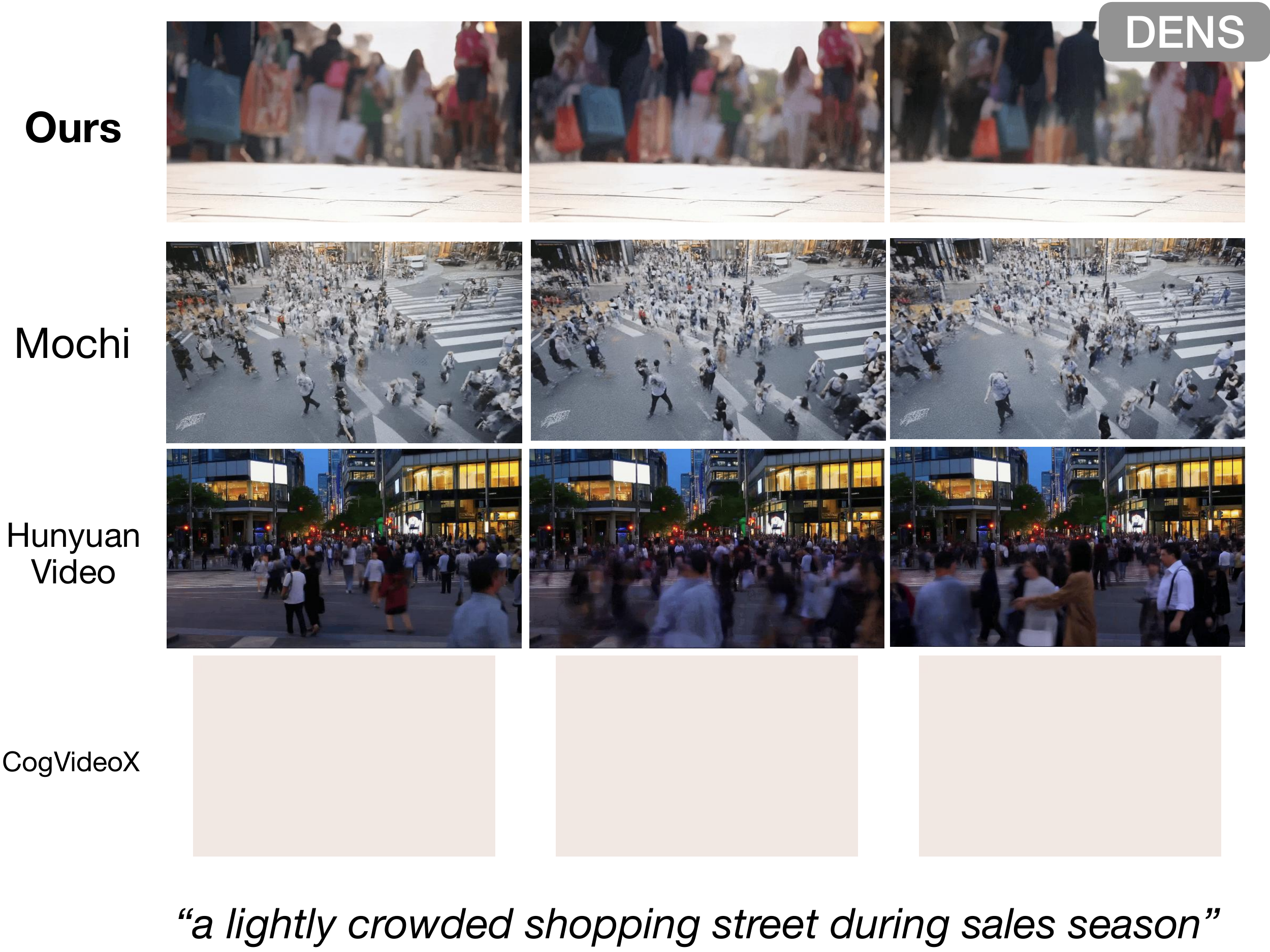}
\caption{\textbf{DENS.} Prompt: ``a lightly crowded shopping street during sales season.''}
\label{fig:supp_dens_1}
\end{figure*}


\begin{figure*}[t]
\centering
\includegraphics[width=0.95\linewidth]{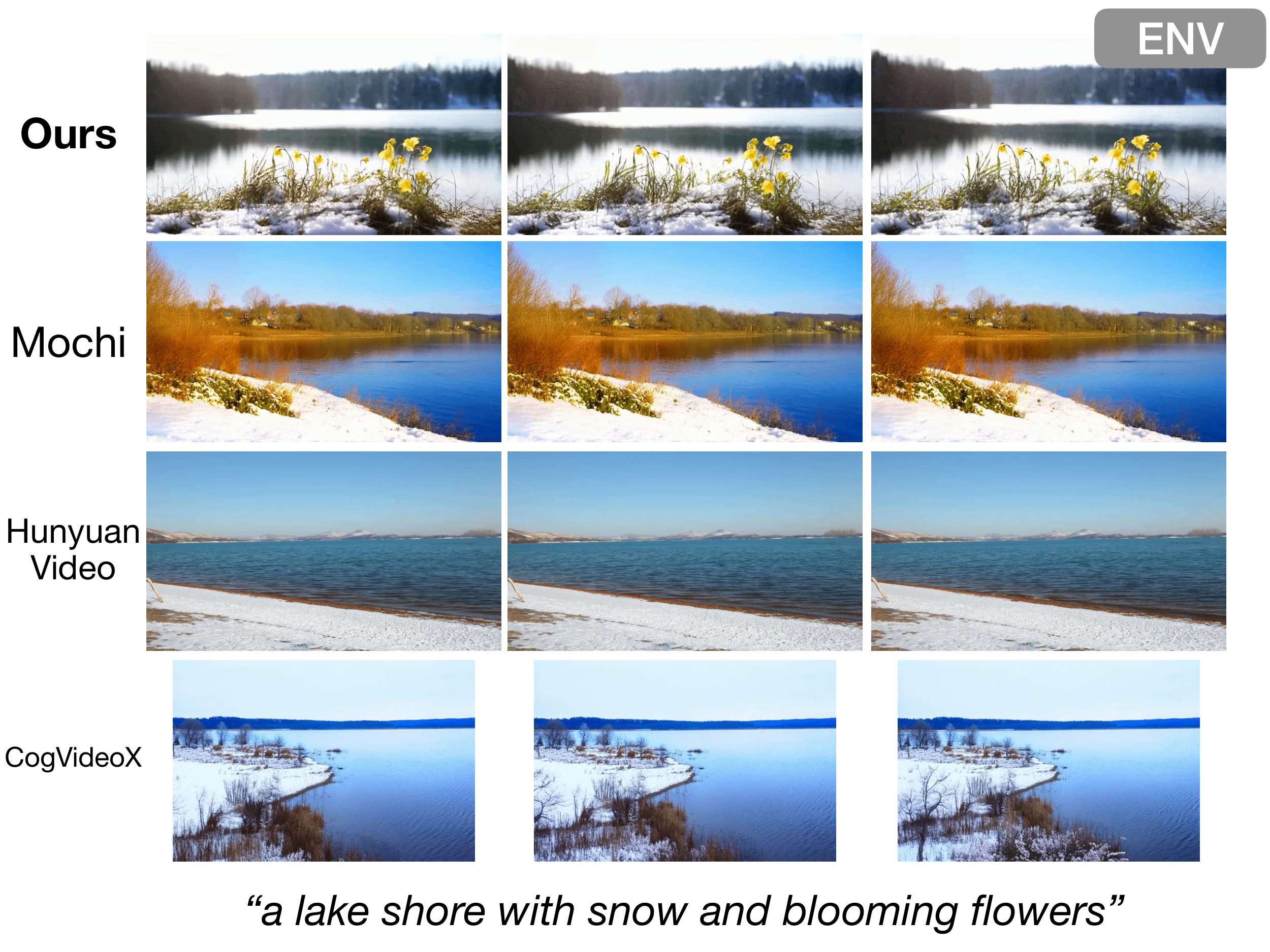}
\caption{\textbf{ENV.} Prompt: ``a lake shore with snow and blooming flowers.''}
\label{fig:supp_env_2}
\end{figure*}

\begin{figure*}[t]
\centering
\includegraphics[width=0.95\linewidth]{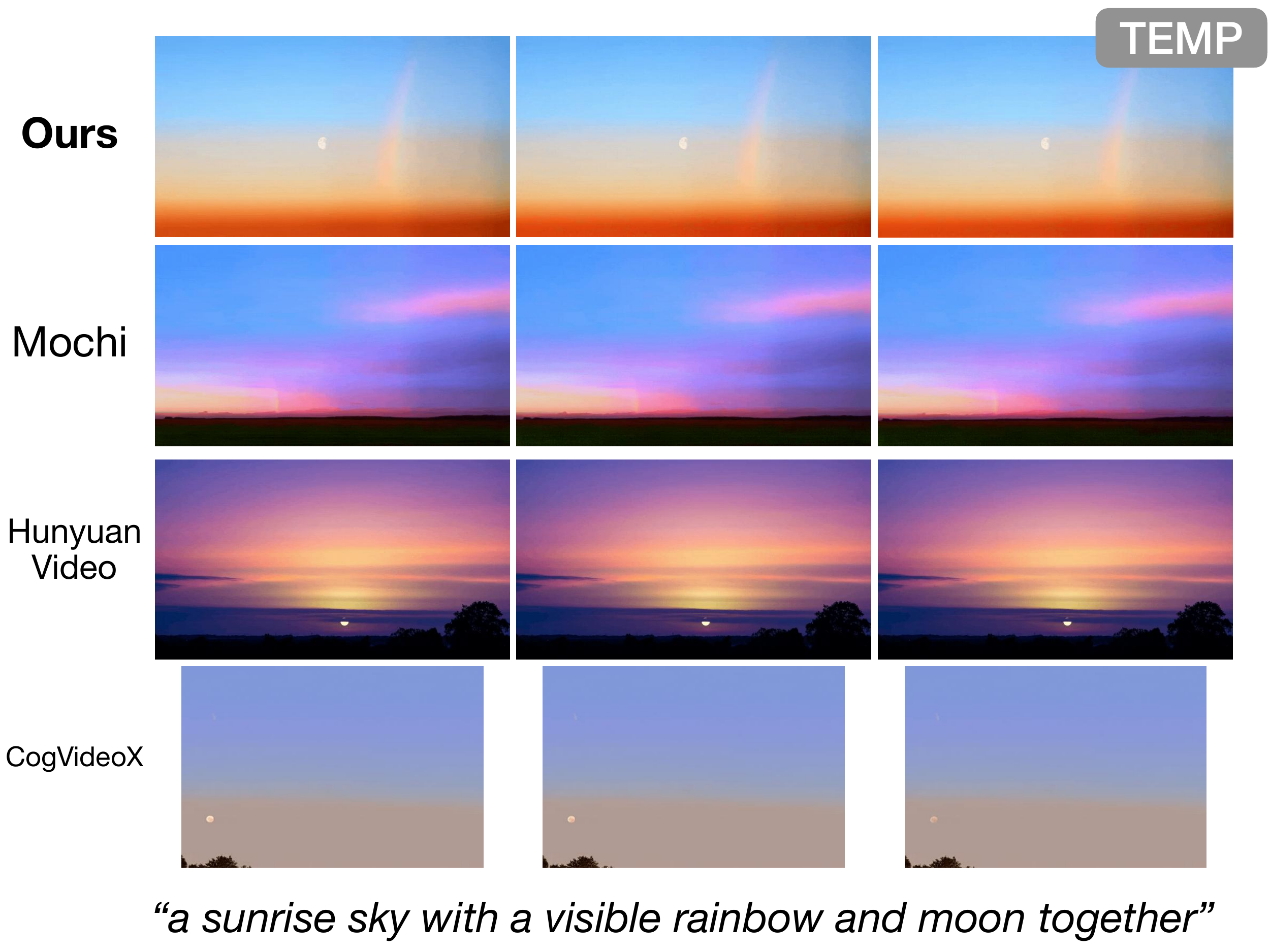}
\caption{\textbf{TEMP.} Prompt: ``a sunrise sky with a visible rainbow and moon together.''}
\label{fig:supp_temp_2}
\end{figure*}

\begin{figure*}[t]
\centering
\includegraphics[width=0.95\linewidth]{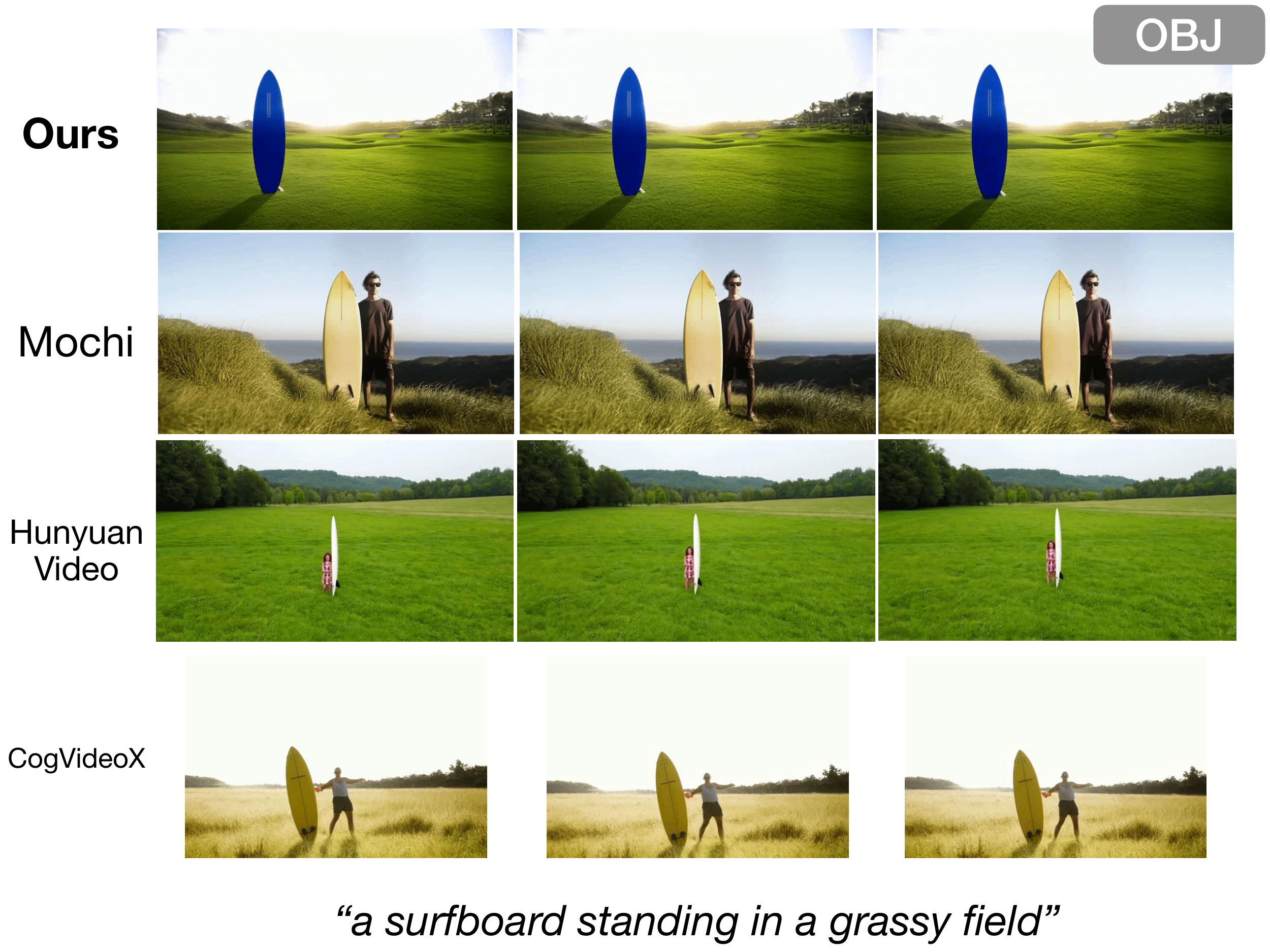}
\caption{\textbf{OBJ.} Prompt: ``a surfboard standing in a grassy field.''}
\label{fig:supp_obj_2}
\end{figure*}

\begin{figure*}[t]
\centering
\includegraphics[width=0.95\linewidth]{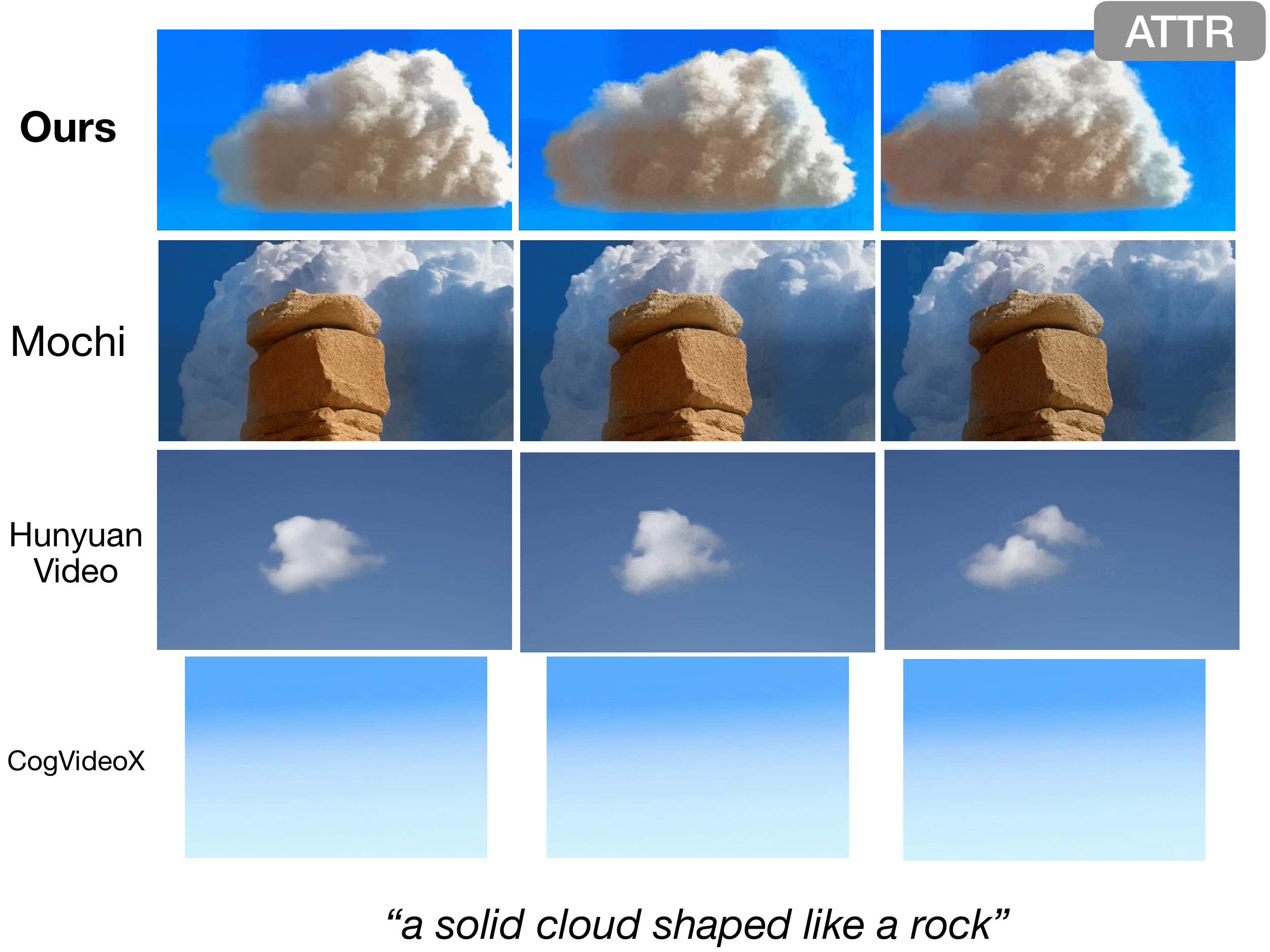}
\caption{\textbf{ATTR.} Prompt: ``a solid cloud shaped like a rock.''}
\label{fig:supp_attr_2}
\end{figure*}

\begin{figure*}[t]
\centering
\includegraphics[width=0.95\linewidth]{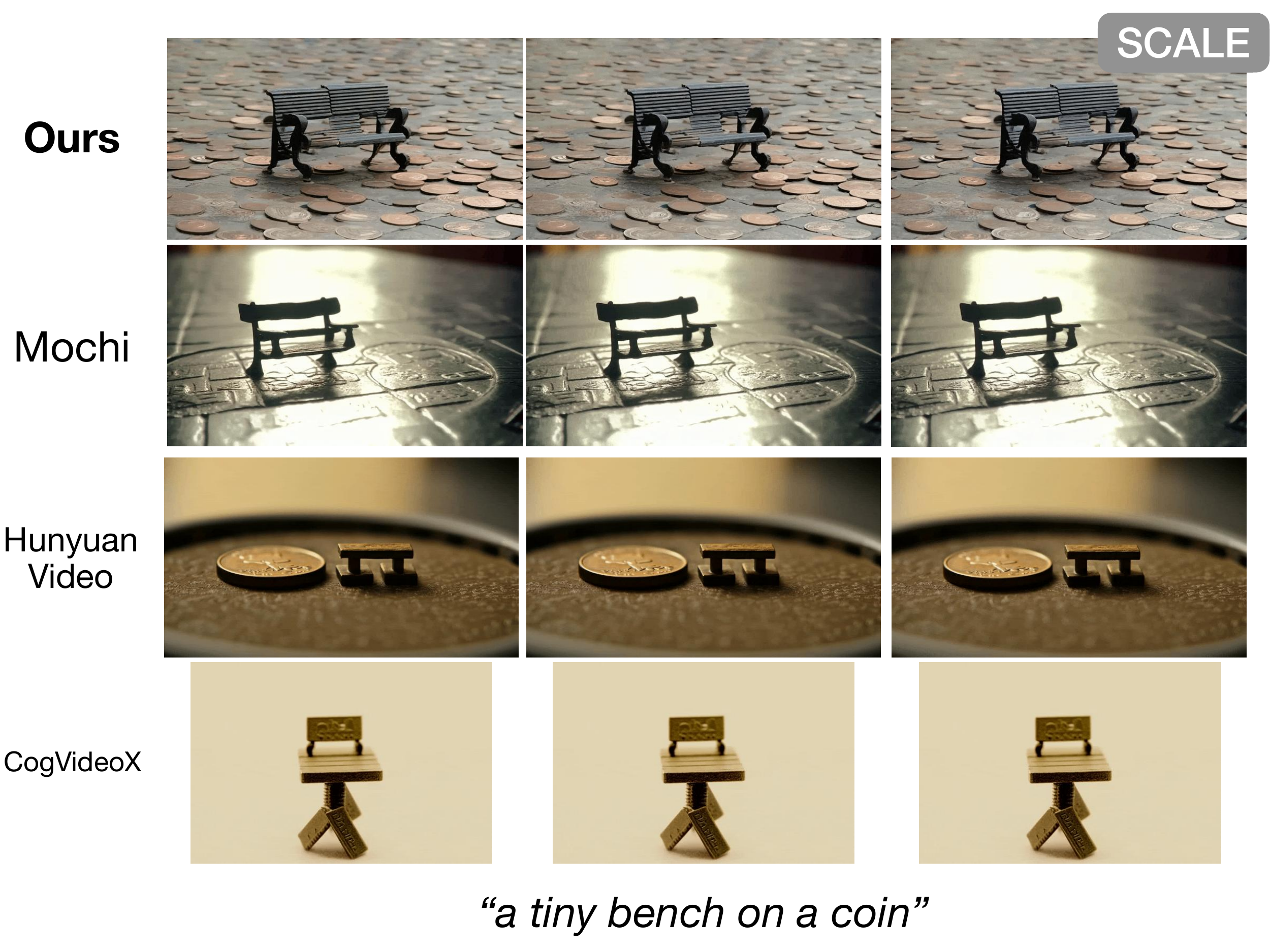}
\caption{\textbf{SCALE.} Prompt: ``a tiny bench on a coin.''}
\label{fig:supp_scale_2}
\end{figure*}

\begin{figure*}[t]
\centering
\includegraphics[width=0.95\linewidth]{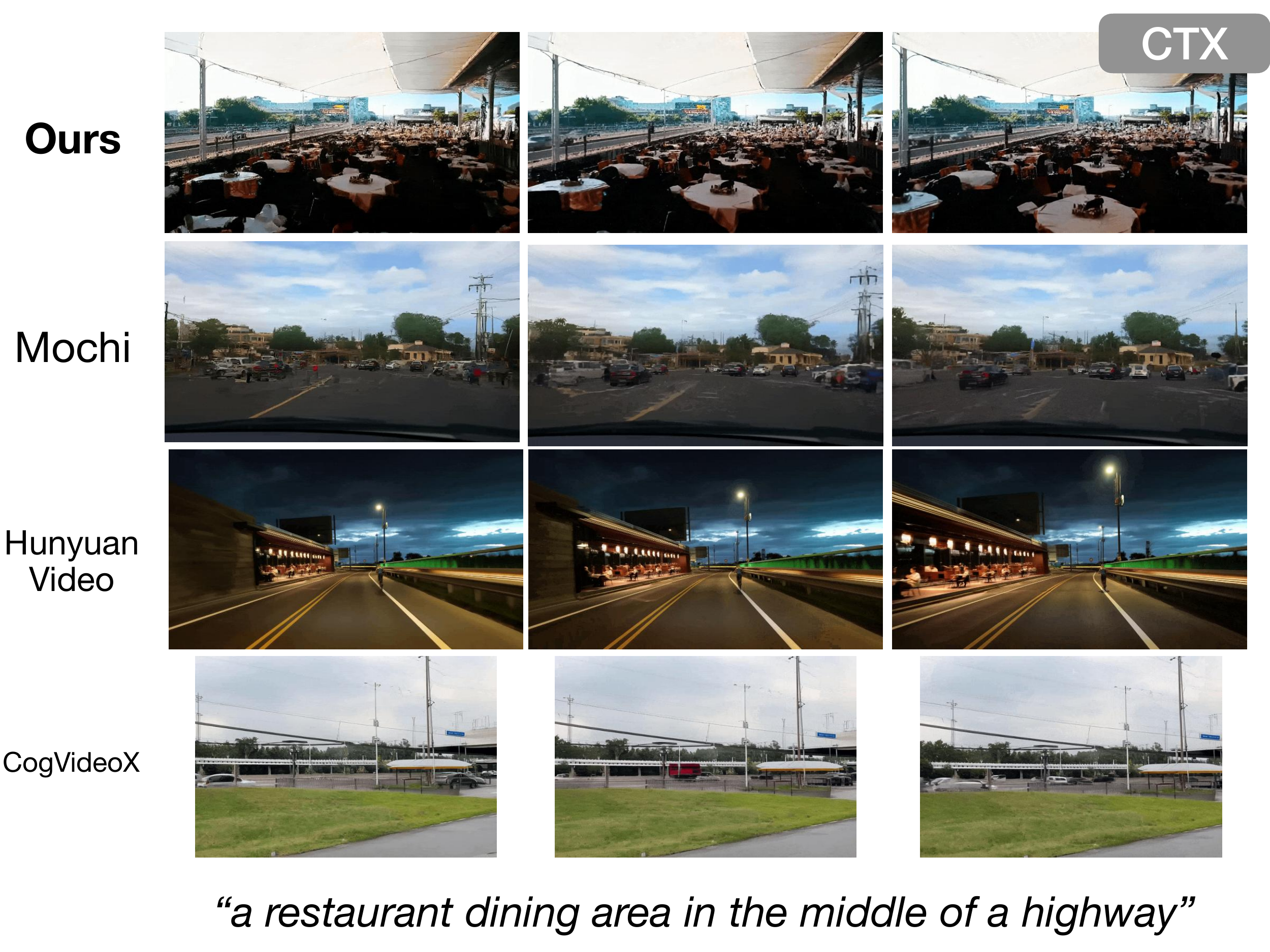}
\caption{\textbf{CTX.} Prompt: ``a restaurant dining area in the middle of a highway.''}
\label{fig:supp_ctx_2}
\end{figure*}

\begin{figure*}[t]
\centering
\includegraphics[width=0.95\linewidth]{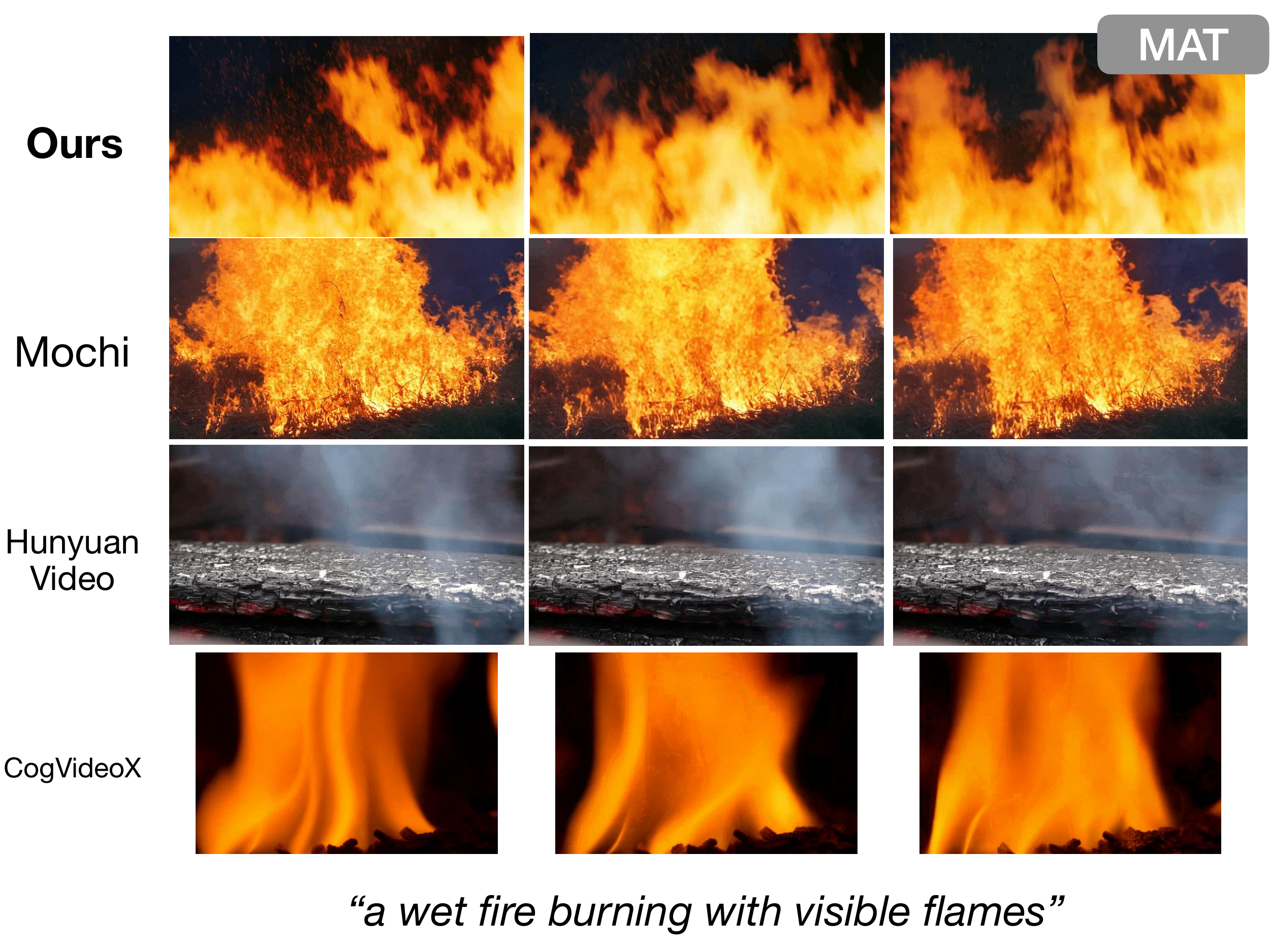}
\caption{\textbf{MAT.} Prompt: ``a wet fire burning with visible flames.''}
\label{fig:supp_mat_2}
\end{figure*}

\begin{figure*}[t]
\centering
\includegraphics[width=0.95\linewidth]{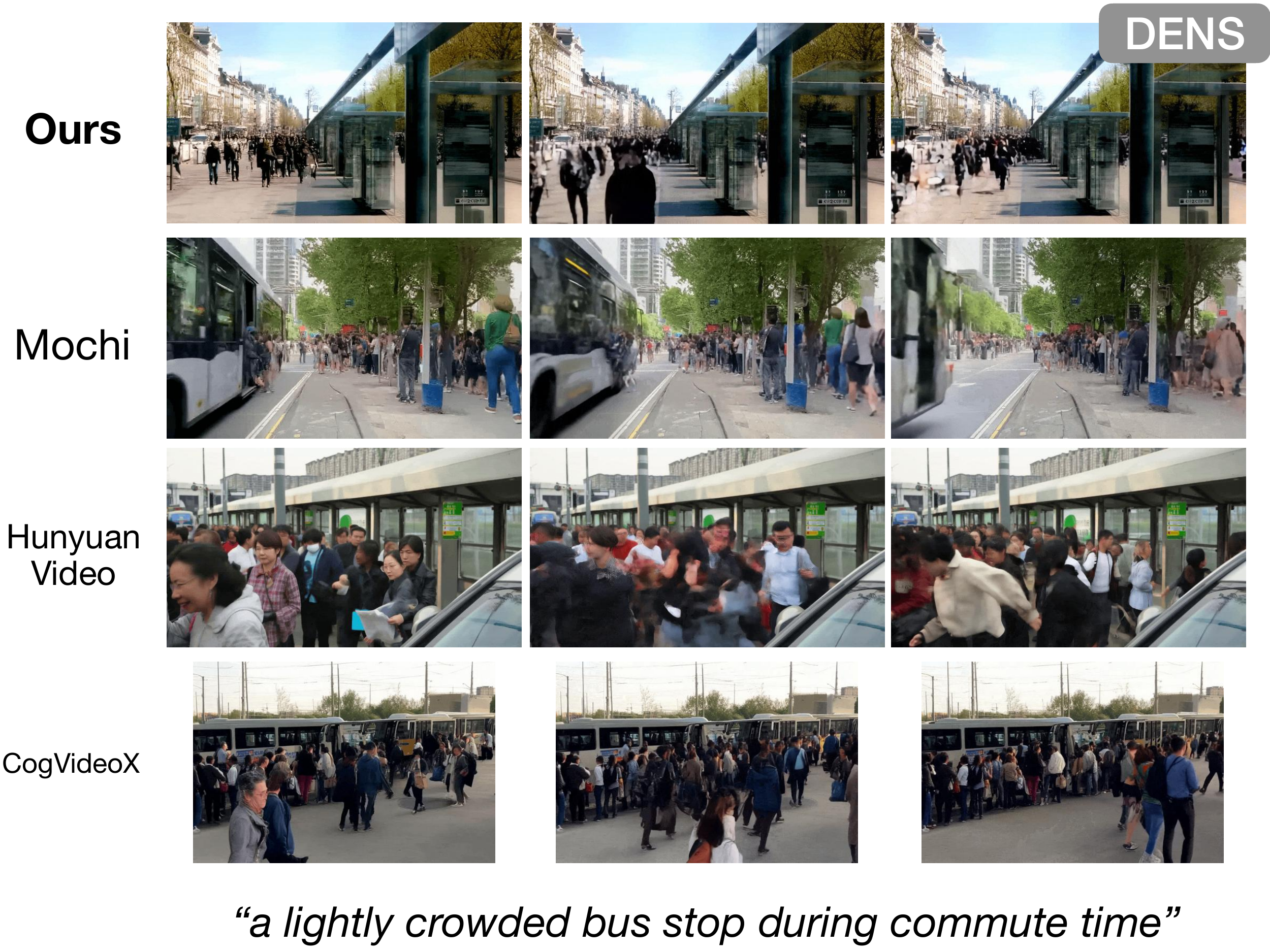}
\caption{\textbf{DENS.} Prompt: ``a lightly crowded bus stop during commute time.''}
\label{fig:supp_dens_2}
\end{figure*}

\end{document}